\documentclass[11pt]{article}

\usepackage[final]{acl}

\usepackage{times}
\usepackage{latexsym}

\usepackage[T1]{fontenc}

\usepackage[utf8]{inputenc}

\usepackage{microtype}

\usepackage{inconsolata}

\usepackage{graphicx}
\usepackage{soul}
\usepackage{bm} 
\usepackage{amsmath} 
\usepackage{amssymb}
\usepackage{booktabs}
\usepackage{multirow}

\usepackage{CJKutf8}
\usepackage{graphicx} 
\def\myrobotgood{\resizebox{0.6cm}{!}{\includegraphics{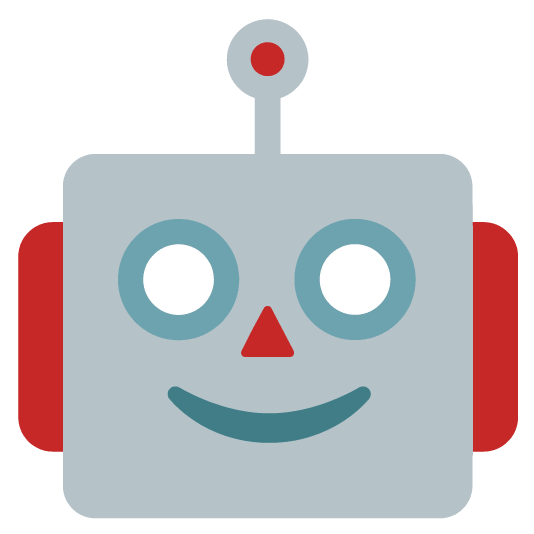}}} 
\def\myrobotthink{\resizebox{0.6cm}{!}{\includegraphics{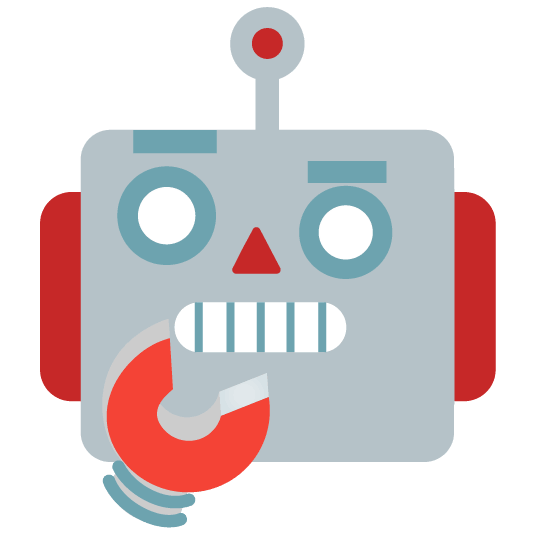}}} 
\def\mythink{\resizebox{0.5cm}{!}{\includegraphics{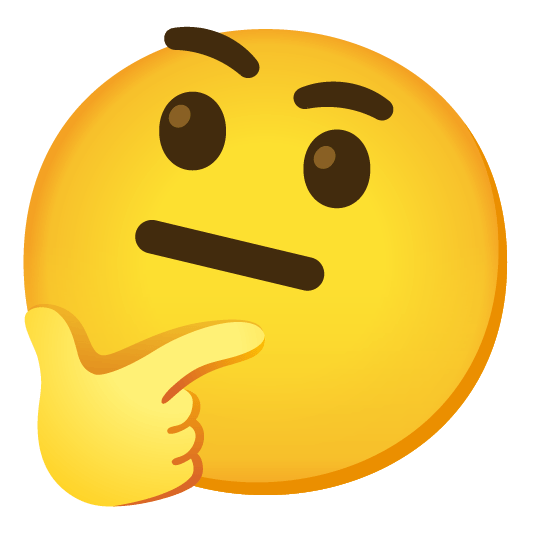}}} 
\def\myright{\resizebox{0.3cm}{!}{\includegraphics{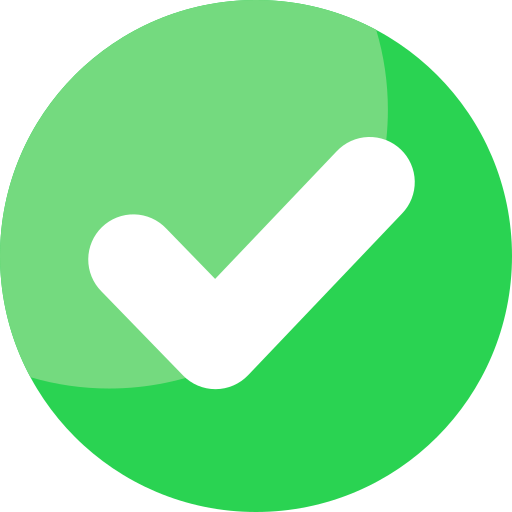}}} 
\def\mywrong{\resizebox{0.3cm}{!}{\includegraphics{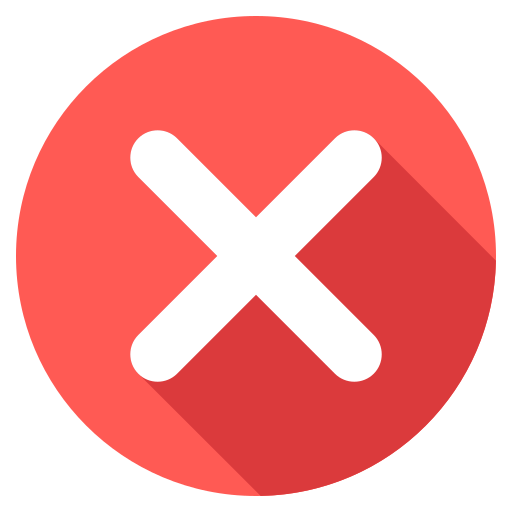}}} 
\def\mydoubt{\resizebox{0.5cm}{!}{\includegraphics{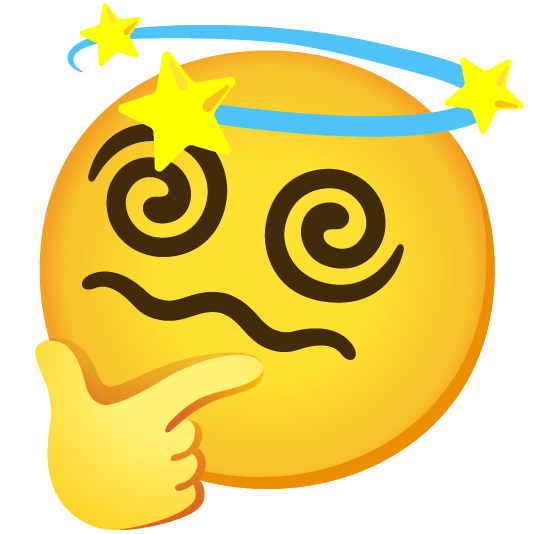}}} 

\definecolor{high-side}{RGB}{246, 205, 205}
\definecolor{high-inside}{RGB}{250, 228, 227}
\definecolor{medium-side}{RGB}{150, 199, 228}
\definecolor{medium-inside}{RGB}{228, 240, 247}
\definecolor{low-side}{RGB}{255, 204, 1}
\definecolor{low-inside}{RGB}{255, 252, 227}
\usepackage{tikz}

\usepackage{pgfplots}
\usepackage{subcaption}
\pgfplotsset{compat=newest}
\definecolor{tiffanyblue}{RGB}{129,216,208}
\definecolor{bangdiblue}{RGB}{0,149,182}
\definecolor{kleinblue}{RGB}{0,47,167}
\definecolor{purple}{RGB}{138,43,226}
\usepackage{pgfplotstable}
\usetikzlibrary{shapes}
\usepackage{circledtext}
\usetikzlibrary{arrows,decorations.pathmorphing,backgrounds,positioning,fit,petri}
\usetikzlibrary{arrows.meta,fit,shapes.arrows}
\usepgfplotslibrary{groupplots}
\usepackage{tikz}
\usetikzlibrary{positioning,shapes,arrows,shadows,patterns}
\usetikzlibrary{calc}
\usepackage{graphicx}

\definecolor{upurple}{RGB}{155,89,182}
\definecolor{ured}{RGB}{231,76,60}
\definecolor{udark}{RGB}{77,153,77}
\definecolor{udpdark}{HTML}{85827a}
\definecolor{usemidark}{HTML}{8c564b}
\definecolor{ublue}{RGB}{52,152,219}
\definecolor{udpblue}{HTML}{0419fb}
\definecolor{usemiblue}{HTML}{17becf}
\definecolor{uorange}{HTML}{ffcc00}
\definecolor{udporange}{HTML}{bcbd22}
\usepackage{makecell}
\usepackage{tabularx}
\usepackage{xcolor}
\usepackage[table]{xcolor} 
\usepackage{xspace}
\usepackage{arydshln}
\usepackage[super]{nth}
\newcommand{\headercolorSLAM}{\rowcolor[RGB]{221, 232, 250}}

\usepackage[ruled,vlined,linesnumbered]{algorithm2e}
\SetAlgoLined
\DontPrintSemicolon

\usepackage[ruled,vlined,linesnumbered]{algorithm2e}
\DontPrintSemicolon
\SetAlgoLined

\SetKwInput{KwHyper}{Hyperparameters}
\SetKwInput{KwIn}{Input}
\SetKwInput{KwInit}{Initialization}

\SetKwFor{For}{for}{do}{end}
\SetKwFor{ForEach}{foreach}{do}{end}
\SetKwIF{If}{ElseIf}{Else}{if}{then}{else if}{else}{end}

\newcommand{\ours}{\textsc{Lang}\xspace}

\usepackage{enumitem}

\usepackage{tcolorbox}
\usepgfplotslibrary{fillbetween}
\pgfplotsset{compat=1.16}  

\tcbuselibrary{most}

\tcbset{
  findingbox/.style={
    colback=gray!5,             
    colframe=black!30,          
    coltitle=white,             
    colbacktitle=black!60,      
    fonttitle=\small\bfseries,      
    boxrule=0.3pt,              
    arc=2pt,                    
    left=6pt, right=6pt, top=2pt, bottom=2pt, 
    enhanced,
    boxed title style={size=small, colframe=black!60, colback=black!60, sharp corners,
    top=1pt, bottom=1pt, left=2pt, right=2pt
    },
  }
}

\title{\ours: Reinforcement Learning for Multilingual Reasoning with Language-Adaptive Hint Guidance}

\author{Yuchun Fan\textsuperscript{1}\thanks{Work done during internship at Meituan.}, Bei Li\textsuperscript{2}\thanks{\xspace\xspace Corresponding author.}, Peiguang Li\textsuperscript{2}, Yilin Wang\textsuperscript{1}, Yongyu Mu\textsuperscript{1},  Jian Yang\textsuperscript{2}, Xin Chen\textsuperscript{2}, \\ 
\bf Rongxiang Weng\textsuperscript{2}, Jingang Wang\textsuperscript{2}, Xunliang Cai\textsuperscript{2}, Jingbo Zhu\textsuperscript{1,3},
{\bf Tong Xiao\textsuperscript{1,3}\footnotemark[2]  }\\
	\textsuperscript{1} \normalsize{NLP Lab, School of Computer Science and Engineering, Northeastern University, Shenyang, China} \\  
    \textsuperscript{2}Meituan Inc.
	\textsuperscript{3}NiuTrans Research, Shenyang, China\\
	\ttfamily{yuchunfan\_neu@outlook.com} \ttfamily{\{xiaotong,zhujingbo\}@mail.neu.edu.cn}
}

\begin{document}
\maketitle
\begin{abstract}

Reinforcement learning has proven effective for enhancing multi-step reasoning in large language models (LLMs), yet its benefits have not fully translated to multilingual contexts. Existing methods struggle with a fundamental trade-off: prioritizing input-language consistency severely hampers reasoning quality, while prioritizing reasoning often leads to unintended language drift toward English. We address this challenge with \ours, a novel framework that leverages language-conditioned hints to guide exploration in non-English reasoning tasks. Our method incorporates two key mechanisms to prevent dependency on these hints: a progressive decay schedule that gradually withdraws scaffolding, and a language-adaptive switch that tailors learning horizons to specific language difficulties. Empirical results on challenging multilingual mathematical benchmarks reveal that \ours substantially enhances reasoning performance without compromising language consistency. Moreover, we show that our framework generalizes beyond mathematics, fostering more consistent language alignment across model layers\footnote{The project will be available at: \url{https://github.com/fmm170/LANG}}.

\end{abstract}

\section{Introduction}
Recent years have witnessed the rise of large reasoning models such as OpenAI-o3~\citep{openai2024o3o4} and DeepSeek-R1~\citep{guo2025deepseek}, which leverage reinforcement learning with verifiable rewards (RLVR) to incentivize multi-step reasoning capabilities. However, these advances remain largely English-centric, leaving a critical performance gap in non-English settings, especially for low-resource languages~\citep{wang2025polymath,luo-etal-2025-mmath}.

\begin{figure}[!t]
    \centering
    \begin{tikzpicture}
\definecolor{my-gray}{RGB}{247, 247, 247}
\definecolor{my-purple}{RGB}{217, 217, 249}
\definecolor{my-blue}{RGB}{233, 246, 252}
\definecolor{upgray}{RGB}{128, 128, 128}
\definecolor{my-pink}{RGB}{255, 252, 227}
\tikzstyle{background} = [draw, fill=white, dashed, drop shadow, dash pattern=on 1.5pt off 1.5pt,line width=0.5pt, rounded corners=1.5pt,minimum width=7.4cm,minimum height=7.1cm]
\tikzstyle{background1} = [draw=white,, fill=my-gray, drop shadow, line width=0.5pt, rounded corners=1.5pt,minimum width=7cm,minimum height=3.4cm]
\tikzstyle{background2} = [draw=white,, fill=my-gray, drop shadow, line width=0.5pt, rounded corners=1.5pt,minimum width=7cm,minimum height=3cm]
\tikzstyle{user} = [draw=white,, fill=my-purple, drop shadow, line width=0.5pt, rounded corners=3pt,minimum width=6cm,minimum height=0.5cm]
\tikzstyle{robot-text} = [draw=white,, fill=my-blue, drop shadow, line width=0.5pt, rounded corners=3pt,minimum width=6cm,minimum height=0.5cm]
\tikzstyle{person-text} = [draw=white,, fill=my-pink, drop shadow, line width=0.5pt, rounded corners=3pt,minimum width=6cm,minimum height=0.5cm]
  \node[background,anchor=north](background) at (0.0, 0cm){};
  \node[background1,anchor = south,font=\tiny] (background1) at ([xshift = -0.05cm,yshift=-3.6cm]background.north){};
  \node[anchor=south, font=\tiny] (text1)
  at ([xshift=-1.7cm,yshift=-0.4cm]background1.north)
  {{\fontfamily{ptm}\selectfont\bfseries \scriptsize{Language-Constrain Reasoning}}};
  \node[user,anchor = south,align=left,font=\tiny] (user1) at ([xshift = 1.8cm,yshift=-1.2cm]text1.north){User query: \begin{CJK}{UTF8}{mj} 196의 양의 정수 약수는 모두 몇 개입니까?\end{CJK}\\\begin{CJK}{UTF8}{mj} 한국어로 생각하고 답변하세요. \end{CJK}\textcolor{upgray}{(Use Korean to think and answer.)} };

\node[anchor = south,font=\footnotesize] (robot1) at ([xshift = -3.1cm,yshift=-2cm]background1.north) {\myrobotthink};
\node[robot-text,anchor = south,align=left,font=\tiny] (robot1-text) at ([xshift = 0.3cm,yshift=-2.5cm]background1.north){<think> \begin{CJK}{UTF8}{mj} 먼저 유클리드 호제법을 사용하여 $196=2\times98$로 \end{CJK}\\\begin{CJK}{UTF8}{mj} 나타낸다. 이어서 98을 소인수분해한다[...]. $196$의 약수에\end{CJK}\\\begin{CJK}{UTF8}{mj} 대한 소인수분해의 경우의 수는 $3\times2=6$가지이다.</think>\end{CJK}\\\begin{CJK}{UTF8}{mj}정답은\texttt{\textbackslash boxed\{6\}}이다.\end{CJK}};
\node[anchor = south,font=\footnotesize] (right1) at ([xshift = 2.6cm,yshift=-1.3cm]robot1-text.north) {\mywrong};
\node[person-text,anchor = south,align=left,font=\tiny] (person1-text) at ([xshift = -0.5cm,yshift=-1.9cm]robot1-text.north){The reasoning trace is \textbf{language-consistent}, but the answer is \textbf{incorrect}.};
\node[anchor = south,font=\footnotesize] (person1) at ([xshift = 3.45cm,yshift=-0.66cm]person1-text.north) {\mythink};

\node[background2,anchor = south,font=\tiny] (background2) at ([xshift = 0cm,yshift=-6.7cm]background1.north){};
\node[anchor=south, font=\tiny] (text2)
  at ([xshift=-1.8cm,yshift=-0.4cm]background2.north)
  {{\fontfamily{ptm}\selectfont\bfseries \scriptsize{Language-Drifting Reasoning}}};
\node[user,anchor = south,align=left,font=\tiny] (user2) at ([xshift = 1.8cm,yshift=-1cm]text2.north){User query: \begin{CJK}{UTF8}{mj} 196의 양의 정수 약수는 모두 몇 개입니까?\end{CJK}};
\node[anchor = south,font=\footnotesize] (robot2) at ([xshift = -3.1cm,yshift=-1.8cm]background2.north) {\myrobotgood};
\node[robot-text,anchor = south,align=left,font=\tiny] (robot2-text) at ([xshift = 0.3cm,yshift=-2.1cm]background2.north){<think> First prime factorize $196=2^2\cdot7^2$. The prime factor \\decomposition of any divisor [...] to distinct integers.</think> \\The answer is \texttt{\textbackslash boxed\{9\}}.};
\node[person-text,anchor = south,align=left,font=\tiny] (person2-text) at ([xshift = -0.5cm,yshift=-1.7cm]robot2-text.north){The answer is \textbf{correct}, but the reasoning trace is \textbf{language-inconsistent}.};
\node[anchor = south,font=\footnotesize] (person2) at ([xshift = 3.45cm,yshift=-0.66cm]person2-text.north) {\mydoubt};
\node[anchor = south,font=\footnotesize] (right1) at ([xshift = 2.6cm,yshift=-1.1cm]robot2-text.north) {\myright};
\end{tikzpicture}
    \caption{The trade-off between answer accuracy and the language of reasoning trace in multilingual reasoning scenarios. The Korean question in English means “How many positive integer divisors are there of 196?”}
    \label{fig:traing_entropy}
\end{figure}
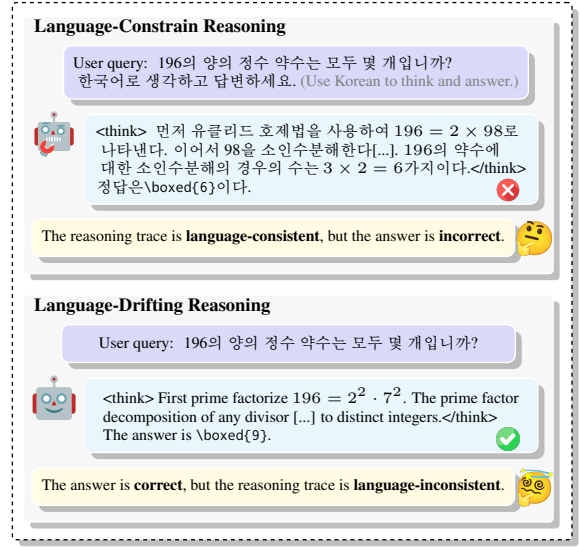

Beyond accuracy, when it comes to multilingual scenarios, language-consistent reasoning is equally crucial for native-user interpretability. However, as illustrated in Figure~\ref{fig:traing_entropy}, current large language models (LLMs) face two coupled challenges. First, when we explicitly require models to reason in the input language, strong reasoning capability in English does not reliably transfer across languages, leading to significant performance degradation in non-English settings. Second, when such language constraints are relaxed, models frequently drift toward English, reducing readability for native users. Consequently, it remains non-trivial to balance language consistency and reasoning performance.

To address these challenges, recent studies primarily follow two directions. One line of work introduces explicit prompt constraints to steer models to reason in the language of the user’s query during reinforcement learning (RL) rollouts or at inference time~\citep{qi-etal-2025-models,mGRPO}. However, such prompt-based steering offers limited control and often comes at the cost of reasoning quality. Building on this, another line of work further introduces language-consistency rewards in RL, encouraging models to align the output language with the input~\citep{zhang2025think}. Nevertheless, when the model's multilingual reasoning ability is limited, trajectories that satisfy such strict objectives are extremely scarce, exacerbating reward sparsity and hindering effective exploration during multilingual reasoning RL.

Inspired by prior studies that leverage hints (\textit{e.g.}, partial reasoning steps from expert demonstrations) to alleviate reward sparsity during RL training~\citep{zhang2025stephint,wang2025hint,UFT}, we propose \ours, a language-adaptive hint-guided RL framework for multilingual reasoning. \ours uses language-conditioned hints as scaffolding to bootstrap exploration in non-English settings. However, we find that keeping multilingual hints throughout training induces strong hint dependence, leading to substantial performance degradation when hints are unavailable at test time.
To address this, we draw inspiration from scheduled sampling~\citep{bengio2015scheduled,zhang-etal-2019-bridging,qian2021glancing}, and introduce a progressive hint-decay schedule that reduces hint exposure over training, shifting learning from hint-conditioned rollouts to autonomous multilingual reasoning. 
Moreover, considering the difference in learning difficulty across languages, we further employ a language-adaptive switch that sets language-specific horizons for turning off hints.

To validate the effectiveness of \ours, we conduct extensive experiments on two representative LLMs across different sizes, covering two challenging multilingual mathematical reasoning benchmarks: MMATH~\citep{luo-etal-2025-mmath} and PolyMath~\citep{wang2025polymath}. The results show that \ours effectively enhances multilingual reasoning ability while preserving language consistency, improving accuracy over vanilla GRPO by 24.1\% on MMATH and 18.7\% on PolyMath. Moreover, \ours can also be generalized to non-mathematical multilingual tasks. Further analysis highlights that our method achieves consistently language-consistent reasoning across layers.

\section{Related Work}
\subsection{Multilingual Mathematical Reasoning}

Recent advances in LRMs have significantly improved their multi-step reasoning capabilities. However, reasoning performance remains highly uneven across languages~\citep{fan-etal-2025-slam,zhang-etal-2025-cm,zhao2024large}, and models often suffer from language drift (\textit{i.e.}, responding in a language different from the user’s query), degrading both performance and user experience in non-English settings~\citep{wang2025polymath,luo-etal-2025-mmath,fan-etal-2025-language,wang2026dapt,zhang-etal-2025-multilingual}. Current approaches to enhancing multilingual reasoning under language consistency conditions mainly fall into two categories. One line of work~\citep{qi-etal-2025-models,luo-etal-2025-mmath} uses explicit constraints in prompts to guide models to reason in the user-desired language. Another line of research~\citep{park2025cross,yang2025parallel,mGRPO,hwang2025learn} introduces language-consistency rewards in reinforcement learning to further encourage alignment between input and output languages. Building upon this, our work attempts to leverage multilingual reasoning traces as explicit guidance to steer model exploration within specific language spaces, a perspective that remains underexplored.

\subsection{Hint-guided Reinforcement Learning}
A well-known challenge in RLVR is the reward sparsity issue: when task difficulty exceeds model capability, sampled trajectories can all be incorrect, yielding zero group-wise advantage~\citep{yu2025dapo,chen2026know,huang2026bootstrapping}. To mitigate this, prior studies~\citep{zhang2025stephint,wang2025hint,UFT} have focused on injecting hint-like partial solutions into prompts as in-context guidance to improve exploration. This issue is even more pronounced in multilingual reasoning, where non-English reasoning ability typically lags behind English, making it difficult to sample feasible trajectories. Building on this line of work, we introduce a language-adaptive hint decay strategy that scales guidance by language difficulty, effectively breaking the exploration bottleneck in multilingual reasoning.

\section{Preliminary Study}
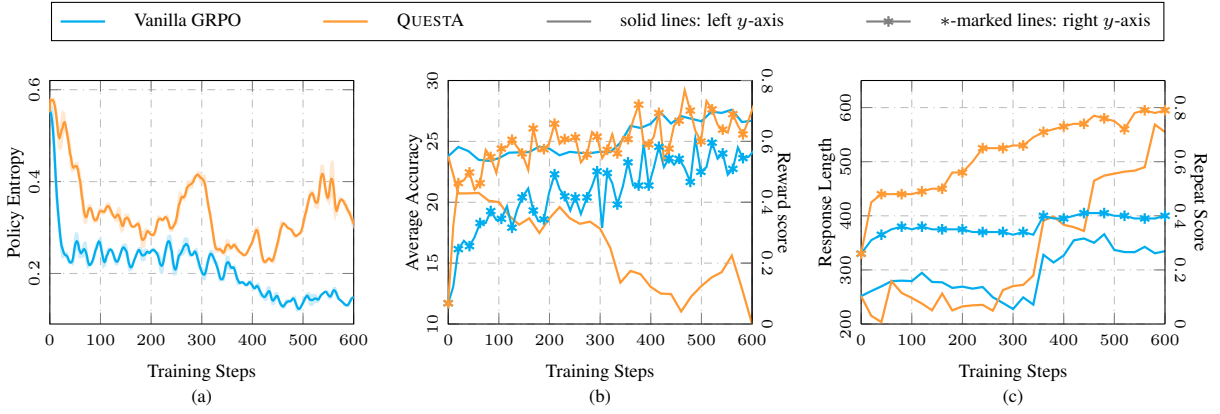
\begin{figure*}[!t]
    \centering
    \definecolor{upgray}{RGB}{128, 128, 128}
\begin{tikzpicture}
\begin{groupplot}[
    group style={
        group size=3 by 1,
        horizontal sep=1.25cm
    },
    width=0.35\textwidth,
    height=0.3\textwidth,
    xmin=0, xmax=600,
    xtick={0,100,200,300,400,500,600},
    ymajorgrids=true, xmajorgrids=true, grid style=dashdotted,
    scaled ticks=false,
    xlabel style={font=\scriptsize},
    ylabel style={font=\scriptsize,yshift=-0.5em},
    tick label style={font=\tiny},
    yticklabel style={font=\tiny, rotate=90},
]

\nextgroupplot[
    xlabel={\shortstack{Training Steps\\(a)}},
    ylabel=Policy Entropy,
    ymin=0.09, ymax=0.62,
    legend to name=shared_legend_training,
    legend columns=4,
legend style={
  draw=black,
  fill=none,
  font=\scriptsize,
  column sep=0.35cm,    
  legend image post style={xscale=0.7},
},
    legend style={
        draw=black,
        fill=none,
        font=\scriptsize,
        legend image post style={xscale=1.4},
        /tikz/every odd column/.append style={column sep=0.3cm},
        /tikz/every even column/.append style={column sep=1cm}
    }
]
\usepgfplotslibrary{fillbetween}

\addplot [name path=cyan_upper, draw=none, forget plot]
  table [x=Step, y=without-hint-upper, col sep=comma]
  {original_data/training-entropy-sample5-with-band.csv};

\addplot [name path=cyan_lower, draw=none, forget plot]
  table [x=Step, y=without-hint-lower, col sep=comma]
  {original_data/training-entropy-sample5-with-band.csv};

\addplot [cyan!20, opacity=0.9, forget plot] 
  fill between[of=cyan_upper and cyan_lower];

\addplot [sharp plot, cyan, line width=0.85pt]
  table [x=Step, y=without-hint-smooth, col sep=comma]
  {original_data/training-entropy-sample5-with-band.csv};
\addlegendentry{Vanilla GRPO}

\addplot [name path=orange_upper, draw=none, forget plot]
  table [x=Step, y=mul-hint-no-attenuate-upper, col sep=comma]
  {original_data/training-entropy-sample5-with-band.csv};

\addplot [name path=orange_lower, draw=none, forget plot]
  table [x=Step, y=mul-hint-no-attenuate-lower, col sep=comma]
  {original_data/training-entropy-sample5-with-band.csv};

\addplot [orange!20, opacity=0.9, forget plot] 
  fill between[of=orange_upper and orange_lower];

\addplot [sharp plot, orange!80, line width=0.85pt]
  table [x=Step, y=mul-hint-no-attenuate-smooth, col sep=comma]
  {original_data/training-entropy-sample5-with-band.csv};
\addlegendentry{\textsc{QuestA}}

\addplot [upgray, thick] coordinates {(0,0) (1,0)};
\addlegendentry{solid lines: left $y$-axis}

\addplot [upgray, thick, mark=asterisk,,mark size=2pt,
          mark options={fill=white, draw=upgray, line width=0.85pt}] 
         coordinates {(0,0) (1,0)};
\addlegendentry{$*$-marked lines: right $y$-axis}

\nextgroupplot[
    xlabel={\shortstack{Training Steps\\(b)}},
    ylabel=Average Accuracy,
    ymin=10, ymax=30,
]
\addplot [sharp plot, cyan, line width=0.85pt] coordinates {
    (0,23.78) (20,24.54) (40,24.20) (60,23.49) (80,23.40) (100,23.62)
    (120,24.07) (140,24.10) (160,24.13) (180,24.57) (200,24.40)
    (220,23.86) (240,24.14) (260,24.06) (280,24.03) (300,24.15)
    (320,24.12) (340,24.88) (360,26.28) (380,26.10) (400,26.45)
    (420,27.35) (440,26.47) (460,27.07) (480,26.86) (500,26.65)
    (520,27.44) (540,27.34) (560,27.60) (580,26.59) (600,26.70)
};

\addplot [sharp plot, orange!80, line width=0.85pt] coordinates {
    (0,23.78) (20,20.72) (40,20.73) (60,20.77) (80,20.13) (100,20.01)
    (120,18.90) (140,18.16) (160,18.68) (180,17.48) (200,18.69)
    (220,19.61) (240,18.69) (260,18.23) (280,18.41) (300,17.83)
    (320,16.26) (340,13.40) (360,14.36) (380,14.09) (400,13.04)
    (420,12.50) (440,12.46) (460,11.03) (480,12.27) (500,13.12)
    (520,13.82) (540,14.26) (560,15.62) (580,12.92) (600,10.01)
};

\nextgroupplot[
    xlabel={\shortstack{Training Steps\\(c)}},
    ylabel=Response Length,
    ymin=200, ymax=650,
    xshift=0.5em,
    ytick={200,300,400,500,600},
    axis y line*=left,
]
\addplot [sharp plot, cyan, line width=0.85pt] coordinates {
  (0,251.51) (20,260.90) (40,269.56) (60,279.21) (80,280.12) (100,279.31)
  (120,294.58) (140,277.74) (160,277.08) (180,266.77) (200,269.11)
  (220,265.35) (240,268.42) (260,249.49) (280,238.85) (300,227.72)
  (320,249.21) (340,235.90) (360,327.87) (380,313.73) (400,326.65)
  (420,354.67) (440,358.08) (460,350.09) (480,365.87) (500,336.95)
  (520,333.02) (540,332.77) (560,342.65) (580,330.73) (600,334.60)
};

\addplot [sharp plot, orange!80, line width=0.85pt] coordinates {
  (0,251.51) (20,215.21) (40,203.64) (60,279.12) (80,257.49) (100,248.24)
  (120,237.83) (140,225.32) (160,256.25) (180,225.16) (200,232.53)
  (220,234.49) (240,235.38) (260,224.61) (280,262.62) (300,270.14)
  (320,272.27) (340,289.71) (360,391.08) (380,398.77) (400,383.33)
  (420,378.99) (440,372.03) (460,464.75) (480,474.23) (500,477.89)
  (520,481.55) (540,483.41) (560,489.89) (580,568.64) (600,554.92)
};

\end{groupplot}

\begin{axis}[
    at={(group c2r1.center)},
    anchor=center,
    width=0.35\textwidth,
    height=0.3\textwidth,
    xmin=0, xmax=600,
    axis x line=none,
    axis y line*=right,
    ymin=0, ymax=0.8,
ytick={0,0.2,0.4,0.6,0.8},
    ylabel=Reward score,
    ylabel style={font=\scriptsize,yshift=0.5em,rotate=180},
    yticklabel style={font=\tiny,rotate=270},
]

\addplot[cyan,mark=asterisk,,mark size=2pt,thick,mark options={fill=white,draw=cyan,line width=0.85pt},mark repeat=2]
  table [x=Step, y=Value, col sep=comma]
  {original_data/reward-no-attenuate-sample10.csv};

\addplot [orange!80,mark=asterisk,,mark size=2pt,thick,mark options={fill=white,draw=orange!80,line width=0.85pt},mark repeat=2]
  table [x=Step, y=Value-no-attenuate, col sep=comma]
  {original_data/reward-no-attenuate-sample10.csv};

\end{axis}

\begin{axis}[
    at={(group c3r1.center)},
    anchor=center,
    width=0.35\textwidth,
    height=0.3\textwidth,
    xmin=0, xmax=600,
    ymin=0, ymax=0.9,
    ytick={0,0.2,0.4,0.6,0.8},
    axis x line=none,
    axis y line*=right,
    ylabel=Repeat Score,
    ylabel style={font=\scriptsize,yshift=0.5em,rotate=180},
    yticklabel style={font=\tiny,rotate=90,rotate=180},
]
\addplot [sharp plot, cyan, line width=0.85pt, mark=asterisk,mark size=2pt,thick,mark options={fill=white,draw=cyan,line width=0.85pt},mark repeat=2] coordinates {
  (0,0.26) (20,0.31) (40,0.33) (60,0.35) (80,0.36) (100,0.35)
  (120,0.36) (140,0.35) (160,0.35) (180,0.35) (200,0.35)
  (220,0.34) (240,0.34) (260,0.34) (280,0.34) (300,0.33)
  (320,0.34) (340,0.33) (360,0.40) (380,0.39) (400,0.39)
  (420,0.40) (440,0.41) (460,0.41) (480,0.41) (500,0.40)
  (520,0.40) (540,0.39) (560,0.39) (580,0.39) (600,0.40)
};

\addplot [sharp plot, orange!80, line width=0.85pt, mark=asterisk,mark size=2pt,thick,mark options={fill=white,draw=orange!80,line width=0.85pt},mark repeat=2] coordinates {
  (0,0.26) (20,0.45) (40,0.48) (60,0.48) (80,0.48) (100,0.48)
  (120,0.49) (140,0.50) (160,0.50) (180,0.56) (200,0.56)
  (220,0.60) (240,0.65) (260,0.65) (280,0.65) (300,0.66)
  (320,0.66) (340,0.69) (360,0.71) (380,0.72) (400,0.73)
  (420,0.74) (440,0.74) (460,0.77) (480,0.76) (500,0.75)
  (520,0.72) (540,0.78) (560,0.79) (580,0.78) (600,0.79)
};

\end{axis}
\node[anchor=south] at ($(group c2r1.north)+(0.1,1.0em)$) {%
  \makebox[\textwidth][c]{\pgfplotslegendfromname{shared_legend_training}}%
};

\end{tikzpicture}
\vspace{-0.2em}
    \vspace{-1.5em}
    \caption{The comparison of Vanilla GRPO and \textsc{QuestA} during RL training. \textcolor{cyan}{Blue curves} denote Vanilla GRPO and \textcolor{orange!80}{orange curves} denote \textsc{QuestA}. In the middle and right panels, solid lines correspond to the left $y$-axis, and $*$-marked lines correspond to the right $y$-axis.}
    \label{fig:prelimiary}
\end{figure*}
In this section, we conduct a pilot study of hint-guided RL for multilingual reasoning and analyze its training dynamics and generation behaviors, providing insights into the design of \ours.

\subsection{Experimental Setup for Pilot Study}

We adopt \textsc{QuestA}~\citep{li2025questa} as a representative hint-guided approach that augments training prompts with a fixed portion of distilled reasoning trajectories to guide exploration. We use Qwen2.5-7B-Instruct as the policy model and replace the original trajectories with multilingual counterparts constructed from DeepMath-103K~\citep{he2025deepmath}. We compare this variant with vanilla GRPO and evaluate its performance on MMATH~\citep{luo-etal-2025-mmath}. More details are provided in Appendix~\ref{sec:Pilot_study_setup}.
\subsection{Observations}
\label{Observations}
We report two observations below that highlight the limitations of naively applying multilingual hints during RL training.
\begin{tcolorbox}[findingbox, title=Finding 1]
High policy entropy and training reward do not reliably translate into better test-time multilingual reasoning performance.
\end{tcolorbox}
As shown in Figure~\ref{fig:prelimiary} (a--b), \textsc{QuestA} maintains higher policy entropy and achieves higher reward than vanilla GRPO, yet performs worse at test time. This gap indicates a severe training--inference discrepancy: \textit{the model is optimized under hint-conditioned rollouts during training but must reason autonomously when hints are absent at inference.} The issue is amplified in multilingual scenarios, where reasoning capabilities vary significantly across languages. For low-resource languages, rewarding trajectories are harder to reach without hints, incentivizing reliance on hint-conditioned shortcuts over autonomous multilingual reasoning. 
\begin{tcolorbox}[findingbox, title=Finding 2]
Increased response length does not necessarily indicate stronger reasoning ability and may exacerbate the repeat curse.
\end{tcolorbox} 
Figure~\ref{fig:prelimiary} (c) shows that \textsc{QuestA} substantially increases response length while also sharply increasing the repetition score, indicating that the added length is dominated by repetitive generation rather than reflecting richer reasoning patterns~\citep{gandhi2025cognitive}.
This provides further evidence of the training–inference discrepancy: \textit{once multilingual hints are removed, the model struggles to produce successful trajectories autonomously and degenerates into repetitive patterns}.
These results suggest that narrowing the training--inference discrepancy is crucial for transferring multilingual capabilities learned during training to test time.

\section{Methodology}
\begin{figure*}[!t]
    \centering
    \input{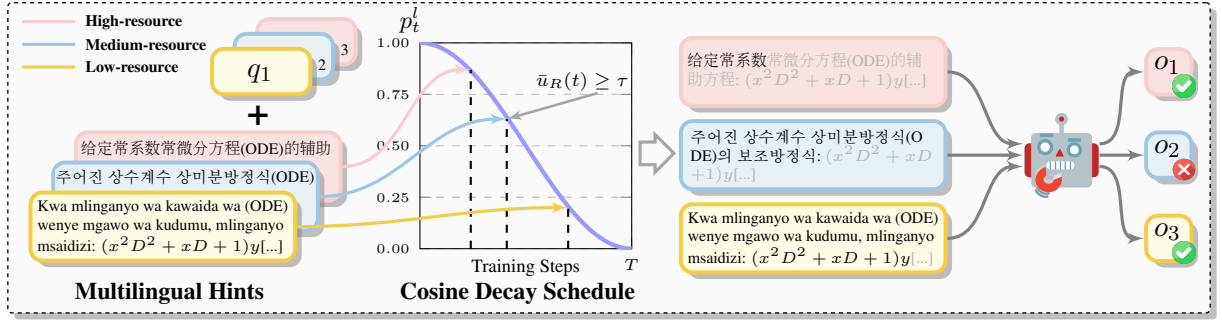}
    \caption{An overview of \ours: our method mitigates reward sparsity by incorporating multilingual hints to guide the model in generating correct multilingual reasoning, combined with a cosine annealing decay schedule and a language-adaptive switch that adjusts hint injection based on each language's learning difficulty.}
    \label{fig:method}
\end{figure*}

In this section, we present \ours, a language-adaptive hint-guided RL framework for multilingual reasoning, as shown in Figure~\ref{fig:method}. \ours consists of two key components: (1) \textbf{Scheduled Multilingual Hint Decay}, which augments training prompts with language-conditioned multilingual reasoning traces and progressively reduces hint exposure to encourage autonomous reasoning; and (2) \textbf{Language-adaptive Switch}, which sets language-specific decay horizons to match their varying learning difficulty.
\subsection{Scheduled Multilingual Hint Decay}

Given a question $q$ in language $l$, we assume access to a multilingual reasoning trace $h=(h_1,\ldots,h_L)$ produced by a teacher model in the same language, where $L$ is the trace length. At training step $t$, we inject a prefix of $h$ with length:
\begin{equation}
k_t^{l} = \left\lfloor p_t^{l}\,L \right\rfloor,
\label{eq:ktl}
\end{equation}
and construct the hint-conditioned prompt:
\begin{equation}
q_t^{l}=
\begin{cases}
q \oplus (h_1,\ldots,h_{k_t^{l}}), & t \le T,\\
q, & t > T,
\end{cases}
\label{eq:qstar}
\end{equation}
where $p_t^{l}\in[0,1]$ denotes the hint ratio at step $t$, and $T$ denotes the step after which hint injection is turned off.
By introducing hint guidance to bootstrap exploration, the probability of sampling successful trajectories in non-English settings is effectively increased, thereby mitigating the reward sparsity issue during the early training phase.

\paragraph{Cosine decay schedule.} To avoid hint dependence and narrow the training--inference gap, for $t\le T$, we instantiate $p_t^{l}$ using a cosine schedule:
\begin{equation}
\begin{aligned}
p_{t}^{l} = \begin{cases}
\frac{1}{2}\left(1 + \cos\left(\pi\frac{t}{T}\right)\right), & t \leq T, \\
0, & t > T.
\end{cases}
\end{aligned}
\label{eq:hint-strategy}
\end{equation}
This schedule starts with full guidance $(p_0^{l}=1)$ and smoothly anneals to zero, yielding a gradual transition from hint-conditioned rollouts to autonomous multilingual reasoning.

\subsection{Language-adaptive Switch}
Due to the substantial gaps in multilingual reasoning ability across languages, a single global switch step is suboptimal. 
To this end, we introduce a language-adaptive switching strategy that tailors learning horizons to specific language difficulties.
Specifically, we partition languages into resource groups $\mathcal{R}\in\{\text{high},\text{mid},\text{low}\}$.
For each rollout step $t$, we compute an effective-update rate $u_{\mathcal{R}}(t)$, defined as the fraction of instances in the group batch $\mathcal B_{\mathcal R}(t)$ whose rollout group contains at least one trajectory with positive advantage:

\begin{equation}
\begin{aligned}
u_{\mathcal R}(t)
&=\frac{1}{\left|\mathcal B_{\mathcal R}(t)\right|}
\sum_{x\in \mathcal B_{\mathcal R}(t)}
\mathbb{I}\Big[\exists\, i\in\{1,\ldots,G\}\\
&\qquad\text{s.t.}\ A_i(x,t)>0\Big],
\end{aligned}
\label{eq:effective_update_rate}
\end{equation}
where $\mathbb{I}$ is the indicator function.

To reduce variance, we adopt an exponential moving average with $\alpha=0.5$ to control the smoothing strength:
\begin{equation}
\bar u_{\mathcal{R}}(t)=\alpha\,\bar u_{\mathcal{R}}(t-1)+(1-\alpha)\,u_{\mathcal{R}}(t).
\label{eq:ema}
\end{equation}
A high $\bar u_{\mathcal{R}}(t)$ indicates that successful trajectories are consistently reachable, serving as an effective signal to remove scaffolding.
\paragraph{Switch criterion.} We switch a resource group $\mathcal{R}$ to the zero-hint regime once $\bar u_{\mathcal{R}}(t)$ exceeds a global threshold $\tau$ and define the switch step as:
\begin{equation}
T_{\mathcal R}=\min\{t \mid \bar u_{\mathcal R}(t)\ge \tau\}.
\label{eq:switch}
\end{equation}
As a result, low-resource groups retain hints longer to alleviate reward sparsity, while high-resource groups switch earlier to autonomous reasoning, reducing hint dependence.

\subsection{Policy Optimization via GRPO}
We optimize the policy model $\pi_{\theta}$ using GRPO~\citep{shao2024deepseekmath}. At each step, we sample a group of $G$ outputs $\{{o_i}\}_{i=1}^{G}\sim \pi_{\theta_{\text{old}}}(\cdot\mid q_t^{l})$ from the current policy, obtain rewards ${r_i}$, and compute standardized group-wise advantages:
\begin{equation}
A_i=
\frac{r_i-\mathrm{mean}(\{r_1,\ldots,r_G\})}
{\mathrm{std}(\{r_1,\ldots,r_G\})}.
\label{eq:adv}
\end{equation}
GRPO then updates $\pi_{\theta}$ by maximizing the clipped surrogate objective with KL regularization:
\begin{equation}
\small
\begin{aligned}
\mathcal{J}(\theta)
&= \mathbb{E}\Bigg[
\frac{1}{G}\sum_{i=1}^{G}
\min\Big(
\rho_i A_i,\,
\mathrm{clip}(\rho_i,1-\varepsilon,1+\varepsilon)A_i
\Big)
\Bigg],
\end{aligned}
\end{equation}
where $\rho_{i} = \frac{\pi_\theta(o_{i} \mid q_t^{l})}{\pi_{\theta_{\mathrm{old}}}(o_{i} \mid q_t^{l})}$. We omit the KL divergence term following the common practice as $\beta = 0$.
\paragraph{Reward.}
We use a binary, outcome-style reward that jointly enforces (i) \textit{language consistency}, (ii) \textit{output format}, and (iii) \textit{answer correctness}. 
Given a model output $o$ with a reasoning trace $o_t$ and a final answer $o_a$, we define:
(1) $R_{\text{lc}}(o)=1$ if both $o_t$ and $o_a$ are in the same language\footnote{Following~\citet{wang2025polymath}, we use the \texttt{langdetect} library to identify the language.} as the input question, and $0$ otherwise;
(2) $R_{\text{format}}(o)=1$ if $o$ contains the thinking tag \texttt{<think>}...\texttt{</think>} and provides the final answer within \verb|\boxed{}|, and $0$ otherwise;
(3) $R_{\text{acc}}(o)=1$ if the extracted final answer matches the ground truth under a rule-based verifier, and $0$ otherwise.
The overall reward is the conjunction:
\begin{equation}
\small
\label{reward}
R(o)=\mathbb{I}\big[R_{\text{lc}}(o)=1 \wedge R_{\text{format}}(o)=1 \wedge R_{\text{acc}}(o)=1\big].
\end{equation}

\section{Experimental Setups}
\subsection{Evaluation Datasets}
To assess the effectiveness of \ours, our main experiments are conducted on two challenging multilingual mathematical reasoning benchmarks. Detailed dataset statistics are provided Appendix~\ref{sec:Details_MMATH_PolyMath}
\paragraph{MMATH~\citep{luo-etal-2025-mmath}} comprises problems in ten languages, translated from the AIME\footnote{\url{https://maa.org/maa-invitational-competitions/}}, CNMO\footnote{\url{https://www.cms.org.cn/}}, and MATH-500~\citep{2024MATHCobbe} using GPT-4o-mini~\citep{2023-ChatGPT}.
\paragraph{PolyMath~\citep{wang2025polymath}} spans eighteen languages and four easy-to-hard difficulty levels, providing a comprehensive assessment of multilingual reasoning capabilities.

\subsection{Evaluation Metrics}
Following~\citet{wang2025polymath}, our evaluation primarily focuses on 
: \textit{language consistency}, \textit{accuracy}, and their conjunction. Detailed definitions are provided in Appendix~\ref{sec:Details_Details_Metrics}.
\paragraph{Language Consistency Ratio (LCR).}
LCR measures whether the model responds in the same language as the input question, computed based on the language-consistency criterion in Equation~\ref{reward}.

\paragraph{Accuracy (Acc).}
We compute accuracy by comparing the extracted final answer with ground-truth, regardless of response language~\citep{luo-etal-2025-mmath}.

\paragraph{Language Consistency \& Accuracy (LC\&Acc).}
LC\&Acc is our primary metric. It counts an output as correct only if the final answer is correct and the entire response is language-consistent with the input~\citep{wang2025polymath}.

\begin{table*}[t]
    \centering
    \small
    \definecolor{upgray}{RGB}{128, 128, 128}
\resizebox{\linewidth}{!}{%
\begin{tabular}{lccccccccccccc}
\toprule
\multirow{2}{*}{\textbf{Method}} & \multicolumn{7}{c}{\textbf{In-Domain Languages}} & \multicolumn{5}{c}{\textbf{Out-of-Domain Languages}} & \multirow{2}{*}{\textbf{ALL-Avg.}} \\
\cmidrule(lr){2-8} \cmidrule(lr){9-13}
 & \textbf{Ar} & \textbf{Th} & \textbf{Fr} & \textbf{Ja} & \textbf{Zh} & \textbf{En} & \textbf{Avg.} & \textbf{Vi} & \textbf{Ko} & \textbf{Pt} & \textbf{Es} & \textbf{Avg.} & \\
\midrule

\rowcolor{gray!30}
\textit{\textbf{Qwen2.5-3B-Instruct}} & \phantom{0}3.0 & \phantom{0}0.9 & \phantom{0}0.7 & \phantom{0}2.1 & 13.7 & 23.0 & \phantom{0}7.2 & \phantom{0}5.8 & \phantom{0}0.9 & \phantom{0}0.4 & \phantom{0}5.1 & \phantom{0}3.1 & \phantom{0}5.6 \\
& \textcolor{upgray} {13.1} & \textcolor{upgray}{11.0} & \textcolor{upgray}{16.7} & \textcolor{upgray}{13.7} & \textcolor{upgray}{15.1} & \textcolor{upgray}{23.0} & \textcolor{upgray}{15.4} & \textcolor{upgray}{15.9} & \textcolor{upgray}{13.0} &\textcolor{upgray}{18.5} & \textcolor{upgray}{16.6} & \textcolor{upgray}{16.0} & \textcolor{upgray}{15.7} \\
\hspace{2mm}+ LCP & 13.7 & 16.6 & 16.4 & 12.4 & 14.3 & 21.3 & 15.8 & 13.1 & 10.9 & 15.8 & 13.3 & 13.3 & 14.8 \\
\hspace{2mm}+ DIT & 12.5 & 17.6 & 17.0 & 12.3 & 15.7 & \textbf{23.2} & 16.4 & 14.9 & 12.3 & 17.7 & 12.4 & 14.3 & 15.6 \\
\hspace{2mm}+ QRT & 12.4 & 18.5 & 17.2 & 13.2 & 16.6 & 22.3 & 16.7 & 14.0 & 12.2 & 16.7 & 10.9 & 13.5 & 15.4 \\
\cdashline{1-14}
\hspace{2mm}+ M-SFT & 12.8 & 18.9 & 18.5 & 13.1 & 15.3 & 22.4 & 16.8 & 13.3 & 11.9 & 16.4 & 12.1 & 13.4 & 15.5 \\
\hspace{2mm}+ Vanilla GRPO & \phantom{0}2.5 & \phantom{0}4.2 & \phantom{0}9.0 & \phantom{0}6.9 & \textbf{17.2} & 22.4 & 10.4 & \phantom{0}4.6 & \phantom{0}1.1 & \phantom{0}1.1 & \phantom{0}1.3 & \phantom{0}9.5 & \phantom{0}7.3 \\
\hspace{2mm}+ LC-GRPO & 14.3 & 17.7 & 19.4 & 14.6 & 14.9 & 22.9 & 17.3 & 17.3 & 13.7 & \textbf{18.2} & 15.1 & 16.1 & 16.8 \\
\hspace{2mm}+ M-Thinker & 12.8 & 17.5 & 19.6 & 11.8 & 13.8 & 22.4 & 16.3 & 14.6 & 12.1 & 17.1 & 11.1 & 13.7 & 15.3 \\
\hspace{2mm}+ mGRPO & \textbf{15.0} & 15.3 & 18.2 & 12.2 & 14.1 & 19.9 & 15.8 & 16.3 & 14.5 & 14.9 & \textbf{16.7} & 15.6 & 15.7 \\
\headercolorSLAM
\hspace{2mm}+ \ours & 14.5 & \textbf{19.5} & \textbf{20.7} & \textbf{15.4} & 16.9 & 23.1 & \textbf{18.3} & \textbf{19.6} & \textbf{14.8} & 18.1 & 14.6 & \textbf{16.8} & \textbf{17.7} \\

\midrule

\rowcolor{gray!30}
\textit{\textbf{Qwen2.5-7B-Instruct}} & \phantom{0}0.3 & \phantom{0}0.5 & \phantom{0}0.2 & \phantom{0}3.6 & 21.0 & 28.2 & \phantom{0}9.0 & \phantom{0}1.3 & \phantom{0}1.8 & \phantom{0}0.7 & \phantom{0}2.1 & \phantom{0}1.5 & \phantom{0}6.0 \\
& \textcolor{upgray}{20.3} & \textcolor{upgray}{21.5} & \textcolor{upgray}{24.1} & \textcolor{upgray}{25.4} & \textcolor{upgray}{23.8} & \textcolor{upgray}{28.2} & \textcolor{upgray}{23.9} & \textcolor{upgray}{20.8} & \textcolor{upgray}{24.6} & \textcolor{upgray}{24.1} & \textcolor{upgray}{25.1} & \textcolor{upgray}{23.6} &\textcolor{upgray}{23.8} \\
\hspace{2mm}+ LCP & 18.0 & 16.1 & 23.2 & 18.6 & 19.5 & 29.6 & 20.8 & 19.8 & 19.8 & 23.4 & 22.9 & 21.5 & 21.1 \\
\hspace{2mm}+ DIT & 18.3 & 16.8 & 19.2 & 19.8 & 19.4 & 29.0 & 20.4 & 18.4 & 13.3 & 19.8 & 21.4 & 18.2 & 19.5 \\
\hspace{2mm}+ QRT & 17.7 & 16.1 & 20.3 & 19.8 & 23.1 & 27.7 & 20.8 & 22.4 & 14.3 & 16.2 & 23.6 & 19.1 & 20.1 \\
\cdashline{1-14}
\hspace{2mm}+ M-SFT & 12.9 & 15.2 & 12.6 & 19.8 & 19.4 & 32.4 & 18.7 & 12.0 & 16.3 & 16.3 & 17.1 & 15.4 & 17.4 \\
\hspace{2mm}+ Vanilla GRPO & \phantom{0}0.0 & \phantom{0}0.7 & \phantom{0}0.0 & \phantom{0}0.1 & \phantom{0}0.7 & 30.9 & \phantom{0}5.4 & \phantom{0}0.2 & \phantom{0}0.0 & \phantom{0}0.0 & \phantom{0}0.0 & \phantom{0}0.1 & \phantom{0}3.3 \\
\hspace{2mm}+ LC-GRPO & 24.3 & 23.6 & 28.2 & 25.0 & 24.7 & 30.2 & 26.0 & 25.2 & 24.9 & 29.2 & 27.8 & 26.8 & 26.3 \\
\hspace{2mm}+ M-Thinker & 21.4 & 20.7 & 22.7 & 24.5 & \textbf{25.7} & 28.6 & 23.9 & 24.6 & 22.8 & 26.7 & 26.7 & 25.2 & 24.4 \\
\hspace{2mm}+ mGRPO & 24.7 & 20.5 & 27.8 & 26.5 & 20.1 & 31.0 & 25.1 & 25.9 & \textbf{25.0} & 29.9 & 28.6 & 27.3 & 26.0 \\
\headercolorSLAM
\hspace{2mm}+ \ours & \textbf{26.3} & \textbf{28.5} & \textbf{31.1} & \textbf{28.0} & 22.5 & \textbf{32.1} & \textbf{28.1} & \textbf{30.1} & 24.5 & \textbf{32.2}& \textbf{31.2} & \textbf{29.5} & \textbf{28.6} \\

\bottomrule
\end{tabular}%
}

  \caption{The LC\&Acc (\%) on MMATH test sets. “Avg.” denotes the average performance within each split, and “ALL-Avg.” denotes the overall average across all languages. The highest score among systems of the same size is highlighted in \textbf{bold}. And \textcolor{upgray}{gray-colored} text indicates the accuracy (\%) without considering language consistency.} 
  \label{table:mmath-LC-Acc-maintable}
\end{table*}

\subsection{Baselines}
We compare \ours against two categories of baselines: \textit{prompting-based} and \textit{training-based}. See Appendix~\ref{sec:all_baselines} for implementation details.
\subsubsection{Prompting-based Methods}
\textbf{Language-Constraint Prompting (LCP)}~\citep{wang2025polymath}, \textbf{Discourse-Initiated Thinking (DIT)}~\citep{luo-etal-2025-mmath}, and \textbf{Question-Restatement Thinking (QRT)}~\citep{luo-etal-2025-mmath} simply prompt LLMs to generate language-constraint reasoning responses. Detailed implementation of these methods refers to Appendix~\ref{sec:prompting-based-methods}.

\subsubsection{Training-based Methods}
\paragraph{Multilingual Supervised Fine-tuning (M-SFT)} fine-tunes models on constructed multilingual mathematical data to improve performance.

\paragraph{Vanilla GRPO~\citep{guo2025deepseek}} applies the standard GRPO algorithm with format and accuracy rewards to perform RL training on multilingual mathematical data.

\paragraph{Language-Consistency GRPO (LC-GRPO)} extends GRPO algorithm with a language-consistency reward, constraining the model to respond in the language of questions.

\paragraph{M-Thinker~\citep{zhang2025think}} extends GRPO algorithm by evaluating cross-lingual thinking alignment with an LLM-as-judge and adding a language-consistency reward, thereby enhancing the multilingual reasoning alignment.

\paragraph{mGRPO~\citep{mGRPO}} encourages the model to sample multilingual trajectories within each group via prompting, mitigating drift toward English in the reasoning traces.

\subsection{Experimental Details}
We conduct experiments on the multilingual DeepMath-103K dataset, selecting ten languages in PolyMath as in-domain and reserving the remaining languages for out-of-domain evaluation. Subsequently, for each in-domain language, we sample 0.3K instances for cold-start training and then sample an additional 3K instances for RL training from the remaining data. To verify the effectiveness of our method, we conduct experiments on Qwen2.5-3B/7B/32B-Instruct~\citep{yang2025qwen3} and Llama3.1-8B-Instruct~\citep{llama-series}. For more details about training, refer to Appendix~\ref{sec:Experimental_Details}.

\begin{table*}[t]
    \centering
    \definecolor{upgray}{RGB}{128, 128, 128}
\resizebox{\linewidth}{!}{%
\begin{tabular}{lcccccccccccccccccccc}
\toprule
\multirow{2}{*}{\textbf{Method}} & \multicolumn{11}{c}{\textbf{In-Domain Languages}} & \multicolumn{8}{c}{\textbf{Out-of-Domain Languages}} & \multirow{2}{*}{\textbf{ALL-Avg.}} \\
\cmidrule(lr){2-12} \cmidrule(lr){13-20}
 & \textbf{Ar} & \textbf{Bn} & \textbf{Th} & \textbf{Sw} & \textbf{Ja} & \textbf{Zh} & \textbf{De} & \textbf{Fr} & \textbf{Ru} & \textbf{En} & \textbf{Avg.} & \textbf{Te} & \textbf{Ko} & \textbf{Vi} & \textbf{It} & \textbf{Id} & \textbf{Pt} & \textbf{Es} & \textbf{Avg.} & \\
\midrule

\rowcolor{gray!30}
\textit{\textbf{Qwen2.5-3B-Instruct}} &
\phantom{0}8.3 & \phantom{0}4.5 & \phantom{0}7.5 & \phantom{0}1.3 & \phantom{0}8.5 & \phantom{0}7.9 & \phantom{0}9.2 & \phantom{0}9.6 & \phantom{0}8.7 & 11.4 & \phantom{0}7.7 &
\phantom{0}1.0 & \phantom{0}8.4 & \phantom{0}9.9 & \phantom{0}9.5 & \phantom{0}9.6 & 11.4 & 11.2 & \phantom{0}8.7 & \phantom{0}8.1 \\
&
\textcolor{upgray}{\phantom{0}8.4} & \textcolor{upgray}{\phantom{0}4.5} & \textcolor{upgray}{\phantom{0}7.6} & \textcolor{upgray}{\phantom{0}1.3} & \textcolor{upgray}{\phantom{0}8.6} & \textcolor{upgray}{\phantom{0}9.4} & \textcolor{upgray}{\phantom{0}9.3} & \textcolor{upgray}{\phantom{0}9.6} & \textcolor{upgray}{\phantom{0}8.9} & \textcolor{upgray}{11.4} & \textcolor{upgray}{\phantom{0}7.9} &
\textcolor{upgray}{\phantom{0}1.0} & \textcolor{upgray}{\phantom{0}8.4} & \textcolor{upgray}{\phantom{0}9.9} & \textcolor{upgray}{10.0} & \textcolor{upgray}{\phantom{0}9.7} & \textcolor{upgray}{11.8} & \textcolor{upgray}{11.4} & \textcolor{upgray}{\phantom{0}8.9} & \textcolor{upgray}{\phantom{0}8.3} \\
\hspace{2mm}+ LCP &
\phantom{0}8.2 & \phantom{0}3.6 & \phantom{0}8.4 & \phantom{0}\textbf{3.4} & \phantom{0}6.8 & 10.2 & \phantom{0}9.9 & \phantom{0}9.2 & \phantom{0}8.8 & \textbf{14.1} & \phantom{0}8.3 &
\phantom{0}2.4 & \phantom{0}7.4 & \phantom{0}9.2 & \phantom{0}9.5 & \phantom{0}9.8 & 10.9 & \phantom{0}9.9 & \phantom{0}8.4 & \phantom{0}8.3 \\
\hspace{2mm}+ DIT &
\phantom{0}7.6 & \phantom{0}5.3 & \phantom{0}6.4 & \phantom{0}3.1 & \phantom{0}7.4 & \phantom{0}9.2 & \textbf{11.7} & \phantom{0}9.9 & 10.3 & 11.7 & \phantom{0}8.3 &
\phantom{0}1.0 & \phantom{0}8.2 & \phantom{0}8.8 & \phantom{0}9.5 & \phantom{0}9.2 & 10.2 & 10.2 & \phantom{0}8.2 & \phantom{0}8.2 \\
\hspace{2mm}+ QRT &
\textbf{10.8} & \phantom{0}3.2 & \phantom{0}6.3 & \phantom{0}1.5 & \phantom{0}9.2 & \phantom{0}6.9 & \phantom{0}9.5 & 10.9 & 10.8 & 12.7 & \phantom{0}8.2 &
\phantom{0}1.5 & \phantom{0}7.5 & 11.4 & 10.6 & 10.5 & \phantom{0}9.4 & \phantom{0}9.7 & \phantom{0}8.7 & \phantom{0}8.4 \\
\cdashline{1-21}
\hspace{2mm}+ M-SFT &
\phantom{0}2.7 & \phantom{0}2.8 & \phantom{0}4.3 & \phantom{0}0.9 & \phantom{0}2.7 & \phantom{0}4.6 & \phantom{0}3.0 & \phantom{0}2.9 & \phantom{0}5.9 & \phantom{0}4.5 & \phantom{0}3.4 &
\phantom{0}0.4 & \phantom{0}5.0 & \phantom{0}5.4 & \phantom{0}3.1 & \phantom{0}2.2 & \phantom{0}3.7 & \phantom{0}5.7 & \phantom{0}3.6 & \phantom{0}3.5 \\
\hspace{2mm}+ Vanilla GRPO &
\phantom{0}9.1 & \phantom{0}0.4 & \phantom{0}8.1 & \phantom{0}1.5 & \phantom{0}8.5 & 10.7 & 10.2 & 11.1 & 10.6 & 12.7 & \phantom{0}8.3 &
\phantom{0}2.6 & \phantom{0}8.5 & \phantom{0}9.7 & \phantom{0}7.8 & \textbf{12.2} & 10.0 & \textbf{11.6} & \phantom{0}8.9 & \phantom{0}8.6 \\
\hspace{2mm}+ LC-GRPO &
\phantom{0}9.9 & \phantom{0}6.6 & 10.5 & \phantom{0}3.0 & \phantom{0}9.7 & 10.8 & \phantom{0}9.9 & 11.7 & 10.2 & 12.4 & \phantom{0}9.5 &
\phantom{0}3.0 & \phantom{0}8.5 & 10.5 & 10.8 & 11.9 & 11.5 & 10.9 & \phantom{0}9.6 & \phantom{0}9.5 \\
\hspace{2mm}+ M-Thinker &
\phantom{0}9.1 & \phantom{0}4.8 & \phantom{0}6.5 & \phantom{0}2.3 & \phantom{0}7.3 & \phantom{0}8.7 & \phantom{0}9.7 & \phantom{0}8.3 & 10.7 & 12.5 & \phantom{0}8.0 &
\phantom{0}1.1 & \phantom{0}8.1 & \phantom{0}8.9 & \phantom{0}9.4 & \phantom{0}7.5 & \phantom{0}9.8 & \phantom{0}9.2 & \phantom{0}7.7 & \phantom{0}7.9 \\
\hspace{2mm}+ mGRPO &
\phantom{0}8.7 & \phantom{0}5.8 & \phantom{0}8.7 & \phantom{0}1.0 & \phantom{0}7.5 & \phantom{0}8.3 & 10.0 & \textbf{11.7} & \phantom{0}9.0 & 12.7 & \phantom{0}8.4 &
\phantom{0}3.1 & \phantom{0}7.8 & \phantom{0}8.6 & 10.1 & \phantom{0}8.0 & \phantom{0}8.4 & 10.1 & \phantom{0}8.0 & \phantom{0}8.2 \\
\headercolorSLAM
\hspace{2mm}+ \ours &
\phantom{0}8.9 & \textbf{\phantom{0}7.4} & \textbf{10.8} & \phantom{0}2.7 & \textbf{\phantom{0}9.9} & \textbf{13.5} & 10.9 & 11.1 & \textbf{11.6} & 13.3 & \textbf{10.0} &
\textbf{\phantom{0}4.5} & \textbf{\phantom{0}9.6} & \textbf{12.2} & \textbf{11.8} & 10.5 & \textbf{12.1} & 10.9 & \textbf{10.2} & \textbf{10.1} \\

\midrule

\rowcolor{gray!30}
\textit{\textbf{Qwen2.5-7B-Instruct}} &
12.1 & 11.8 & 12.7 & \phantom{0}2.2 & 13.8 & 15.3 & 15.3 & 15.2 & 14.8 & 17.3 & 13.1&\phantom{0}6.2 &
12.4 & 14.4 & 13.4 & \phantom{0}9.7 & 13.6 & 15.1 & 12.0 & 12.7  \\
&\textcolor{upgray}{12.5} & \textcolor{upgray}{12.0} & \textcolor{upgray}{13.3} & \textcolor{upgray}{\phantom{0}5.0} & \textcolor{upgray}{13.8} & \textcolor{upgray}{17.4} & \textcolor{upgray}{16.2} & \textcolor{upgray}{15.4} & \textcolor{upgray}{15.7} & \textcolor{upgray}{17.3} & \textcolor{upgray}{13.9} &
\textcolor{upgray}{\phantom{0}6.2} & \textcolor{upgray}{12.5} & \textcolor{upgray}{14.5} & \textcolor{upgray}{13.4} & \textcolor{upgray}{13.9} & \textcolor{upgray}{14.8} & \textcolor{upgray}{15.7} & \textcolor{upgray}{13.0} & \textcolor{upgray}{13.5} \\
\hspace{2mm}+ LCP &
\phantom{0}9.0 & \phantom{0}5.1 & \phantom{0}8.3 & \phantom{0}1.3 & \phantom{0}8.2 & \phantom{0}6.8 & \phantom{0}8.9 & \phantom{0}9.3 & 10.9 & 14.2 & \phantom{0}8.2 &
\phantom{0}1.0 & \phantom{0}9.4 & \phantom{0}9.2 & 10.0 & \phantom{0}7.9 & 10.5 & 10.7 & \phantom{0}8.4 & \phantom{0}8.3 \\
\hspace{2mm}+ DIT &
12.1 & 10.8 & 11.8 & \phantom{0}2.8 & 13.2 & 13.7 & 13.5 & 16.6 & 15.7 & 16.6 & 12.7 &
\phantom{0}6.6 & 11.6 & 11.9 & 16.8 & 13.5 & 14.4 & 14.3 & 12.7 & 12.7 \\
\hspace{2mm}+ QRT &
12.2 & \textbf{11.5} & 12.1 & \textbf{\phantom{0}4.9} & 14.5 & 13.4 & 14.8 & 16.1 & 15.0 & 17.5 & 13.2 &
\phantom{0}7.7 & 11.5 & 14.0 & 13.6 & 13.7 & 14.9 & 16.5 & 13.1 & 13.2 \\
\cdashline{1-21}
\hspace{2mm}+ M-SFT &
\phantom{0}5.1 & \phantom{0}2.8 & \phantom{0}6.7 & \phantom{0}1.3 & \phantom{0}4.8 & \phantom{0}3.7 & \phantom{0}4.1 & \phantom{0}5.1 & \phantom{0}5.3 & \phantom{0}6.1 & \phantom{0}4.5 &
\phantom{0}1.4 & \phantom{0}4.4 & \phantom{0}6.5 & \phantom{0}7.1 & \phantom{0}3.4 & \phantom{0}7.5 & \phantom{0}5.9 & \phantom{0}5.2 & \phantom{0}4.8 \\
\hspace{2mm}+ Vanilla GRPO &
13.1 & \phantom{0}4.5 & \phantom{0}1.4 & \phantom{0}2.1 & 13.1 & \textbf{15.8} & 16.6 & 15.9 & 15.3 & 17.4 & 11.5 &
\phantom{0}4.2 & 12.4 & 15.8 & 15.6 & 14.1 & 16.4 & 16.2 & 13.5 & 12.3 \\
\hspace{2mm}+ LC-GRPO &
14.3 & \phantom{0}9.0 & 11.2 & \phantom{0}2.5 & 13.7 & 13.4 & 17.2 & 17.0 & 18.3 & 17.6 & 13.4 &
\phantom{0}6.0 & 13.3 & 16.5 & 16.6 & 14.5 & 17.1 & 18.7 & 14.7 & 13.9 \\
\hspace{2mm}+ M-Thinker &
13.6 & \phantom{0}8.8 & \phantom{0}9.0 & \phantom{0}3.1 & 11.7 & 12.1 & 13.2 & 14.8 & 14.7 & 14.3 & 11.5 &
\textbf{\phantom{0}7.8} & 10.8 & 12.3 & 14.3 & 13.4 & 13.8 & 15.5 & 12.6 & 12.0 \\
\hspace{2mm}+ mGRPO &
14.9 & 10.6 & 11.5 & \phantom{0}2.7 & 12.7 & 13.1 & 13.8 & 15.4 & 16.4 & 15.3 & 12.6 &
\phantom{0}6.5 & 12.8 & 13.8 & 15.7 & 13.3 & 16.4 & 16.5 & 13.6 & 13.0 \\
\headercolorSLAM
\hspace{2mm}+ \ours &
\textbf{16.2} & 11.0 & \textbf{15.3} & \phantom{0}3.4 & \textbf{15.3} & 15.7 & \textbf{17.9} & \textbf{17.9} & \textbf{19.4} & \textbf{17.9} & \textbf{15.0} &
\phantom{0}7.6 & \textbf{16.1} & \textbf{17.2} & \textbf{20.5} & \textbf{17.4} & \textbf{17.7} & \textbf{19.0} & \textbf{16.5} & \textbf{15.6} \\

\bottomrule
\end{tabular}%
}

  \caption{The LC\&Acc (\%) on PolyMath test sets.}
  \label{table:polymath-LC-Acc-maintable}
\end{table*}

\section{Experimental Results}

\subsection{Main Results}
We present the main metric LC\&Acc on the MMATH and PolyMath benchmarks in Table~\ref{table:mmath-LC-Acc-maintable} and~\ref{table:polymath-LC-Acc-maintable}, respectively. Detailed results for LCR and Accuracy are provided in Appendix~\ref{sec:Experiment_of_Acc_and_LC}. The results of Qwen2.5-32B-Instruct and Llama3.1-8B-Instruct refer to Appendix~\ref{sec:qwen32B-llama}.

\paragraph{Current LLMs struggle to generate language-consistent reasoning traces while preserving accuracy.} 
As shown in Tables~\ref{table:mmath-LC-Acc-maintable} and~\ref{table:polymath-LC-Acc-maintable}, Qwen-family models achieve strong English accuracy, yet their LC\&Acc degrades substantially once the intermediate reasoning traces are required to remain in the input language. 
Moreover, existing approaches further reveal a consistent trade-off between language consistency and accuracy.  
In particular, on Qwen2.5-3B-Instruct, prompting-based methods can greatly increase language consistency (e.g., QRT boosts the MMATH LCR from 23.3\% to 85.1\%), but this gain comes with an 11.3\% drop in accuracy. 
Conversely, training-based approaches without explicit language constraints achieve notable accuracy improvements but reduce language consistency. Specifically, on PolyMath with Qwen2.5-7B-Instruct, vanilla GRPO yields a 9.0\% relative accuracy gain, while its LCR drops by 16.5\% relative to QRT.  
Overall, these results indicate that generating language-consistent reasoning traces with correct answers remains challenging.

\paragraph{\ours effectively improves multilingual reasoning ability without sacrificing language consistency.}
As shown in Tables~\ref{table:mmath-LC-Acc-maintable} and ~\ref{table:polymath-LC-Acc-maintable}, \ours consistently outperforms all competitive baselines on both MMATH and PolyMath in the in-domain setting. Across the four evaluation models, \ours improves average LC\&Acc by 24.1\% on MMATH and 18.7\% on PolyMath relative to LC-GRPO. Moreover, \ours achieves near-perfect language consistency as illustrated in Appendix~\ref{sec:Experiment_of_Acc_and_LC} (Figure~\ref{table:mmath-3B-LC}-\ref{table:polymath-7B-LC}). Remarkably, these gains are particularly pronounced in low-resource languages. For example, with Qwen2.5-7B-Instruct, \ours yields LC\&Acc gains of 39.0\% on Thai in MMATH and 24.6\% on Vietnamese in PolyMath over mGRPO. This indicates that \ours leverages fine-grained multilingual hints with explicit language-consistency supervision, mitigating the trade-off between accuracy and language consistency.

\paragraph{\ours exhibits strong out-of-distribution and cross-model generalization.}
\ours consistently improves LC\&Acc across diverse evaluation languages and outperforms all strong baselines on out-of-domain languages as shown in Tables~\ref{table:mmath-LC-Acc-maintable} and~\ref{table:polymath-LC-Acc-maintable}. Specifically, \ours achieves average LC\&Acc gains of 79.3\% on Llama3.1-8B-Instruct and 11.2\% on Qwen2.5-7B-Instruct over LC-GRPO across two benchmarks.
This superior generalization demonstrates that the multilingual reasoning ability learned by \ours can be effectively transferred to unseen languages. Additionally, we note consistent improvements across model families and scales. 
As shown in Tables~\ref{table:mmath-LC-Acc-llama} and~\ref{table:polymath-LC-Acc-llama}, \ours achieves average LC\&Acc gains over LC-GRPO of 41.1\% for Llama3.1-8B-Instruct and 4.9\% for Qwen2.5-32B-Instruct across two benchmarks. These consistent gains across languages and model variants demonstrate the robustness of \ours in guiding models to sample language-consistent reasoning trajectories.

\subsection{Ablation Study}
We conduct ablation studies to validate \ours, with additional ablations reported in Appendix~\ref{sec:additional_ablations}.
\begin{figure}[!t]
    \centering
\definecolor{upurple}{RGB}{155,89,182}
\definecolor{ublue}{RGB}{52,152,219}
\definecolor{ured}{RGB}{231,76,60}
\definecolor{udark}{RGB}{77,153,77}
\definecolor{ugreen}{RGB}{46,204,113}
\definecolor{upink}{HTML}{fcd4d4}
\definecolor{ucyan}{HTML}{e3eeff}
\definecolor{uedgecyan}{HTML}{6d97e0}
\definecolor{uedgepink}{HTML}{cc0000}
\definecolor{usemiblue}{HTML}{17becf}
\definecolor{uorange}{HTML}{ffcc00}

\tikzset{global scale/.style={
    scale=#1,
    every node/.append style={scale=#1}
  }
}

\begin{tikzpicture}
\pgfmathsetmacro{\T}{1000}       
\pgfmathsetmacro{\cparam}{\T/2}  
\pgfmathsetmacro{\kparam}{100}   
\pgfmathsetmacro{\lam}{6}        

\begin{axis}[
  name=right,                    
  global scale = 0.63,
  width=0.6\linewidth,
  height=.3\textwidth,
  scale only axis=true,
  at={(0.3em,1em)},             
  anchor=south west,
  xmin=0, xmax=\T,
  ymin=0, ymax=1,
  xlabel={Training Steps},
  ylabel={Decay Function Value},
  ylabel style={yshift=0.5em},
  tick label style={font=\scriptsize},
  grid=both,
  major grid style={dashed},
  xtick={0,200,400,600,800,1000},
  ytick={0,0.25,0.5,0.75,1.00},
  legend style={
    at={(1.08,1.2)},
    anchor=south,
    fill=none,
    legend columns=1,
    /tikz/every even column/.append style={column sep=5pt},
    draw=none,
    legend image post style={xscale=0.5},
    inner sep=1pt,
    legend cell align=left,
    font=\scriptsize
  },
  scaled y ticks=false,
yticklabel style={
  font=\scriptsize,
  /pgf/number format/fixed,
  /pgf/number format/fixed zerofill,
  /pgf/number format/precision=2
},
  domain=0:\T,
  samples=300,
]
  \addplot[ublue!80, thick] {exp(-\lam * x / \T)};
  \addlegendentry{Exponential decay}

  \addplot[uorange!60, thick] {1 - x / \T};
  \addlegendentry{Linear decay}

  \addplot[red!90, thick] {0.5 * (1 + cos(180 * x / \T))};
  \addlegendentry{Cosine annealing decay}
\end{axis}

\begin{axis}[
  name=left,                     
  global scale = 0.63,
  width=0.6\linewidth,
  height=.3\textwidth,
  scale only axis=true,
  at={(right.east)},            
  anchor=west,
  xshift=4.4em,                   
  ymajorgrids,
  grid style=dashed,
  ybar,
  xmin=0.8, xmax=1.7,
  xtick={1,1.5},
  xticklabels={MMATH,PolyMath},
  enlarge x limits=false,
  xtick align=inside,
  bar width=0.6em,
  nodes near coords,
  nodes near coords align={vertical},
  nodes near coords style={
    font=\tiny,
    scale=0.9,
    /pgf/number format/fixed,
    /pgf/number format/fixed zerofill,
    /pgf/number format/precision=1
  },
  ymin=6, ymax=30,
  ytick={6.0,10,14,18,22,26,30},
  yticklabel pos=left,
  yticklabel style={
  font=\tiny,
    /pgf/number format/fixed,
    /pgf/number format/fixed zerofill,
    /pgf/number format/precision=1
  },
  ylabel={LC\&Acc (\%)},
  ylabel style={font=\normalsize, yshift=-1.0em},
  ylabel shift=1.5em, 
  xlabel style={yshift=0.3em,align=center},
  legend style={
    draw=none,
    line width=1pt,
    at={(0.5,1.0)},
    anchor=south,
    cells={align=left}
  },
  axis on top=true,
]
  \addplot[fill=usemiblue!40,draw=usemiblue!60, area legend] coordinates {
    (1,21.1) (1.5,8.26)
  };

  \addplot[fill=ublue!40,draw=ublue!60, area legend, nodes near coords style={yshift=-1pt}] coordinates {
    (1,24.1) (1.5,14.3)
  };

  \addplot[fill=uorange!40, draw=uorange!60, area legend, nodes near coords style={yshift=2pt}] coordinates {
    (1,27.1) (1.5,14.6)
  };

  \addplot[fill=red!40, draw=red!60, area legend] coordinates {
    (1,28.6) (1.5,15.6)
  };
\end{axis}

\node[rectangle,draw=usemiblue!60,fill=usemiblue!40,
      inner sep=2pt,minimum height=0.4em,minimum width=0.8em,
      anchor=north west] (leg0)
  at ([xshift=11mm,yshift=-2mm]left.north west) {};

\node[rectangle,draw=ublue!60,fill=ublue!40,
      inner sep=2pt,minimum height=0.4em,minimum width=0.8em,
      anchor=north west] (leg1)
  at ([yshift=-0.25em]leg0.south west) {};

\node[rectangle,draw=uorange!60,fill=uorange!40,
      inner sep=2pt,minimum height=0.4em,minimum width=0.8em,
      anchor=north west] (leg2)
  at ([yshift=-0.25em]leg1.south west) {};

\node[rectangle,draw=red!60,fill=red!40,
      inner sep=2pt,minimum height=0.4em,minimum width=0.8em,
      anchor=north west] (leg3)
  at ([yshift=-0.25em]leg2.south west) {};

\node[anchor=west, scale=0.7] at ([xshift=0.35em]leg0.center) {\tiny Qwen2.5-7B-Instruct};
\node[anchor=west, scale=0.7] at ([xshift=0.35em]leg1.center) {\tiny Exponential decay};
\node[anchor=west, scale=0.7] at ([xshift=0.35em]leg2.center) {\tiny Linear decay};
\node[anchor=west, scale=0.7] at ([xshift=0.35em]leg3.center) {\tiny Cosine annealing decay};

\node[anchor=west, scale=0.7] at ([xshift=-10em,yshift=-7.6em]leg2.south west) {(a)};
\node[anchor=west, scale=0.7, align=left] at ([xshift=0em,yshift=-7.5em]leg2.south west)
{(b)};
\end{tikzpicture}
    \vspace{-1.5em}
    \caption{(a) Examples of decay schedules (b) The average performance on MMATH and PolyMath dataset.}
    \label{fig:decay-compare}
\end{figure}
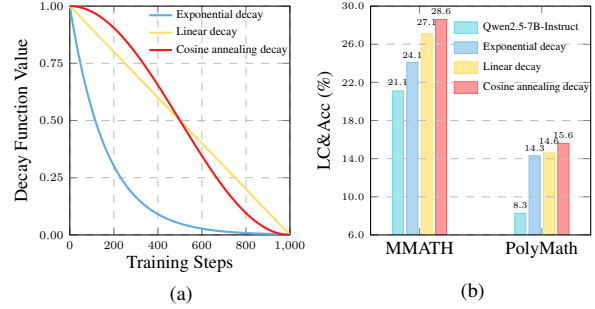
\paragraph{Effect of different sampling strategies.}

To evaluate the effectiveness of the cosine annealing sampling strategy, we conduct ablation studies on Qwen2.5-7B-Instruct by comparing it with exponential and linear decay. As shown in Figure~\ref{fig:decay-compare} (a), exponential decay reduces the hint length too quickly in the early stage, whereas linear decay applies a uniform schedule that ignores the model’s evolving reasoning abilities. In contrast, cosine annealing better matches the desired pattern by providing stronger guidance early and encouraging more autonomous exploration later, thereby achieving the best average LC\&Acc on both benchmarks in Figure~\ref{fig:decay-compare} (b). These results suggest that cosine annealing alleviates early reward sparsity while enabling a timely transition to independent exploration, thus improving consistency between training and inference.

\begin{figure}[t!]
    \centering
    \definecolor{c1}{HTML}{81BECE}
\definecolor{c2}{HTML}{378BA4}
\definecolor{ublue}{HTML}{5fa0d1}  
\definecolor{ured}{HTML}{a9c1f7}   
\definecolor{udpblue}{HTML}{0419fb}
\definecolor{usemiblue}{HTML}{17becf}
\raisebox{-10em}{%
    \begin{tikzpicture}
\path[use as bounding box] (-4em,5.5em) rectangle (32em,13em);
  \clip (-4em,5.5em) rectangle (32em,16.9em);
    \tiny{
    \begin{axis}[
      at={(0,13em)},
      legend entries={de2en},
      ymajorgrids,
      xmajorgrids,
      grid style=dashed,
      xbar,
      legend image code/.code={%
                    \draw[#1, draw=none] (0cm,-0.1cm) rectangle (0.6cm,0.1cm);
                }, 
      height=.29\textwidth,
      width=.45\textwidth,
      bar width=0.8em,
      xlabel={\scriptsize{LC\&Acc (\%)}}, 
      xtick={0,5,10,15,20,25,30},
      xmax=32, 
      xmin=0,
      symbolic y coords={{PLAST \\ w/o cold-start},{PLAST \\ w/o $R_{\text{lc}}$},{PLAST \\ w/o $p_{t}^{l}$},{PLAST}},
      yticklabels={{\ours},{w/o cold-start},{ w/o $R_{\text{lc}}$},{ w/o $p_{t}^{l}$}},
      yticklabel style={align=right,font=\tiny},
      ytick=data,
      nodes near coords,
      nodes near coords align={horizontal},
      legend style={at={(25em,9.3em)}},
legend reversed,
      enlarge y limits=0.15,xticklabel style={/pgf/number format/fixed,/pgf/number format/fixed zerofill,/pgf/number format/precision=1},
      enlarge x limits=0,xticklabel style={/pgf/number format/fixed,/pgf/number format/fixed zerofill,/pgf/number format/precision=1},
      nodes near coords style={
  /pgf/number format/fixed,
  /pgf/number format/fixed zerofill,
  /pgf/number format/precision=1
},
]
    \addplot[fill=ublue!40,draw=ublue,area legend] coordinates {
    (28.6,PLAST)
    (27.5,{PLAST \\ w/o cold-start})
    (3.1,{PLAST \\ w/o $R_{\text{lc}}$})
    (10.1,{PLAST \\ w/o $p_{t}^{l}$})
};
    \addplot[fill=ured!40,draw=ured,area legend] coordinates {
    (15.6,PLAST)
    (15.0,{PLAST \\ w/o cold-start})
    (9.1,{PLAST \\ w/o $R_{\text{lc}}$})
    (3.2,{PLAST \\ w/o $p_{t}^{l}$})
};
      \addlegendentry{MMATH\phantom{l}}
      \addlegendentry{PolyMath}

    \end{axis}

  }

    \end{tikzpicture}
}
    \vspace{0.5em}
    \caption{The average performance of training different components on MMATH and PolyMath datasets.}
    \label{fig:ablation-diff-module}
\end{figure}
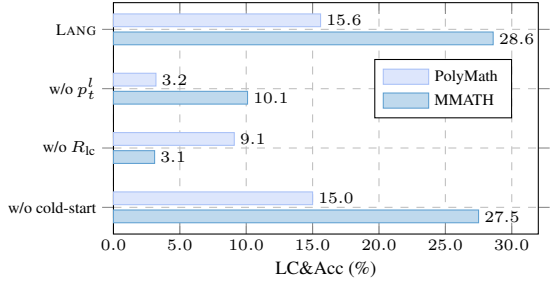
\paragraph{Effect of key ingredients in the \ours training framework.} 
To further assess the necessity of the language-consistency reward, the hint annealing phase, and cold-start training, we conduct ablation studies on Qwen2.5-7B-Instruct. 
As shown in Figure~\ref{fig:ablation-diff-module}, without the hint annealing phase $p_{t}^{l}$, maintaining full hints throughout training amplifies the training inference discrepancy and consequently degrades performance. 
Furthermore, removing the language-consistency reward $R_{\text{lc}}$ causes pronounced language drifting of the reasoning traces into English, undermining the goal of language-consistency reasoning. 
Additionally, removing the cold-start stage results in a 3.0\% average performance drop, indicating that cold-start training improves the initial policy’s compliance with instruction-specified output language and formatting constraints, thereby stabilizing and facilitating subsequent multilingual RL training.

\section{Further Analysis}
\begin{table}[t]
    \centering
    
\definecolor{upgreen}{RGB}{88, 166, 93}
\definecolor{downred}{RGB}{210,70,70}

\centering
\resizebox{0.48\textwidth}{!}{%
\begin{tabular}{l c c c l}
\toprule
\textbf{Method} & \textbf{MMLU-ProX} & \textbf{XWinograd} & \textbf{XStoryCloze} & \textbf{XCOPA} \\
\midrule
\textit{\textbf{Qwen2.5-7B-Instruct}} & 35.9\phantom{0000002} & 65.7\phantom{0000000} & 60.9\phantom{000001} & 60.7\phantom{0000000} \\
\hspace{2mm}+ Vanilla GRPO & 21.0\phantom{0000002} & 65.7\phantom{0000000} &57.5\phantom{000001} & 58.9\phantom{0000000} \\
\hspace{2mm}+ LC-GRPO & 39.9\phantom{0000002} & 62.2\phantom{0000000} &60.2\phantom{000001} & 60.5\phantom{0000000} \\
\hspace{2mm}+ \ours                & 41.0\,{\textcolor{upgreen}{$\uparrow$\,\textbf{14.2\%}}} & 79.9\,{\textcolor{upgreen}{$\uparrow$\,\textbf{21.6\%}}} & 63.5\,{\textcolor{upgreen}{$\uparrow$\,\textbf{4.3\%}}} & 62.9\,{\textcolor{upgreen}{$\uparrow$\,\textbf{3.6\%}}} \\
\bottomrule
\end{tabular}

}

  \caption{The average extract match and accuracy (\%) on non-mathematical benchmarks. \textcolor{upgreen}{\textbf{Arrows}} indicate the relative improvement over Qwen2.5-7B-Instruct model. }
  \label{table:4datasets}
\end{table}

\subsection{Scalability of \ours Beyond Multilingual Mathematical Reasoning}
To evaluate the scalability of \ours, we extend our method on MMLU-ProX~\citep{xuan2025mmlu}, XWinograd~\citep{muennighoff2022crosslingual,tikhonov2021heads}, XStoryCloze~\citep{DBLP:journals/corr/abs-2112-10668}, XCOPA~\citep{ponti-etal-2020-xcopa} benchmarks, which cover multilingual understanding and generation tasks across domains. We utilize lm-evaluation-harness\footnote{\url{https://github.com/EleutherAI/lm-evaluation-harness}} as our evaluation framework (Detail results are provided in Appendix~\ref{sec:Experiment_of_Xtasks}). Notably, as shown in Table~\ref{table:4datasets}, \ours achieves significant average performance improvements of 10.9\% over four benchmarks. 
The results demonstrate that \ours transfers effectively to diverse multilingual tasks, suggesting that multilingual guidance during RL training improves general multilingual capability beyond the target reasoning domain.

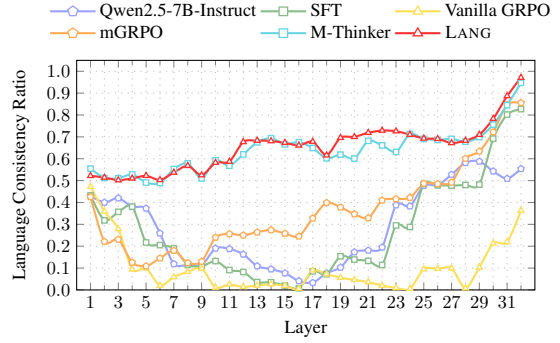
\begin{figure}[t!]
    \centering
    \definecolor{upurple}{RGB}{155,89,182}
\definecolor{ured}{RGB}{231,76,60}
\definecolor{udark}{RGB}{77,153,77}
\definecolor{udpdark}{HTML}{85827a}
\definecolor{usemidark}{HTML}{8c564b}
\definecolor{ublue}{RGB}{52,152,219}
\definecolor{udpblue}{HTML}{0419fb}
\definecolor{usemiblue}{HTML}{17becf}
\definecolor{uorange}{HTML}{ffcc00}
\definecolor{udporange}{HTML}{bcbd22}
\tikzset{global scale/.style={
    scale=#1,
    every node/.append style={scale=#1}
  }
}

\pgfplotsset{
    width=1.4\textwidth,
   height=.6\textheight,
   grid=major,
   major grid style={dotted},
   enlarge y limits={upper,value=0.05},
   legend style={
      fill,
      at={(0.50,16.5em)},
      legend columns=5,
      legend cell align=left,
      anchor=south
      },
   }
\begin{tikzpicture}

   \begin{axis}[
global scale = 0.63,
   at={(0em,16em)},
    legend entries={Qwen2.5-7B-Instruct,SFT,Vanilla GRPO,mGRPO,M-Thinker,\ours},
    height=.4\textwidth,
    width=.7\textwidth,
    ymajorgrids=true,
    xmajorgrids=true,
    legend style={draw=none,
    line width=1pt,
    at={(11.5em,13.5em)},
    legend cell align=left,
    legend columns=3
    column sep=3pt,
    yshift=-5pt,
    anchor=south
    },
    enlarge x limits={abs=1},
    xlabel=Layer,
    ylabel={Language Consistency Ratio},
    xlabel style={yshift=-0.5em},
    ylabel style={yshift=1em},
    restrict x to domain=0:33,
   xtick={1,3,5,7,9,11,13,15,17,19,21,23,25,27,29,31},
   minor xtick={2,4,6,8,10,12,14,16,18,20,22,24,26,28,30,32},  
   xminorgrids=true,                                              
   minor grid style={dotted},                                     
   ymin=0.0, ymax=1.0,
   ytick={0.0,0.1,0.2,0.3,0.4,0.5,0.6,0.7,0.8,0.9,1.0},
   yticklabel style={/pgf/number format/fixed,/pgf/number format/fixed zerofill,/pgf/number format/precision=1,rotate=0}
   ]
   \addplot[sharp plot,udpblue!40,smooth,thick,line width=0.5pt,mark=pentagon*,mark size=2pt,thick,mark options={fill=white,draw=udpblue!40,line width=0.5pt}] plot coordinates{
(1, 0.43) (2, 0.4) (3, 0.42) 
(4, 0.38) (5, 0.373) (6, 0.259) 
(7, 0.119) (8, 0.114) (9, 0.113) 
(10, 0.190) (11, 0.188) (12, 0.163) 
(13, 0.109) (14, 0.095) (15, 0.077) 
(16, 0.041) (17, 0.033) (18, 0.077) 
(19, 0.103) (20, 0.173) (21, 0.181) 
(22, 0.195) (23, 0.388) (24, 0.383) 
(25, 0.477) (26, 0.479) (27, 0.528) 
(28, 0.582) (29, 0.588) (30, 0.543) 
(31, 0.509) (32, 0.554) 
      };
   \addplot[sharp plot,udark!65,smooth,thick,line width=0.5pt,mark=square*,mark size=1.6pt,thick,mark options={fill=white,draw=udark!65,line width=0.5pt}] plot coordinates {
(1, 0.432) (2, 0.318) (3, 0.357) 
(4, 0.381) (5, 0.217) (6, 0.205) 
(7, 0.190) (8, 0.116) (9, 0.108) 
(10, 0.132) (11, 0.091) (12, 0.082) 
(13, 0.036) (14, 0.035) (15, 0.020) 
(16, 0.007) (17, 0.085) (18, 0.072) 
(19, 0.154) (20, 0.139) (21, 0.133) 
(22, 0.114) (23, 0.295) (24, 0.288) 
(25, 0.484) (26, 0.479) (27, 0.477) 
(28, 0.480) (29, 0.482) (30, 0.692) 
(31, 0.802) (32, 0.827)   
      };
    \addplot[sharp plot,uorange!65,smooth,thick,line width=0.5pt,mark=triangle*,mark size=2pt,thick,mark options={fill=white,draw=uorange!65,line width=0.5pt}] plot coordinates {
(1, 0.472) (2, 0.359) (3, 0.279) 
(4, 0.094) (5, 0.099) (6, 0.020) 
(7, 0.059) (8, 0.083) (9, 0.098) 
(10, 0.011) (11, 0.025) (12, 0.014) 
(13, 0.020) (14, 0.026) (15, 0.019) 
(16, 0.005) (17, 0.090) (18, 0.072) 
(19, 0.056) (20, 0.046) (21, 0.036) 
(22, 0.020) (23, 0.009) (24, 0.005) 
(25, 0.095) (26, 0.097) (27, 0.099) 
(28, 0.008) (29, 0.103) (30, 0.213) 
(31, 0.219) (32, 0.363) 
      };
   \addplot[sharp plot,orange!65,smooth,thick,line width=0.5pt,mark=pentagon*,mark size=2pt,thick,mark options={fill=white,draw=orange!65,line width=0.5pt}] plot coordinates{
(1, 0.426) (2, 0.221) (3, 0.231) 
(4, 0.126) (5, 0.109) (6, 0.145) 
(7, 0.180) (8, 0.124) (9, 0.130) 
(10, 0.240) (11, 0.256) (12, 0.250) 
(13, 0.264) (14, 0.274) (15, 0.258) 
(16, 0.245) (17, 0.328) (18, 0.399) 
(19, 0.378) (20, 0.347) (21, 0.329) 
(22, 0.410) (23, 0.416) (24, 0.422) 
(25, 0.487) (26, 0.485) (27, 0.492) 
(28, 0.601) (29, 0.634) (30, 0.722) 
(31, 0.847) (32, 0.856) 
      };
   \addplot[sharp plot,usemiblue!65,smooth,thick,line width=0.5pt,mark=square*,mark size=1.6pt,thick,mark options={fill=white,draw=usemiblue!65,line width=0.5pt}] plot coordinates {
(1, 0.555) (2, 0.512) (3, 0.511) 
(4, 0.528) (5, 0.491) (6, 0.489) 
(7, 0.553) (8, 0.579) (9, 0.510) 
(10, 0.593) (11, 0.568) (12, 0.620) 
(13, 0.675) (14, 0.694) (15, 0.666) 
(16, 0.676) (17, 0.649) (18, 0.602) 
(19, 0.619) (20, 0.601) (21, 0.682) 
(22, 0.661) (23, 0.631) (24, 0.713) 
(25, 0.691) (26, 0.686) (27, 0.691) 
(28, 0.678) (29, 0.699) (30, 0.756) 
(31, 0.845) (32, 0.947) 
      };
    \addplot[sharp plot,red!80,smooth,thick,line width=0.5pt,mark=triangle*,mark size=2pt,thick,mark options={fill=white,draw=red!80,line width=0.5pt}] plot coordinates {
(1, 0.522) (2, 0.514) (3, 0.502) 
(4, 0.510) (5, 0.522) (6, 0.503) 
(7, 0.537) (8, 0.568) (9, 0.525) 
(10, 0.582) (11, 0.587) (12, 0.677) 
(13, 0.683) (14, 0.681) (15, 0.673) 
(16, 0.661) (17, 0.678) (18, 0.616) 
(19, 0.696) (20, 0.700) (21, 0.719) 
(22, 0.729) (23, 0.726) (24, 0.710) 
(25, 0.693) (26, 0.691) (27, 0.672) 
(28, 0.683) (29, 0.710) (30, 0.783) 
(31, 0.887) (32, 0.971) 
      };

   \end{axis}

\end{tikzpicture}
    \caption{The layer-wise comparison of the language-consistency rate of intermediate decoded outputs with the question language across different training methods.}
    \label{fig:logit-analysis-across-layers}
\end{figure}

\subsection{Impact of Translation Quality on Multilingual Performance}
We further extend our method by constructing the multilingual training data with different translation models to examine the impact of translation quality on the resulting gains in multilingual performance. As shown in Table~\ref{table:diff-transtor}, the model’s multilingual gains remain stable across training data constructed with different translation models. This indicates that \ours does not rely heavily on translation quality, but instead helps the model internalize multilingual reasoning patterns, leading to strong robustness.

\begin{table}[t]
    \centering
    
\definecolor{upgreen}{RGB}{88, 166, 93}
\definecolor{downred}{RGB}{210,70,70}

\centering
\resizebox{0.48\textwidth}{!}{%
\begin{tabular}{lcc}
\toprule
\textbf{Translation Model} & \textbf{MMATH} & \textbf{PolyMath} \\
\midrule
GPT-4o-mini~\citep{2023-ChatGPT} & 28.6 & 15.6 \\
Claude-Opus-4.5~\citep{anthropic2025claudeopus45} & 27.9 &  15.0\\
NLLB-200-3.3B~\citep{NLLB}   & 25.1 & 13.8 \\
\bottomrule
\end{tabular}

}

  \caption{The average LC\&Acc (\%) on MMATH and PolyMath under different translation models for multilingual training data construction, using Qwen2.5-7B-Instruct as the backbone. }
  \label{table:diff-transtor}
\end{table}

\begin{figure*}[!t]
    \centering
    \input{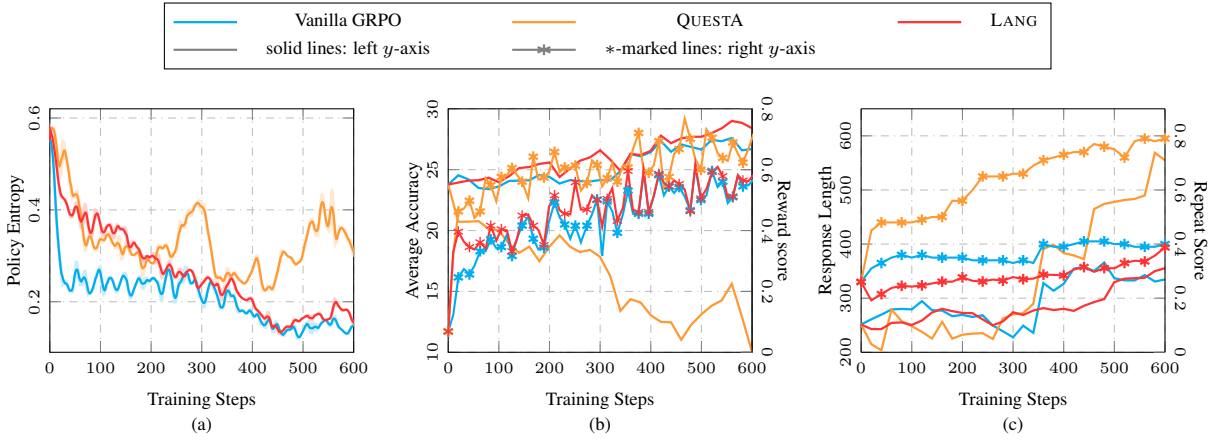}
    \vspace{-1.5em}
    \caption{The comparison of Vanilla GRPO and \textsc{QuestA} during RL training. \textcolor{cyan}{Blue curves} denote Vanilla GRPO, \textcolor{orange!80}{orange curves} denote \textsc{QuestA}, and \textcolor{red!80}{red curves} denote \ours. In the middle and right panels, solid lines correspond to the left $y$-axis, and $*$-marked lines correspond to the right $y$-axis.}
    \label{fig:prelimiary2}
\end{figure*}

\subsection{Layer-wise Analysis of Language-Consistent Reasoning}
To examine the extent to which the model maintains language-consistent reasoning across layers, rather than reasoning in English in intermediate layers and translating into the target language only at upper layers, which leads to inconsistency in multilingual reasoning and hinders the model's ability to maintain coherent and accurate reasoning across languages~\citep{zhao2024how}. 
We decode intermediate hidden states into layer-wise outputs with the logit lens~\citep{nostalgebraist2020logitlens} and measure language consistency with the input question on MMATH. 
As shown in Figure~\ref{fig:logit-analysis-across-layers}, \ours maintains consistently high language-consistency rates across layers, thereby mitigating disrupted reasoning continuity induced by cross-layer language switching. This result confirms that our method encourages the model to internalize language-consistent reasoning throughout its intermediate representations, rather than relying on superficial language conversion at the final output layer.

\subsection{Revisiting Findings after Training with \ours}
To facilitate comparisons, we compare the metrics discussed in Section~\ref{Observations} before and after training with \ours. As shown in Figure~\ref{fig:prelimiary2}, our method attains a higher average reward score during RL training compared to Vanilla GRPO. Furthermore, the higher reward reliably translated into better test-time multilingual reasoning performance. Additionally, \ours increases response length without inducing repetitive generation. This highlights that \ours narrows the training--inference discrepancy, enabling the multilingual capabilities learned during training to transfer effectively to test time.

\subsection{Impact of Teacher Model on Multilingual Performance}
\label{appendix:teacher_model}
To verify that our method is not dependent on a specific teacher model, we further extend our experiments by replacing DeepSeek-R1 with GPT-4o-mini 
to generate reasoning-trace hints for constructing the multilingual training data. As shown in Table~\ref{tab:teacher_mmath} and Table~\ref{tab:teacher_polymath}, the model's multilingual gains remain stable across training data constructed with different teacher models. Specifically, after switching to GPT-4o-mini, \textsc{Lang} still brings improvements of +22.5 and +2.8 points on MMATH and PolyMath respectively, compared to the Qwen2.5-7B-Instruct baseline. This indicates that \textsc{Lang} does not rely heavily on the choice of teacher reasoning model, but instead helps the student model internalize multilingual reasoning patterns from diverse reasoning traces, confirming that our approach is \emph{teacher-agnostic}.

\section{Conclusion}
In this work, we propose \ours, a language-adaptive hint-guided RL framework for multilingual reasoning.
\ours bootstraps exploration with language-conditioned multilingual hints in early training, then progressively reduces hint exposure to mitigate the training-inference discrepancy. A key innovation is our language-adaptive switch, which adjusts the learning pace based on difficulty, specifically preserving guidance for languages that need it while accelerating independence for those that do not. Experiments on MMATH and PolyMath across two LLM families and multiple scales show that \ours substantially improves multilingual reasoning while preserving language consistency. Ablations further verify the necessity of both progressive hint decay and the language-adaptive switching mechanism.

\section*{Limitations}
\label{sec:limitations}
Our work presents several limitations worth noting. First, for constructing multilingual hints, we use the widely adopted DeepSeek-R1 as the sole distillation source. Future work will involve extending our method by using data distilled from additional teacher models to more comprehensively evaluate the generalizability of \ours. Second, while our method achieves substantial improvements in accuracy while maintaining language consistency between input and output across both in-domain and out-of-domain test sets across model families and scales, the degrees of performance across different languages result in performance trade-offs. We hypothesize that these trade-offs may arise from imbalances in multilingual training data during the pre-training stage. However, our work starts from the perspective of RL training to enhance the multilingual ablities of LLMs. In future work, we will explore data-centric strategies for improving multilingual capabilities, with a particular focus on data selection and data augmentation.

\section*{Ethics Statement}
This work does not require ethical considerations. All the data used in this paper is sourced from open-source materials. Throughout the experimental process, all data and models were strictly utilized following their intended purposes and respective licenses.  
Additionally, this paper may contain offensive text related to the case study. We have all referenced them elliptically and will not present the complete harmful content within the paper.
\label{sec:Ethics}

\section*{Acknowledgements}
In particular, we sincerely thank Lei Huang for his insightful and constructive suggestions. His generous encouragement has been instrumental in sustaining my efforts to complete this work. This work was supported in part by the National Natural
Science Foundation of China (Nos. U24A20334 and
62276056), the Yunnan Fundamental Research Projects
(No.202401BC070021), the Yunnan Science and Technology
Major Project (No. 202502AD080014), the Fundamental
Research Funds for the Central Universities (Nos.
N25BSS054 and N25BSS094), and the Program of Introducing
Talents of Discipline to Universities, Plan 111
(No.B16009).
\bibliography{custom,acl2026}

@article{bengio2015scheduled,
  title={Scheduled sampling for sequence prediction with recurrent neural networks},
  author={Bengio, Samy and Vinyals, Oriol and Jaitly, Navdeep and Shazeer, Noam},
  journal={Advances in neural information processing systems},
  volume={28},
  year={2015}
}

@inproceedings{qian2021glancing,
  title={Glancing transformer for non-autoregressive neural machine translation},
  author={Qian, Lihua and Zhou, Hao and Bao, Yu and Wang, Mingxuan and Qiu, Lin and Zhang, Weinan and Yu, Yong and Li, Lei},
  booktitle={Proceedings of the 59th Annual Meeting of the Association for Computational Linguistics and the 11th International Joint Conference on Natural Language Processing (Volume 1: Long Papers)},
  pages={1993--2003},
  year={2021}
}

@inproceedings{zhang-etal-2019-bridging,
    title = "Bridging the Gap between Training and Inference for Neural Machine Translation",
    author = "Zhang, Wen  and
      Feng, Yang  and
      Meng, Fandong  and
      You, Di  and
      Liu, Qun",
    editor = "Korhonen, Anna  and
      Traum, David  and
      M{\`a}rquez, Llu{\'i}s",
    booktitle = "Proceedings of the 57th Annual Meeting of the Association for Computational Linguistics",
    month = jul,
    year = "2019",
    address = "Florence, Italy",
    publisher = "Association for Computational Linguistics",
    url = "https://aclanthology.org/P19-1426/",
    doi = "10.18653/v1/P19-1426",
    pages = "4334--4343",
}

@article{huang2026bootstrapping,
  title={Bootstrapping Exploration with Group-Level Natural Language Feedback in Reinforcement Learning},
  author={Huang, Lei and Cheng, Xiang and Zhao, Chenxiao and Shen, Guobin and Yang, Junjie and Feng, Xiaocheng and Gu, Yuxuan and Yu, Xing and Qin, Bing},
  journal={arXiv preprint arXiv:2603.04597},
  year={2026}
}

@inproceedings{fan-etal-2025-language,
    title = "Language-Specific Layer Matters: Efficient Multilingual Enhancement for Large Vision-Language Models",
    author = "Fan, Yuchun  and
      Wang, Yilin  and
      Mu, Yongyu  and
      Huang, Lei  and
      Li, Bei  and
      Feng, Xiaocheng  and
      Xiao, Tong  and
      Zhu, JingBo",
    editor = "Christodoulopoulos, Christos  and
      Chakraborty, Tanmoy  and
      Rose, Carolyn  and
      Peng, Violet",
    booktitle = "Findings of the Association for Computational Linguistics: EMNLP 2025",
    month = nov,
    year = "2025",
    address = "Suzhou, China",
    publisher = "Association for Computational Linguistics",
    url = "https://aclanthology.org/2025.findings-emnlp.666/",
    doi = "10.18653/v1/2025.findings-emnlp.666",
    pages = "12473--12500",
    ISBN = "979-8-89176-335-7"
}

@inproceedings{fan-etal-2025-slam,
    title = "{SLAM}: Towards Efficient Multilingual Reasoning via Selective Language Alignment",
    author = "Fan, Yuchun  and
      Mu, Yongyu  and
      Wang, YiLin  and
      Huang, Lei  and
      Ruan, Junhao  and
      Li, Bei  and
      Xiao, Tong  and
      Huang, Shujian  and
      Feng, Xiaocheng  and
      Zhu, Jingbo",
    editor = "Rambow, Owen  and
      Wanner, Leo  and
      Apidianaki, Marianna  and
      Al-Khalifa, Hend  and
      Eugenio, Barbara Di  and
      Schockaert, Steven",
    booktitle = "Proceedings of the 31st International Conference on Computational Linguistics",
    month = jan,
    year = "2025",
    address = "Abu Dhabi, UAE",
    publisher = "Association for Computational Linguistics",
    url = "https://aclanthology.org/2025.coling-main.637/",
    pages = "9499--9515"
}

@article{wang2026dapt,
  title={DaPT: A Dual-Path Framework for Multilingual Multi-hop Question Answering},
  author={Wang, Yilin and Fan, Yuchun and Li, Jiaoyang and Zhu, Ziming and Mu, Yongyu and He, Qiaozhi and Xiao, Tong and Zhu, Jingbo},
  journal={arXiv preprint arXiv:2603.19097},
  year={2026}
}

@inproceedings{joshi-etal-2020-state,
    title = "The State and Fate of Linguistic Diversity and Inclusion in the {NLP} World",
    author = "Joshi, Pratik  and
      Santy, Sebastin  and
      Budhiraja, Amar  and
      Bali, Kalika  and
      Choudhury, Monojit",
    editor = "Jurafsky, Dan  and
      Chai, Joyce  and
      Schluter, Natalie  and
      Tetreault, Joel",
    booktitle = "Proceedings of the 58th Annual Meeting of the Association for Computational Linguistics",
    month = jul,
    year = "2020",
    address = "Online",
    publisher = "Association for Computational Linguistics",
    url = "https://aclanthology.org/2020.acl-main.560/",
    doi = "10.18653/v1/2020.acl-main.560",
    pages = "6282--6293"
}

@inproceedings{
zhao2024large,
title={Large Language Model is not a (Multilingual) Compositional Relation Reasoner},
author={Jinman Zhao and Xueyan Zhang},
booktitle={First Conference on Language Modeling},
year={2024},
url={https://openreview.net/forum?id=wLQ3I0F1oj}
}

@article{Zhao_Min_Wu_Li_Sun_Cai_Wang_Chen_Penn_2026, title={Beyond Step Pruning: Information Theory Based Step-level Optimization for Self-Refining Large Language Models}, volume={40}, url={https://ojs.aaai.org/index.php/AAAI/article/view/40798}, DOI={10.1609/aaai.v40i41.40798}, number={41}, journal={Proceedings of the AAAI Conference on Artificial Intelligence}, author={Zhao, Jinman and Min, Erxue and Wu, Hui and Li, Ziheng and Sun, Zexu and Cai, Hengyi and Wang, Shuaiqiang and Chen, Xu and Penn, Gerald}, year={2026}, month={Mar.}, pages={34941-34949} }

@article{DeekSeek-R1,
  author       = {DeepSeek{-}AI},
  title        = {DeepSeek-R1: Incentivizing Reasoning Capability in LLMs via Reinforcement
                  Learning},
  journal      = {CoRR},
  volume       = {abs/2501.12948},
  year         = {2025},
  url          = {https://doi.org/10.48550/arXiv.2501.12948},
  doi          = {10.48550/ARXIV.2501.12948},
  eprinttype    = {arXiv},
  eprint       = {2501.12948},
  bibsource    = {dblp computer science bibliography, https://dblp.org}
}

@article{wang2025polymath,
  title={Polymath: Evaluating mathematical reasoning in multilingual contexts},
  author={Wang, Yiming and Zhang, Pei and Tang, Jialong and Wei, Haoran and Yang, Baosong and Wang, Rui and Sun, Chenshu and Sun, Feitong and Zhang, Jiran and Wu, Junxuan and others},
  journal={arXiv preprint arXiv:2504.18428},
  year={2025}
}

@inproceedings{luo-etal-2025-mmath,
    title = "{MMATH}: A Multilingual Benchmark for Mathematical Reasoning",
    author = "Luo, Wenyang  and
      Zhao, Wayne Xin  and
      Sha, Jing  and
      Wang, Shijin  and
      Wen, Ji-Rong",
    editor = "Christodoulopoulos, Christos  and
      Chakraborty, Tanmoy  and
      Rose, Carolyn  and
      Peng, Violet",
    booktitle = "Findings of the Association for Computational Linguistics: EMNLP 2025",
    month = nov,
    year = "2025",
    address = "Suzhou, China",
    publisher = "Association for Computational Linguistics",
    url = "https://aclanthology.org/2025.findings-emnlp.598/",
    doi = "10.18653/v1/2025.findings-emnlp.598",
    pages = "11187--11202",
    ISBN = "979-8-89176-335-7"
}

@inproceedings{qi-etal-2025-models,
    title = "When Models Reason in Your Language: Controlling Thinking Language Comes at the Cost of Accuracy",
    author = "Qi, Jirui  and
      Chen, Shan  and
      Xiong, Zidi  and
      Fern{\'a}ndez, Raquel  and
      Bitterman, Danielle  and
      Bisazza, Arianna",
    editor = "Christodoulopoulos, Christos  and
      Chakraborty, Tanmoy  and
      Rose, Carolyn  and
      Peng, Violet",
    booktitle = "Findings of the Association for Computational Linguistics: EMNLP 2025",
    month = nov,
    year = "2025",
    address = "Suzhou, China",
    publisher = "Association for Computational Linguistics",
    url = "https://aclanthology.org/2025.findings-emnlp.1103/",
    doi = "10.18653/v1/2025.findings-emnlp.1103",
    pages = "20279--20296",
    ISBN = "979-8-89176-335-7"
}

@article{hwang2025learn,
  title={Learn globally, speak locally: Bridging the gaps in multilingual reasoning},
  author={Hwang, Jaedong and Tanmay, Kumar and Lee, Seok-Jin and Agrawal, Ayush and Palangi, Hamid and Ayush, Kumar and Fiete, Ila and Liang, Paul Pu},
  journal={arXiv preprint arXiv:2507.05418},
  year={2025}
}

@article{park2025cross,
  title={Cross-lingual collapse: How language-centric foundation models shape reasoning in large language models},
  author={Park, Cheonbok and Kim, Jeonghoon and Lee, Joosung and Bae, Sanghwan and Choo, Jaegul and Yoo, Kang Min},
  journal={arXiv preprint arXiv:2506.05850},
  year={2025}
}

@article{zhang2025think,
  title={Think Natively: Unlocking Multilingual Reasoning with Consistency-Enhanced Reinforcement Learning},
  author={Zhang, Xue and Liang, Yunlong and Meng, Fandong and Zhang, Songming and Huang, Kaiyu and Chen, Yufeng and Xu, Jinan and Zhou, Jie},
  journal={arXiv preprint arXiv:2510.07300},
  year={2025}
}

@article{yang2025parallel,
  title={Parallel Scaling Law: Unveiling Reasoning Generalization through A Cross-Linguistic Perspective},
  author={Yang, Wen and Wu, Junhong and Li, Chong and Zong, Chengqing and Zhang, Jiajun},
  journal={arXiv preprint arXiv:2510.02272},
  year={2025}
}

@article{shi2025efficient,
  title={Efficient reinforcement finetuning via adaptive curriculum learning},
  author={Shi, Taiwei and Wu, Yiyang and Song, Linxin and Zhou, Tianyi and Zhao, Jieyu},
  journal={arXiv preprint arXiv:2504.05520},
  year={2025}
}

@article{wang2025hint,
  title={HINT: Helping Ineffective Rollouts Navigate Towards Effectiveness},
  author={Wang, Xinyi and Han, Jinyi and Jiang, Zishang and Li, Tingyun and Liang, Jiaqing and Jiang, Sihang and Dai, Zhaoqian and Ma, Shuguang and Yu, Fei and Xiao, Yanghua},
  journal={arXiv preprint arXiv:2510.09388},
  year={2025}
}

@article{li2025questa,
  title={Questa: Expanding reasoning capacity in llms via question augmentation},
  author={Li, Jiazheng and Lin, Hongzhou and Lu, Hong and Wen, Kaiyue and Yang, Zaiwen and Gao, Jiaxuan and Wu, Yi and Zhang, Jingzhao},
  journal={arXiv preprint arXiv:2507.13266},
  year={2025}
}

@inproceedings{yao-etal-2025-understanding,
    title = "Understanding the Repeat Curse in Large Language Models from a Feature Perspective",
    author = "Yao, Junchi  and
      Yang, Shu  and
      Xu, Jianhua  and
      Hu, Lijie  and
      Li, Mengdi  and
      Wang, Di",
    editor = "Che, Wanxiang  and
      Nabende, Joyce  and
      Shutova, Ekaterina  and
      Pilehvar, Mohammad Taher",
    booktitle = "Findings of the Association for Computational Linguistics: ACL 2025",
    month = jul,
    year = "2025",
    address = "Vienna, Austria",
    publisher = "Association for Computational Linguistics",
    url = "https://aclanthology.org/2025.findings-acl.406/",
    doi = "10.18653/v1/2025.findings-acl.406",
    pages = "7787--7815",
    ISBN = "979-8-89176-256-5"
}

@article{he2025deepmath,
  title={Deepmath-103k: A large-scale, challenging, decontaminated, and verifiable mathematical dataset for advancing reasoning},
  author={He, Zhiwei and Liang, Tian and Xu, Jiahao and Liu, Qiuzhi and Chen, Xingyu and Wang, Yue and Song, Linfeng and Yu, Dian and Liang, Zhenwen and Wang, Wenxuan and others},
  journal={arXiv preprint arXiv:2504.11456},
  year={2025}
}

@article{shao2024deepseekmath,
  title={Deepseekmath: Pushing the limits of mathematical reasoning in open language models},
  author={Shao, Zhihong and Wang, Peiyi and Zhu, Qihao and Xu, Runxin and Song, Junxiao and Bi, Xiao and Zhang, Haowei and Zhang, Mingchuan and Li, YK and Wu, Yang and others},
  journal={arXiv preprint arXiv:2402.03300},
  year={2024}
}

@article{guo2025deepseek,
  title={Deepseek-r1: Incentivizing reasoning capability in llms via reinforcement learning},
  author={Guo, Daya and Yang, Dejian and Zhang, Haowei and Song, Junxiao and Zhang, Ruoyu and Xu, Runxin and Zhu, Qihao and Ma, Shirong and Wang, Peiyi and Bi, Xiao and others},
  journal={arXiv preprint arXiv:2501.12948},
  year={2025}
}

@misc{mGRPO,
  title        = {{mGRPO}: Unlocking LLM Reasoning through Multilingual Thinking},
  author       = {{mGRPO}},
  howpublished = {OpenReview},
  year         = {2025},
  month        = oct,
  url          = {https://openreview.net/forum?id=QtfALPluAj},
  note         = {OpenReview submission},
}

@inproceedings{Shi2024mCOT,
  author       = {Freda Shi and
                  Mirac Suzgun and
                  Markus Freitag and
                  Xuezhi Wang and
                  Suraj Srivats and
                  Soroush Vosoughi and
                  Hyung Won Chung and
                  Yi Tay and
                  Sebastian Ruder and
                  Denny Zhou and
                  Dipanjan Das and
                  Jason Wei},
  title        = {Language models are multilingual chain-of-thought reasoners},
  booktitle    = {The Eleventh International Conference on Learning Representations,
                  {ICLR} 2023, Kigali, Rwanda, May 1-5, 2023},
  publisher    = {OpenReview.net},
  year         = {2023},
  url          = {https://openreview.net/forum?id=fR3wGCk-IXp},
  bibsource    = {dblp computer science bibliography, https://dblp.org}
}

@inproceedings{2024MATHCobbe,
  author       = {Hunter Lightman and
                  Vineet Kosaraju and
                  Yuri Burda and
                  Harrison Edwards and
                  Bowen Baker and
                  Teddy Lee and
                  Jan Leike and
                  John Schulman and
                  Ilya Sutskever and
                  Karl Cobbe},
  title        = {Let's Verify Step by Step},
  booktitle    = {The Twelfth International Conference on Learning Representations,
                  {ICLR} 2024, Vienna, Austria, May 7-11, 2024},
  publisher    = {OpenReview.net},
  year         = {2024},
  url          = {https://openreview.net/forum?id=v8L0pN6EOi},
  bibsource    = {dblp computer science bibliography, https://dblp.org}
}

@article{2023-ChatGPT,
  author       = {OpenAI},
  title        = {{GPT-4} Technical Report},
  journal      = {CoRR},
  volume       = {abs/2303.08774},
  year         = {2023},
  url          = {https://doi.org/10.48550/arXiv.2303.08774},
  doi          = {10.48550/ARXIV.2303.08774},
  eprinttype    = {arXiv},
  eprint       = {2303.08774},
  bibsource    = {dblp computer science bibliography, https://dblp.org}
}

@misc{openai2024o3o4,
  author       = {OpenAI},
  title        = {Openai o3 and o4-mini system card.},
  month        = {April},
  year         = {2025},
  url = {https://openai.com/index/o3-o4-mini-system-card/}
}

@article{zhang2025stephint,
  title={StepHint: Multi-level Stepwise Hints Enhance Reinforcement Learning to Reason},
  author={Zhang, Kaiyi and Lv, Ang and Li, Jinpeng and Wang, Yongbo and Wang, Feng and Hu, Haoyuan and Yan, Rui},
  journal={arXiv preprint arXiv:2507.02841},
  year={2025}
}

@article{UFT,
author       = {Liu, Mingyang and Farina, Gabriele and Ozdaglar, Asuman},
title        = {UFT: Unifying Supervised and Reinforcement Fine-Tuning},
journal      = {arXiv preprint arXiv:2505.16984},
year         = {2025}
}

@article{yang2025qwen3,
  title={Qwen3 technical report},
  author={Yang, An and Li, Anfeng and Yang, Baosong and Zhang, Beichen and Hui, Binyuan and Zheng, Bo and Yu, Bowen and Gao, Chang and Huang, Chengen and Lv, Chenxu and others},
  journal={arXiv preprint arXiv:2505.09388},
  year={2025}
}

@inproceedings{ponti-etal-2020-xcopa,
    title = "{XCOPA}: A Multilingual Dataset for Causal Commonsense Reasoning",
    author = "Ponti, Edoardo Maria  and
      Glava{\v{s}}, Goran  and
      Majewska, Olga  and
      Liu, Qianchu  and
      Vuli{\'c}, Ivan  and
      Korhonen, Anna",
    editor = "Webber, Bonnie  and
      Cohn, Trevor  and
      He, Yulan  and
      Liu, Yang",
    booktitle = "Proceedings of the 2020 Conference on Empirical Methods in Natural Language Processing (EMNLP)",
    month = nov,
    year = "2020",
    address = "Online",
    publisher = "Association for Computational Linguistics",
    url = "https://aclanthology.org/2020.emnlp-main.185/",
    doi = "10.18653/v1/2020.emnlp-main.185",
    pages = "2362--2376"
}

@article{DBLP:journals/corr/abs-2112-10668,
  author    = {Xi Victoria Lin and
               Todor Mihaylov and
               Mikel Artetxe and
               Tianlu Wang and
               Shuohui Chen and
               Daniel Simig and
               Myle Ott and
               Naman Goyal and
               Shruti Bhosale and
               Jingfei Du and
               Ramakanth Pasunuru and
               Sam Shleifer and
               Punit Singh Koura and
               Vishrav Chaudhary and
               Brian O'Horo and
               Jeff Wang and
               Luke Zettlemoyer and
               Zornitsa Kozareva and
               Mona T. Diab and
               Veselin Stoyanov and
               Xian Li},
  title     = {Few-shot Learning with Multilingual Language Models},
  journal   = {CoRR},
  volume    = {abs/2112.10668},
  year      = {2021},
  url       = {https://arxiv.org/abs/2112.10668},
  eprinttype = {arXiv},
  eprint    = {2112.10668},
  bibsource = {dblp computer science bibliography, https://dblp.org}
}

@misc{2023opencompass,
    title={OpenCompass: A Universal Evaluation Platform for Foundation Models},
    author={OpenCompass},
    howpublished = {\url{https://github.com/open-compass/opencompass}},
    year={2023}
}

@inproceedings{deepspeed2,
  author       = {Samyam Rajbhandari and
                  Jeff Rasley and
                  Olatunji Ruwase and
                  Yuxiong He},
  editor       = {Christine Cuicchi and
                  Irene Qualters and
                  William T. Kramer},
  title        = {ZeRO: memory optimizations toward training trillion parameter models},
  booktitle    = {Proceedings of the International Conference for High Performance Computing,
                  Networking, Storage and Analysis, {SC} 2020, Virtual Event / Atlanta,
                  Georgia, USA, November 9-19, 2020},
  pages        = {20},
  publisher    = {{IEEE/ACM}},
  year         = {2020},
  url          = {https://doi.org/10.1109/SC41405.2020.00024},
  doi          = {10.1109/SC41405.2020.00024},
  bibsource    = {dblp computer science bibliography, https://dblp.org}
}

@article{xuan2025mmlu,
  title={Mmlu-prox: A multilingual benchmark for advanced large language model evaluation},
  author={Weihao Xuan and Rui Yang and Heli Qi and Qingcheng Zeng and Yunze Xiao and Aosong Feng and Dairui Liu and Yun Xing and Junjue Wang and Fan Gao and Jinghui Lu and Yuang Jiang and Huitao Li and Xin Li and Kunyu Yu and Ruihai Dong and Shangding Gu and Yuekang Li and Xiaofei Xie and Felix Juefei-Xu and Foutse Khomh and Osamu Yoshie and Qingyu Chen and Douglas Teodoro and Nan Liu and Randy Goebel and Lei Ma and Edison Marrese-Taylor and Shijian Lu and Yusuke Iwasawa and Yutaka Matsuo and Irene Li},
  journal={arXiv preprint arXiv:2503.10497},
  year={2025}
}

@misc{muennighoff2022crosslingual,
      title={Crosslingual Generalization through Multitask Finetuning},
      author={Niklas Muennighoff and Thomas Wang and Lintang Sutawika and Adam Roberts and Stella Biderman and Teven Le Scao and M Saiful Bari and Sheng Shen and Zheng-Xin Yong and Hailey Schoelkopf and Xiangru Tang and Dragomir Radev and Alham Fikri Aji and Khalid Almubarak and Samuel Albanie and Zaid Alyafeai and Albert Webson and Edward Raff and Colin Raffel},
      year={2022},
      eprint={2211.01786},
      archivePrefix={arXiv},
      primaryClass={cs.CL}
}

@misc{tikhonov2021heads,
    title={It's All in the Heads: Using Attention Heads as a Baseline for Cross-Lingual Transfer in Commonsense Reasoning},
    author={Alexey Tikhonov and Max Ryabinin},
    year={2021},
    eprint={2106.12066},
    archivePrefix={arXiv},
    primaryClass={cs.CL}
}

@misc{nostalgebraist2020logitlens,
  author = {Nostalgebraist},
  title = {Interpreting GPT: The Logit Lens},
  howpublished = {LessWrong},
  year = {2020},
  url = {https://www.lesswrong.com/posts/AcKRB8wDpdaN6v6ru/interpreting-gpt-the-logit-lens}
}

@article{zhao2024how,
  author       = {Yiran Zhao and
                  Wenxuan Zhang and
                  Guizhen Chen and
                  Kenji Kawaguchi and
                  Lidong Bing},
  title        = {How do Large Language Models Handle Multilingualism?},
  journal      = {CoRR},
  volume       = {abs/2402.18815},
  year         = {2024},
  url          = {https://doi.org/10.48550/arXiv.2402.18815},
  doi          = {10.48550/ARXIV.2402.18815},
  eprinttype    = {arXiv},
  eprint       = {2402.18815},
  bibsource    = {dblp computer science bibliography, https://dblp.org}
}

@inproceedings{zhang-etal-2025-p,
    title = "{P}-{MME}val: A Parallel Multilingual Multitask Benchmark for Consistent Evaluation of {LLM}s",
    author = "Zhang, Yidan  and
      Wan, Yu  and
      Deng, Boyi  and
      Yang, Baosong  and
      Wei, Hao-Ran  and
      Huang, Fei  and
      Yu, Bowen  and
      Liu, Dayiheng  and
      Lin, Junyang  and
      Huang, Fei  and
      Zhou, Jingren",
    editor = "Christodoulopoulos, Christos  and
      Chakraborty, Tanmoy  and
      Rose, Carolyn  and
      Peng, Violet",
    booktitle = "Proceedings of the 2025 Conference on Empirical Methods in Natural Language Processing",
    month = nov,
    year = "2025",
    address = "Suzhou, China",
    publisher = "Association for Computational Linguistics",
    url = "https://aclanthology.org/2025.emnlp-main.242/",
    doi = "10.18653/v1/2025.emnlp-main.242",
    pages = "4809--4836",
    ISBN = "979-8-89176-332-6"
}

@article{DBLP:journals/corr/abs-2501-14249,
  author       = {Long Phan and
                  Alice Gatti and
                  Ziwen Han and
                  Nathaniel Li and
                  Josephina Hu and
                  Hugh Zhang and
                  Sean Shi and
                  Michael Choi and
                  Anish Agrawal and
                  Arnav Chopra and
                  Adam Khoja and
                  Ryan Kim and
                  Jason Hausenloy and
                  Oliver Zhang and
                  Mantas Mazeika and
                  Daron Anderson and
                  Tung Nguyen and
                  Mobeen Mahmood and
                  Fiona Feng and
                  Steven Y. Feng and
                  Haoran Zhao and
                  Michael Yu and
                  Varun Gangal and
                  Chelsea Zou and
                  Zihan Wang and
                  Jessica P. Wang and
                  Pawan Kumar and
                  Oleksandr Pokutnyi and
                  Robert Gerbicz and
                  Serguei Popov and
                  John{-}Clark Levin and
                  Mstyslav Kazakov and
                  Johannes Schmitt and
                  Geoff Galgon and
                  Alvaro Sanchez and
                  Yongki Lee and
                  Will Yeadon and
                  Scott Sauers and
                  Marc Roth and
                  Chidozie Agu and
                  S{\o}ren Riis and
                  Fabian Giska and
                  Saiteja Utpala and
                  Zachary Giboney and
                  Gashaw M. Goshu and
                  Joan of Arc Xavier and
                  Sarah{-}Jane Crowson and
                  Mohinder Maheshbhai Naiya and
                  Noah Burns and
                  Lennart Finke and
                  Zerui Cheng and
                  Hyunwoo Park and
                  Francesco Fournier{-}Facio and
                  John Wydallis and
                  Mark Nandor and
                  Ankit Singh and
                  Tim Gehrunger and
                  Jiaqi Cai and
                  Ben McCarty and
                  Darling Duclosel and
                  Jungbae Nam and
                  Jennifer Zampese and
                  Ryan G. Hoerr and
                  Aras Bacho and
                  Gautier Abou Loume and
                  Abdallah Galal and
                  Hangrui Cao and
                  Alexis C. Garretson and
                  Damien Sileo and
                  Qiuyu Ren and
                  Doru Cojoc and
                  Pavel Arkhipov and
                  Usman Qazi and
                  Lianghui Li and
                  Sumeet Motwani and
                  Christian Schr{\"{o}}der de Witt and
                  Edwin Taylor and
                  Johannes Veith and
                  Eric Singer and
                  Taylor D. Hartman and
                  Paolo Rissone and
                  Jaehyeok Jin and
                  Jack Wei Lun Shi and
                  Chris G. Willcocks and
                  Joshua Robinson and
                  Aleksandar Mikov and
                  Ameya Prabhu and
                  Longke Tang and
                  Xavier Alapont and
                  Justine Leon Uro and
                  Kevin Zhou and
                  Emily de Oliveira Santos and
                  Andrey Pupasov Maksimov and
                  Edward Vendrow and
                  Kengo Zenitani and
                  Julien Guillod and
                  Yuqi Li and
                  Joshua Vendrow and
                  Vladyslav Kuchkin and
                  Ng Ze{-}An},
  title        = {Humanity's Last Exam},
  journal      = {CoRR},
  volume       = {abs/2501.14249},
  year         = {2025},
  url          = {https://doi.org/10.48550/arXiv.2501.14249},
  doi          = {10.48550/ARXIV.2501.14249},
  eprinttype    = {arXiv},
  eprint       = {2501.14249},
  bibsource    = {dblp computer science bibliography, https://dblp.org}
}

@inproceedings{DBLP:conf/kr/LevesqueDM12,
  author       = {Hector J. Levesque and
                  Ernest Davis and
                  Leora Morgenstern},
  editor       = {Gerhard Brewka and
                  Thomas Eiter and
                  Sheila A. McIlraith},
  title        = {The Winograd Schema Challenge},
  booktitle    = {Principles of Knowledge Representation and Reasoning: Proceedings
                  of the Thirteenth International Conference, {KR} 2012, Rome, Italy,
                  June 10-14, 2012},
  publisher    = {{AAAI} Press},
  year         = {2012},
  url          = {https://aaai.org/papers/59-4492-the-winograd-schema-challenge/},
  bibsource    = {dblp computer science bibliography, https://dblp.org}
}

@inproceedings{DBLP:conf/aaaiss/RoemmeleBG11,
  author       = {Melissa Roemmele and
                  Cosmin Adrian Bejan and
                  Andrew S. Gordon},
  title        = {Choice of Plausible Alternatives: An Evaluation of Commonsense Causal
                  Reasoning},
  booktitle    = {Logical Formalizations of Commonsense Reasoning, Papers from the 2011
                  {AAAI} Spring Symposium, Technical Report SS-11-06, Stanford, California,
                  USA, March 21-23, 2011},
  publisher    = {{AAAI}},
  year         = {2011},
  url          = {http://www.aaai.org/ocs/index.php/SSS/SSS11/paper/view/2418},
  bibsource    = {dblp computer science bibliography, https://dblp.org}
}

@inproceedings{lin-etal-2022-shot,
    title = "Few-shot Learning with Multilingual Generative Language Models",
    author = "Lin, Xi Victoria  and
      Mihaylov, Todor  and
      Artetxe, Mikel  and
      Wang, Tianlu  and
      Chen, Shuohui  and
      Simig, Daniel  and
      Ott, Myle  and
      Goyal, Naman  and
      Bhosale, Shruti  and
      Du, Jingfei  and
      Pasunuru, Ramakanth  and
      Shleifer, Sam  and
      Koura, Punit Singh  and
      Chaudhary, Vishrav  and
      O{'}Horo, Brian  and
      Wang, Jeff  and
      Zettlemoyer, Luke  and
      Kozareva, Zornitsa  and
      Diab, Mona  and
      Stoyanov, Veselin  and
      Li, Xian",
    editor = "Goldberg, Yoav  and
      Kozareva, Zornitsa  and
      Zhang, Yue",
    booktitle = "Proceedings of the 2022 Conference on Empirical Methods in Natural Language Processing",
    month = dec,
    year = "2022",
    address = "Abu Dhabi, United Arab Emirates",
    publisher = "Association for Computational Linguistics",
    url = "https://aclanthology.org/2022.emnlp-main.616/",
    doi = "10.18653/v1/2022.emnlp-main.616",
    pages = "9019--9052"
}

@inproceedings{mostafazadeh-etal-2016-corpus,
    title = "A Corpus and Cloze Evaluation for Deeper Understanding of Commonsense Stories",
    author = "Mostafazadeh, Nasrin  and
      Chambers, Nathanael  and
      He, Xiaodong  and
      Parikh, Devi  and
      Batra, Dhruv  and
      Vanderwende, Lucy  and
      Kohli, Pushmeet  and
      Allen, James",
    editor = "Knight, Kevin  and
      Nenkova, Ani  and
      Rambow, Owen",
    booktitle = "Proceedings of the 2016 Conference of the North {A}merican Chapter of the Association for Computational Linguistics: Human Language Technologies",
    month = jun,
    year = "2016",
    address = "San Diego, California",
    publisher = "Association for Computational Linguistics",
    url = "https://aclanthology.org/N16-1098/",
    doi = "10.18653/v1/N16-1098",
    pages = "839--849"
}

@article{llama-series,
  author       = {AI@Meta},
  title        = {The Llama 3 Herd of Models},
  journal      = {CoRR},
  volume       = {abs/2407.21783},
  year         = {2024},
  url          = {https://doi.org/10.48550/arXiv.2407.21783},
  doi          = {10.48550/ARXIV.2407.21783},
  eprinttype    = {arXiv},
  eprint       = {2407.21783},
  bibsource    = {dblp computer science bibliography, https://dblp.org}
}

@article{Light-R1-SFTData,
  author       = {Liang Wen and
                  Yunke Cai and
                  Fenrui Xiao and
                  Xin He and
                  Qi An and
                  Zhenyu Duan and
                  Yimin Du and
                  Junchen Liu and
                  Lifu Tang and
                  Xiaowei Lv and
                  Haosheng Zou and
                  Yongchao Deng and
                  Shousheng Jia and
                  Xiangzheng Zhang},
  title        = {Light-R1: Curriculum SFT, {DPO} and {RL} for Long {COT} from Scratch
                  and Beyond},
  journal      = {CoRR},
  volume       = {abs/2503.10460},
  year         = {2025},
  url          = {https://doi.org/10.48550/arXiv.2503.10460},
  doi          = {10.48550/ARXIV.2503.10460},
  eprinttype    = {arXiv},
  eprint       = {2503.10460},
  bibsource    = {dblp computer science bibliography, https://dblp.org}
}

@inproceedings{MAPO,
  author       = {Shuaijie She and
                  Wei Zou and
                  Shujian Huang and
                  Wenhao Zhu and
                  Xiang Liu and
                  Xiang Geng and
                  Jiajun Chen},
  editor       = {Lun{-}Wei Ku and
                  Andre Martins and
                  Vivek Srikumar},
  title        = {{MAPO:} Advancing Multilingual Reasoning through Multilingual-Alignment-as-Preference
                  Optimization},
  booktitle    = {Proceedings of the 62nd Annual Meeting of the Association for Computational
                  Linguistics (Volume 1: Long Papers), {ACL} 2024, Bangkok, Thailand,
                  August 11-16, 2024},
  pages        = {10015--10027},
  publisher    = {Association for Computational Linguistics},
  year         = {2024},
  url          = {https://doi.org/10.18653/v1/2024.acl-long.539},
  doi          = {10.18653/V1/2024.ACL-LONG.539},
  bibsource    = {dblp computer science bibliography, https://dblp.org}
}

@inproceedings{NumeBLEU,
  author       = {Swaroop Mishra and
                  Arindam Mitra and
                  Neeraj Varshney and
                  Bhavdeep Singh Sachdeva and
                  Peter Clark and
                  Chitta Baral and
                  Ashwin Kalyan},
  editor       = {Smaranda Muresan and
                  Preslav Nakov and
                  Aline Villavicencio},
  title        = {NumGLUE: {A} Suite of Fundamental yet Challenging Mathematical Reasoning
                  Tasks},
  booktitle    = {Proceedings of the 60th Annual Meeting of the Association for Computational
                  Linguistics (Volume 1: Long Papers), {ACL} 2022, Dublin, Ireland,
                  May 22-27, 2022},
  pages        = {3505--3523},
  publisher    = {Association for Computational Linguistics},
  year         = {2022},
  url          = {https://doi.org/10.18653/v1/2022.acl-long.246},
  doi          = {10.18653/V1/2022.ACL-LONG.246},
  bibsource    = {dblp computer science bibliography, https://dblp.org}
}

@misc{anthropic2025claudeopus45,
  author       = {{Anthropic}},
  title        = {Claude Opus 4.5},
  year         = {2025},
  howpublished = {\url{https://www.anthropic.com/news/claude-opus-4-5}},
  note         = {Accessed: 2025-12-30}
}

@article{NLLB,
  author       = {Marta R. Costa{-}juss{\`{a}} and
                  James Cross and
                  Onur {\c{C}}elebi and
                  Maha Elbayad and
                  Kenneth Heafield and
                  Kevin Heffernan and
                  Elahe Kalbassi and
                  Janice Lam and
                  Daniel Licht and
                  Jean Maillard and
                  Anna Y. Sun and
                  Skyler Wang and
                  Guillaume Wenzek and
                  Al Youngblood and
                  Bapi Akula and
                  Lo{\"{\i}}c Barrault and
                  Gabriel Mejia Gonzalez and
                  Prangthip Hansanti and
                  John Hoffman and
                  Semarley Jarrett and
                  Kaushik Ram Sadagopan and
                  Dirk Rowe and
                  Shannon Spruit and
                  Chau Tran and
                  Pierre Andrews and
                  Necip Fazil Ayan and
                  Shruti Bhosale and
                  Sergey Edunov and
                  Angela Fan and
                  Cynthia Gao and
                  Vedanuj Goswami and
                  Francisco Guzm{\'{a}}n and
                  Philipp Koehn and
                  Alexandre Mourachko and
                  Christophe Ropers and
                  Safiyyah Saleem and
                  Holger Schwenk and
                  Jeff Wang},
  title        = {No Language Left Behind: Scaling Human-Centered Machine Translation},
  journal      = {CoRR},
  volume       = {abs/2207.04672},
  year         = {2022},
  url          = {https://doi.org/10.48550/arXiv.2207.04672},
  doi          = {10.48550/ARXIV.2207.04672},
  eprinttype    = {arXiv},
  eprint       = {2207.04672},
  bibsource    = {dblp computer science bibliography, https://dblp.org}
}

@inproceedings{
yu2025dapo,
title={{DAPO}: An Open-Source {LLM} Reinforcement Learning System at Scale},
author={Qiying Yu and Zheng Zhang and Ruofei Zhu and Yufeng Yuan and Xiaochen Zuo and YuYue and Weinan Dai and Tiantian Fan and Gaohong Liu and Juncai Liu and LingJun Liu and Xin Liu and Haibin Lin and Zhiqi Lin and Bole Ma and Guangming Sheng and Yuxuan Tong and Chi Zhang and Mofan Zhang and Ru Zhang and Wang Zhang and Hang Zhu and Jinhua Zhu and Jiaze Chen and Jiangjie Chen and Chengyi Wang and Hongli Yu and Yuxuan Song and Xiangpeng Wei and Hao Zhou and Jingjing Liu and Wei-Ying Ma and Ya-Qin Zhang and Lin Yan and Yonghui Wu and Mingxuan Wang},
booktitle={The Thirty-ninth Annual Conference on Neural Information Processing Systems},
year={2025},
url={https://openreview.net/forum?id=2a36EMSSTp}
}

@article{gandhi2025cognitive,
  author       = {Kanishk Gandhi and
                  Ayush Chakravarthy and
                  Anikait Singh and
                  Nathan Lile and
                  Noah D. Goodman},
  title        = {Cognitive Behaviors that Enable Self-Improving Reasoners, or, Four
                  Habits of Highly Effective STaRs},
  journal      = {CoRR},
  volume       = {abs/2503.01307},
  year         = {2025},
  url          = {https://doi.org/10.48550/arXiv.2503.01307},
  doi          = {10.48550/ARXIV.2503.01307},
  eprinttype    = {arXiv},
  eprint       = {2503.01307},
  bibsource    = {dblp computer science bibliography, https://dblp.org}
}

@article{Hallucination-lhuang,
  author       = {Lei Huang and
                  Weijiang Yu and
                  Weitao Ma and
                  Weihong Zhong and
                  Zhangyin Feng and
                  Haotian Wang and
                  Qianglong Chen and
                  Weihua Peng and
                  Xiaocheng Feng and
                  Bing Qin and
                  Ting Liu},
  title        = {A Survey on Hallucination in Large Language Models: Principles, Taxonomy,
                  Challenges, and Open Questions},
  journal      = {{ACM} Trans. Inf. Syst.},
  volume       = {43},
  number       = {2},
  pages        = {42:1--42:55},
  year         = {2025},
  url          = {https://doi.org/10.1145/3703155},
  doi          = {10.1145/3703155},
  bibsource    = {dblp computer science bibliography, https://dblp.org}
}

@inproceedings{Citation-lhuang,
  author       = {Lei Huang and
                  Xiaocheng Feng and
                  Weitao Ma and
                  Yuxuan Gu and
                  Weihong Zhong and
                  Xiachong Feng and
                  Weijiang Yu and
                  Weihua Peng and
                  Duyu Tang and
                  Dandan Tu and
                  Bing Qin},
  editor       = {Lun{-}Wei Ku and
                  Andre Martins and
                  Vivek Srikumar},
  title        = {Learning Fine-Grained Grounded Citations for Attributed Large Language
                  Models},
  booktitle    = {Findings of the Association for Computational Linguistics, {ACL} 2024,
                  Bangkok, Thailand and virtual meeting, August 11-16, 2024},
  pages        = {14095--14113},
  publisher    = {Association for Computational Linguistics},
  year         = {2024},
  url          = {https://doi.org/10.18653/v1/2024.findings-acl.838},
  doi          = {10.18653/V1/2024.FINDINGS-ACL.838},
  bibsource    = {dblp computer science bibliography, https://dblp.org}
}

@inproceedings{Faithful-lhuang,
  author       = {Lei Huang and
                  Xiaocheng Feng and
                  Weitao Ma and
                  Yuchun Fan and
                  Xiachong Feng and
                  Yangfan Ye and
                  Weihong Zhong and
                  Yuxuan Gu and
                  Baoxin Wang and
                  Dayong Wu and
                  Guoping Hu and
                  Bing Qin},
  editor       = {Wanxiang Che and
                  Joyce Nabende and
                  Ekaterina Shutova and
                  Mohammad Taher Pilehvar},
  title        = {Improving Contextual Faithfulness of Large Language Models via Retrieval
                  Heads-Induced Optimization},
  booktitle    = {Proceedings of the 63rd Annual Meeting of the Association for Computational
                  Linguistics (Volume 1: Long Papers), {ACL} 2025, Vienna, Austria,
                  July 27 - August 1, 2025},
  pages        = {16896--16913},
  publisher    = {Association for Computational Linguistics},
  year         = {2025},
  url          = {https://aclanthology.org/2025.acl-long.826/},
  bibsource    = {dblp computer science bibliography, https://dblp.org}
}

@inproceedings{Misalignment-lhuang,
  author       = {Lei Huang and
                  Xiaocheng Feng and
                  Weitao Ma and
                  Yuchun Fan and
                  Xiachong Feng and
                  Yuxuan Gu and
                  Yangfan Ye and
                  Liang Zhao and
                  Weihong Zhong and
                  Baoxin Wang and
                  Dayong Wu and
                  Guoping Hu and
                  Lingpeng Kong and
                  Tong Xiao and
                  Ting Liu and
                  Bing Qin},
  editor       = {Wanxiang Che and
                  Joyce Nabende and
                  Ekaterina Shutova and
                  Mohammad Taher Pilehvar},
  title        = {Alleviating Hallucinations from Knowledge Misalignment in Large Language
                  Models via Selective Abstention Learning},
  booktitle    = {Proceedings of the 63rd Annual Meeting of the Association for Computational
                  Linguistics (Volume 1: Long Papers), {ACL} 2025, Vienna, Austria,
                  July 27 - August 1, 2025},
  pages        = {24564--24579},
  publisher    = {Association for Computational Linguistics},
  year         = {2025},
  url          = {https://aclanthology.org/2025.acl-long.1199/},
  bibsource    = {dblp computer science bibliography, https://dblp.org}
}

@inproceedings{Self-Improving-lhuang,
  author       = {Lei Huang and
                  Xiaocheng Feng and
                  Weitao Ma and
                  Liang Zhao and
                  Yuchun Fan and
                  Weihong Zhong and
                  Dongliang Xu and
                  Qing Yang and
                  Hongtao Liu and
                  Bing Qin},
  editor       = {Yaser Al{-}Onaizan and
                  Mohit Bansal and
                  Yun{-}Nung Chen},
  title        = {Advancing Large Language Model Attribution through Self-Improving},
  booktitle    = {Proceedings of the 2024 Conference on Empirical Methods in Natural
                  Language Processing, {EMNLP} 2024, Miami, FL, USA, November 12-16,
                  2024},
  pages        = {3822--3836},
  publisher    = {Association for Computational Linguistics},
  year         = {2024},
  url          = {https://doi.org/10.18653/v1/2024.emnlp-main.223},
  doi          = {10.18653/V1/2024.EMNLP-MAIN.223},
  bibsource    = {dblp computer science bibliography, https://dblp.org}
}

@inproceedings{chen-etal-2025-clueanchor,
    title = "{C}lue{A}nchor: Clue-Anchored Knowledge Reasoning Exploration and Optimization for Retrieval-Augmented Generation",
    author = "Chen, Hao  and
      Yan, Yukun  and
      Mei, Sen  and
      Che, Wanxiang  and
      Liu, Zhenghao  and
      Shi, Qi  and
      Li, Xinze  and
      Fan, Yuchun  and
      Huang, Pengcheng  and
      Xiong, Qiushi  and
      Liu, Zhiyuan  and
      Sun, Maosong",
    editor = "Christodoulopoulos, Christos  and
      Chakraborty, Tanmoy  and
      Rose, Carolyn  and
      Peng, Violet",
    booktitle = "Findings of the Association for Computational Linguistics: EMNLP 2025",
    month = nov,
    year = "2025",
    address = "Suzhou, China",
    publisher = "Association for Computational Linguistics",
    url = "https://aclanthology.org/2025.findings-emnlp.1049/",
    doi = "10.18653/v1/2025.findings-emnlp.1049",
    pages = "19258--19278",
    ISBN = "979-8-89176-335-7"
}

@article{chen2026know,
  title={Know More, Know Clearer: A Meta-Cognitive Framework for Knowledge Augmentation in Large Language Models},
  author={Chen, Hao and He, Ye and Fan, Yuchun and Yan, Yukun and Liu, Zhenghao and Zhu, Qingfu and Sun, Maosong and Che, Wanxiang},
  journal={arXiv preprint arXiv:2602.12996},
  year={2026}
}

@inproceedings{zhang-etal-2025-cm,
    title = "{CM}-Align: Consistency-based Multilingual Alignment for Large Language Models",
    author = "Zhang, Xue  and
      Liang, Yunlong  and
      Meng, Fandong  and
      Zhang, Songming  and
      Chen, Yufeng  and
      Xu, Jinan  and
      Zhou, Jie",
    editor = "Christodoulopoulos, Christos  and
      Chakraborty, Tanmoy  and
      Rose, Carolyn  and
      Peng, Violet",
    booktitle = "Findings of the Association for Computational Linguistics: EMNLP 2025",
    month = nov,
    year = "2025",
    address = "Suzhou, China",
    publisher = "Association for Computational Linguistics",
    url = "https://aclanthology.org/2025.findings-emnlp.1401/",
    doi = "10.18653/v1/2025.findings-emnlp.1401",
    pages = "25689--25702",
    ISBN = "979-8-89176-335-7"
}

@inproceedings{zhang-etal-2025-multilingual,
    title = "Multilingual Knowledge Editing with Language-Agnostic Factual Neurons",
    author = "Zhang, Xue  and
      Liang, Yunlong  and
      Meng, Fandong  and
      Zhang, Songming  and
      Chen, Yufeng  and
      Xu, Jinan  and
      Zhou, Jie",
    editor = "Rambow, Owen  and
      Wanner, Leo  and
      Apidianaki, Marianna  and
      Al-Khalifa, Hend  and
      Eugenio, Barbara Di  and
      Schockaert, Steven",
    booktitle = "Proceedings of the 31st International Conference on Computational Linguistics",
    month = jan,
    year = "2025",
    address = "Abu Dhabi, UAE",
    publisher = "Association for Computational Linguistics",
    url = "https://aclanthology.org/2025.coling-main.385/",
    pages = "5775--5788"
}

\newpage
\appendix
\setcounter{equation}{0}
\section{Preliminary Study}
\label{sec:Pilot_study_setup}

\subsection{Construction of Training Data}
\label{subsec:train_dataset}
Due to the scarcity of multilingual mathematical training data, we construct our training data from DeepMath-103K~\citep{he2025deepmath}, which contains high-difficulty math problems paired with reasoning traces generated by DeepSeek-R1~\citep{guo2025deepseek}. We translate the questions and their corresponding reasoning traces from the DeepMath-103K dataset~\citep{he2025deepmath} into ten in-domain languages: Arabic, Bengali, Thai, Swahili, Japanese, Chinese, German, French, and Russian using GPT-4o-mini~\citep{2023-ChatGPT}, followed by manual verification and correction.

\subsection{Metrics for Preliminary Study}
We provide a detailed description of the metrics used in the preliminary study below:
\paragraph{Policy Entropy.} 
The policy entropy can be described as follows.
Given a question $q$ and a policy model $\pi_\theta$, the model generates a response $y = (y_{1},...,y_{t},...,y_{T})$, where each token $y_{t}$ is sampled from the conditional distribution $\pi_\theta\left(\cdot \mid y_{<t}, q\right)$.
Following \citet{shi2025efficient}, we measure the average token-level entropy of the policy model over the training dataset $\mathcal{D}$, as defined by the following equation:
\begin{equation}
\begin{aligned}
\mathcal{H}(\pi_\theta, \mathcal{D}) 
&= \mathbb{E}_{q \sim \mathcal{D},\ y \sim \pi_\theta(\cdot \mid q)} \\
&\quad \left[ \frac{1}{|y|} \sum_{t=1}^{|y|} -\log \pi_\theta(y_t \mid y_{<t}, q) \right].
\end{aligned}
\label{eq:entropy}
\end{equation}

\paragraph{Average Accuracy.} 
We report the average accuracy over ten languages on the MMATH test sets by comparing the extracted final answer in the response to the gold answer.

\paragraph{Reward Score.} 
We report the average reward score, defined as the proportion of rollouts with $R(o)=1$, where $R(o)$ equals 1 only when all reward criteria are satisfied, defined as equation \ref{reward}.

\paragraph{Response Length.} 
We track the model’s response length on the MMATH test sets throughout training. For each evaluation checkpoint, we tokenize the generated responses using the model’s own tokenizer and report the average number of output tokens (including both the reasoning trace and final answer) across the ten languages.

\paragraph{Repeat Score.} Following~\cite{yao-etal-2025-understanding}'s work, we measure repetition using the weighted $n$-gram repetition rate. Before computing repetition, we remove mathematical expressions and symbols from each output via regular expressions and compute the metric on the remaining text. 
Given a generated token sequence $y=(t_1,\ldots,t_T)$, the contiguous $n$-gram starting at position $j$ is defined as:

\begin{equation}
\small
\text{$n$-gram}_j = (t_j, t_{j+1}, \ldots, t_{j+n-1}), \quad j=1,\ldots,T-n+1.
\end{equation}
Let $\{n_i\}_{i=1}^{K}$ denote the unique $n$-grams in $y$, with frequency $f_i$ for each $n_i$.
We compute the repeat score as specified by the following equation:
\begin{equation}
\text{Repeat Score}_n(y)=
\frac{\sum_{i=1}^{K} f_i^{\,w}\cdot \mathbb{I}(f_i>1)}
{\sum_{i=1}^{K} \max(f_i,1)^{\,w}},
\end{equation}
where $\mathbb{I}(\cdot)$ is the indicator function and $w$ is a weighting factor. Following~\cite{yao-etal-2025-understanding}'s setup, we set $n=1$, $w=1$ and report the average repeat score over all generated outputs across ten languages on the MMATH benchmark.

\section{Experimental Details of Multilingual Mathematical Reasoning Task}
\subsection{Evaluation Datasets}
\label{sec:Details_MMATH_PolyMath}
The datasets for evaluating the multilingual reasoning ability of LLMs cover two benchmarks: MMATH and PolyMath. The detailed descriptions of these two datasets are as follows. 
\paragraph{MMATH~\citep{luo-etal-2025-mmath}} is a new benchmark for evaluating multilingual complex reasoning. It contains 374 high-quality math problems spanning 10 typologically and geographically diverse languages, resulting in 3,740 test instances in total. The source problems are drawn from 30 questions from AIME 2024, 15 from AIME 2025\footnote{\url{https://maa.org/maa-invitational-competitions/}}, 18 from CNMO\footnote{\url{https://www.cms.org.cn/}}, and 311 filtered problems from MATH500~\citep{2024MATHCobbe}. Each problem is translated and validated through a rigorous pipeline that combines frontier LLMs with human verification, ensuring semantic equivalence across languages.  The test sets include 10 languages: English (en), Chinese (zh), Arabic (ar), Spanish (es), French (fr), Japanese (ja), Korean (ko), Portuguese (pt), Thai (th), and Vietnamese (vi).
\paragraph{PolyMath~\citep{wang2025polymath}} is a multilingual benchmark organized by comprehensive difficulty levels, spanning from K-12 to Olympiad and advanced frontier mathematics. It covers 18 languages: English (en), Chinese (zh), Spanish (es), Arabic (ar), French (fr), Bengali (bn), Portuguese (pt), Russian (ru), Indonesian (id), German (de), Japanese (ja), Swahili (sw), Vietnamese (vi), Italian (it), Telugu (te), Korean (ko), Thai (th), and Malay (ms). And the problems are categorized into four difficulty levels based on Thought Depth and Knowledge Breadth, with 125 problems per level. The data sources for each level are summarized as follows:
\begin{itemize}
    \item \textbf{Low-level:} Problems are sourced from MGSM~\citep{Shi2024mCOT}, a multilingual math word-problem benchmark, with additional translations supplemented by P-MMeval~\citep{zhang-etal-2025-p}.
    \item \textbf{Medium-level:} Problems are collected from \textit{College Math} exercise sets and standardized examinations (e.g., China’s \textit{Gaokao} and postgraduate entrance exams), together with problems from widely used contest archives such as the U.S. AMC and China’s provincial CNMO selection contests.
    \item \textbf{High-level:} Problems are drawn from established competition problem sets, including the U.S. AIME and China’s CNMO.
    \item \textbf{Top-level:} Problems are aggregated from the IMO/IMO Shortlist and major national/regional Olympiads (e.g., CMO, USAMO, Putnam), and complemented with frontier problems from the HLE dataset~\citep{DBLP:journals/corr/abs-2501-14249}.
\end{itemize}

\subsection{Evaluation Details}
\label{sec:Details_Details_Metrics}
\subsubsection{Evaluation Details for MMATH}
Following~\citet{luo-etal-2025-mmath}, we generate outputs using a temperature of t = 0.6, a top-p value of 0.95, and a maximum output length of 32{,}768 tokens. To obtain a more reliable estimate of reasoning accuracy, each evaluation is repeated 4 times, and the average result is recorded. Given the varying complexity of each benchmark subset, we report the final score using macro-average accuracy. To extract the final answer, we employ the math extraction tool from~\citet{2023opencompass}, and the extracted answers are then verified against the ground truth using \texttt{math\_verify}\footnote{\url{https://github.com/huggingface/Math-Verify}}. We utilize the inference prompt from~\citet{luo-etal-2025-mmath}. The evaluation prompts used for different methods on the MMATH test set are shown in Figure~\ref{fig:mmath_eval_prompts}.

\subsubsection{Evaluation Details for PolyMath}
For a fair comparison, we use greedy decoding to ensure the determinism of the outputs and set the maximum generation length to 65{,}536 tokens during inference. To ensure a fair comparison and mitigate the
risk of hallucinations~\cite{Hallucination-lhuang,Citation-lhuang,Faithful-lhuang,chen-etal-2025-clueanchor,Zhao_Min_Wu_Li_Sun_Cai_Wang_Chen_Penn_2026,Misalignment-lhuang,Self-Improving-lhuang}, we adopt the inference prompt from~\citet{luo-etal-2025-mmath}. The evaluation prompts used for different methods on the PolyMath test set are shown in Figure~\ref{fig:polymath_eval_prompts}. 

Following~\citet{luo-etal-2025-mmath}, we report the Difficulty-Weighted Accuracy (DW-ACC)~\citep{luo-etal-2025-mmath} on the PolyMath benchmark. DW-ACC assigns level-specific weights $w_1,w_2,w_3,w_4$ to problems from the \textit{low}, \textit{medium}, \textit{high}, and \textit{top} levels, respectively, with weights doubling as difficulty increases: $w_1=1$, $w_2=2$, $w_3=4$, and $w_4=8$. This weighting scheme reduces the influence of easier problems and places greater emphasis on correctness at higher difficulty levels. The accuracies at the four levels are denoted by $a_1,a_2,a_3,a_4$, and DW-ACC is specified by the following equation:
\begin{equation}
    \text{DW-ACC}=\frac{\sum_{i=1}^{4} w_i a_i}{\sum_{i=1}^{4} w_i}
    =\sum_{i=1}^{4}\left(\frac{2^{i-1}}{15}a_i\right).
\end{equation}

\section{Experimental Details of \ours}
\label{sec:Experimental_Details}
\subsection{Training Datasets}
We train our models on the multilingual DeepMath-103K dataset, with full details of the data construction provided in Appendix~\ref{subsec:train_dataset}.

\subsection{Training Details of Cold-Start training}
We utilize LLaMA-Factory\footnote{\url{https://github.com/hiyouga/LLaMA-Factory}} as our cold-start training framework. All models are trained using NVIDIA H800 GPUs. All models are trained for 1 epoch with a batch size of 256, and the learning rate is set to 2e-5. We set the maximum token length to be 16384. We use Deepspeed stage 2~\citep{deepspeed2} to conduct multi-GPU distributed training, with training precision Bfloat16 enabled.

\subsection{Training Details of RL training}
We utilize the GRPO algorithm implemented by verl\footnote{\url{https://github.com/volcengine/verl}}. All models are trained using 4×8 NVIDIA H800 GPUs. All models are trained for 5 epochs with a batch size of 128 and a PPO mini-batch size of 64. We use a learning rate of $1\times10^{-6}$, 8 rollouts with temperature 1.0, and a KL coefficient of 0.0. The maximum sequence length is 16,384 tokens. Following~\citet{joshi-etal-2020-state}, the languages in this study are classified into three categories based on resource availability: high-resource languages include English, German, French, Spanish, Portuguese, and Italian; mid-resource languages include Japanese, Chinese, Russian, Korean, and Vietnamese; and low-resource languages include Arabic, Bengali, Thai, Swahili, Telugu, and Indonesian.

\section{Detailed Results for LCR and Accuracy}
\label{sec:Experiment_of_Acc_and_LC}
We report the LCR of Qwen2.5-3B-Instruct on MMATH and PolyMath in Figure~\ref{table:mmath-3B-LC} and Figure~\ref{table:polymath-3B-LC}, respectively.
For Qwen2.5-3B-Instruct and Qwen2.5-7B-Instruct on the PolyMath test sets, we report the difficulty-weighted accuracy in Table~\ref{table:polymath-acc}, and accuracy by difficulty level in Tables~\ref{table:polymath-acc-top}--\ref{table:polymath-acc-low}.
For the PolyMath test sets, we report the difficulty-weighted accuracy in Table~\ref{table:polymath-acc}. We additionally report per-tier accuracies for the four difficulty levels in Tables~\ref{table:polymath-acc-top}--\ref{table:polymath-acc-low}.

\definecolor{red2}{RGB}{172,21,28}
\definecolor{blue}{RGB}{39,89,167}
\definecolor{red3}{RGB}{203,104,104}
\definecolor{blue1}{RGB}{104,155,203}
\definecolor{color1}{RGB}{235,164,122}
\definecolor{color2}{RGB}{78,172,183}

\definecolor{upink}{HTML}{fcd4d4}
\definecolor{ucyan}{HTML}{e3eeff}
\definecolor{uedgecyan}{HTML}{6d97e0}
\definecolor{uedgepink}{HTML}{cc0000}

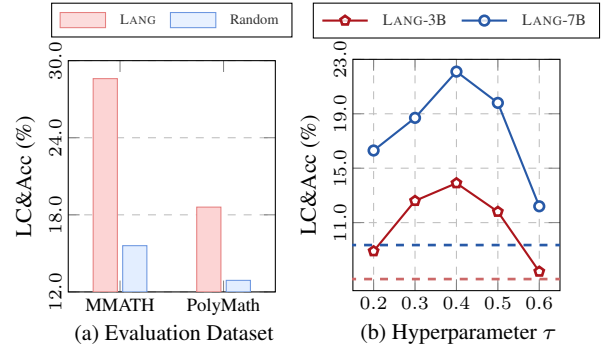
\begin{figure}[t]
    \centering
    \vspace{1.0mm}
\begin{tikzpicture}
    \scriptsize{
\begin{axis}
[
    anchor=north west,
    at={(11.5em,5.5em)},
    ymajorgrids,
    xmajorgrids,
    grid style=dashed,
    width=.27\textwidth,
    height=.29\textwidth,
    yticklabel style={/pgf/number format/precision=2,/pgf/number format/fixed zerofill,scale=1.0},
    xmax=0.65,
    xmin=0.15,
    ymin=6.0,
    ymax=23.0,
    xtick={0.2,0.3,0.4,0.5,0.6},
    ytick={11.0,15.0,19.0,23.0},
    xlabel={\scriptsize{(b) Hyperparameter $\tau$}},
    xlabel style={scale=1.2, yshift=0em, xshift=0.1em},
    ylabel=\scriptsize {LC\&Acc (\%)},
    ylabel style={yshift=0em, scale=1.2},
    legend style={
        at={(0.5,1.09)},  
        anchor=south,    
        font={\tiny},
        cells={anchor=west},
        fill opacity=0.8,
        scale=0.6,
        legend columns=2,  
        column sep=1ex     
    },
    yticklabel style={rotate=90},
    yticklabel style={
        /pgf/number format/precision=1,
        /pgf/number format/fixed zerofill,
        scale=1.0
    }
]

\addplot[red2,mark=pentagon*,mark size=1.8pt,thick,mark options={fill=white,draw=red2,line width=1pt}] coordinates {(0.2,8.9) (0.3,12.6) (0.4,13.9) (0.5,11.8) (0.6,7.4)};
\addlegendentry{\scalebox{1}{\ours-3B}}

\addplot[blue,mark=*,mark size=1.8pt,thick,mark options={fill=white,draw=blue,line width=1pt}] coordinates {(0.2,16.3) (0.3,18.7) (0.4,22.1) (0.5,19.8) (0.6,12.2)};
\addlegendentry{\scalebox{1}{\ours-7B}}  

\draw[red3, dashed,line width=1pt] (axis cs:0.1,6.85) -- (axis cs:0.7,6.85); 
\draw[blue, dashed,line width=1pt] (axis cs:0.1,9.35) -- (axis cs:0.7,9.35); 
\end{axis}

  \begin{axis}[
    at={(-3.8em,-7em)},
    anchor=south west,
    ymajorgrids,
    grid style=dashed,
    legend style={
        at={(0.5,1.1)},  
        anchor=south,    
        font={\tiny},
        cells={anchor=west},
        fill opacity=0.8,
        scale=1.0,
        legend columns=2,  
        column sep=1ex     
    },
    legend cell align={left},
    ybar,
    enlarge x limits=0.5,
    xtick align=inside,
    height=.29\textwidth,
    width=.27\textwidth,
    bar width=1.3em,
    xlabel={\scriptsize{(a) Evaluation Dataset}},
    xlabel style={scale=1.2, yshift=0.2em, xshift=0.1em},
    ylabel=\scriptsize {LC\&Acc (\%)},
    ylabel style={scale=1.2, yshift=2em},
    symbolic x coords={{1}, {2}},
    xtick=data,
    ymin=12.0,
    ymax=30.0,
    ytick={12.0, 18.0, 24.0, 30.0},
    nodes near coords align={vertical},
    xticklabels={MMATH, PolyMath},
    ylabel style={yshift=-2em},
    yticklabel style={/pgf/number format/fixed,/pgf/number format/fixed zerofill,/pgf/number format/precision=1,rotate=0,scale=1.0},
    yticklabel style={rotate=90} 
]
    \addplot[fill=upink, draw=uedgepink!50, area legend] coordinates {({1},28.6) ({2},18.6)};
    \addlegendentry{\scalebox{0.9}{\ours}}
    \addplot[fill=ucyan,draw=uedgecyan, area legend] coordinates {({1},15.6) ({2},12.9)};
    \addlegendentry{\scalebox{0.9}{Random}}
\end{axis}

}   
\end{tikzpicture}
    \vspace{-3mm}
    \caption{ (a) The impact of randomly discarding segments of the hint on model performance. (b) The effect of the choice of the hyperparameter $\tau$ on the performance of \ours. \textcolor{red3}{Red dashed line} denotes the performance of Qwen2.5-3B-Instruct, and \textcolor{blue}{Blue dashed line} denotes the performance of Qwen2.5-7B-Instruct.
    }
    \label{fig:ablation_difference_tau}
\end{figure}

\section{Additional Ablation Studies}
\label{sec:additional_ablations}
\paragraph{Effect of random hint segment discarding.}
To evaluate the necessity of truncating hints from the tail, we conduct ablation studies by maintaining the total length of hint decay, but randomly discarding segments of the hint. As shown in Figure~\ref{fig:ablation_difference_tau} (a), randomly discarding segments causes semantic disruption in the multilingual hints, leading to a sharp decline in the model's average performance across both benchmarks. This suggests that semantically disrupted multilingual hints fail to provide effective guidance during training, making it difficult for the model to learn coherent and logical reasoning traces.

\paragraph{Effect of $\tau$ reveals a meaningful trade-off across language groups yet consistent robustness.}
To demonstrate the impact of the choice of $\tau$ on our method, we conduct ablation studies by training Qwen2.5-3B-Instruct and Qwen2.5-7B-Instruct with different values of $\tau$, reporting their average performance on MMATH and PolyMath. As shown in Figure~\ref{fig:ablation_difference_tau} (b),  selecting too small a value for $\tau$ may cause hints to be turned off too early during training, preventing the model from receiving sufficient multilingual hint guidance and making it difficult to explore correct reasoning traces. In contrast, choosing too large a value for $\tau$ may cause hints to be turned off too late during training, leading the model to become overly reliant on hint guidance and preventing it from developing sufficient autonomous exploration capabilities. More concretely, the hyperparameter $\tau$ mediates a meaningful trade-off across language groups with different resource levels. A smaller $\tau$ removes hints earlier, which benefits high-resource languages by encouraging more autonomous exploration, but deprives low-resource languages of the guidance they critically need. A larger $\tau$ retains hints longer, which provides low-resource languages with extended support but may induce dependency in high-resource ones. To illustrate this, we report the performance of Qwen2.5-7B-Instruct on PolyMath across high-, medium-, and low-resource language groups in Table~\ref{table:ablation_difference_tau}. As shown, $\tau=0.4$ achieves the best overall balance across all three groups, while $\tau=0.3$ favors high-resource languages at the cost of low-resource performance. The sensitivity observed in Figure~\ref{fig:ablation_difference_tau} (b) thus reflects the inherent difficulty gap across language groups rather than a design fragility. Notably, our method consistently outperforms the Qwen2.5-3B/7B-Instruct baselines regardless of the value of $\tau$. Even at the suboptimal setting of $\tau=0.3$, \ours still surpasses LC-GRPO on low-resource languages by 7.7\%, demonstrating strong robustness precisely where improvement is most needed. These results highlight that \ours can robustly enhance multilingual reasoning ability by injecting multilingual hints during the early stages of training to guide correct reasoning path generation, while adaptively removing hints to encourage autonomous exploration.

\begin{table}[t]
  \centering
  \footnotesize
  
  \label{tab:results}
  \resizebox{\columnwidth}{!}{
    \begin{tabular}{lccc}
      \toprule
      & High-Resource & Medium-Resource & Low-Resource \\
      \midrule
      \ours ($\tau$=0.3) & 17.8 & 20.2 & 18.1 \\
      \ours ($\tau$=0.4) & 25.1 & 21.5 & 19.6 \\
      LC-GRPO           & 23.2 & 20.0 & 16.8 \\
      \bottomrule
    \end{tabular}%
  }
  \caption{Performance of Qwen2.5-7B-Instruct on PolyMath across high-, medium-, and low-resource language groups under different $\tau$ values.}
  \label{table:ablation_difference_tau}
\end{table}

\section{Comprehensive Results for Additional Training Models}
\label{sec:qwen32B-llama}
We additionally report LC\&Acc results for Qwen2.5-32B-Instruct and Llama3.1-8B-Instruct. Table~\ref{table:mmath-LC-Acc-llama} summarizes LC\&Acc on the MMATH test set, and Table~\ref{table:polymath-LC-Acc-llama} reports LC\&Acc on the PolyMath test set.

\section{Experimental Details of Multilingual Non-Mathematical Reasoning Tasks}

\label{sec:Experiment_of_Xtasks}
 we evaluate it on multilingual understanding benchmarks, including MMLU-ProX~\citep{xuan2025mmlu}, XWinograd~\citep{muennighoff2022crosslingual,tikhonov2021heads}, and XCOPA~\citep{ponti-etal-2020-xcopa}, and additionally assess its performance on the multilingual generation benchmark XStoryCloze~\citep{DBLP:journals/corr/abs-2112-10668}. For a fair comparison, we utilize lm-evaluation-harness\footnote{\url{https://github.com/EleutherAI/lm-evaluation-harness}} as our evaluation framework. The details of the four benchmarks are summarized as follows:

\paragraph{MMLU-ProX~\citep{xuan2025mmlu}} is a novel multilingual benchmark that builds upon the challenging, reasoning-focused design of MMLU-Pro while extending its coverage to 29 typologically diverse languages, including English (en), Chinese (zh), Japanese (ja), Korean (ko), French (fr), German (de), Spanish (es), Portuguese (pt), Arabic (ar), Thai (th), Hindi (hi), Bengali (bn), Swahili (sw), Afrikaans (af), Czech (cs), Hungarian (hu), Indonesian (id), Italian (it), Marathi (mr), Nepali (ne), Russian (ru), Serbian (sr), Telugu (te), Ukrainian (uk), Urdu (ur), Vietnamese (vi), Wolof (wo), Yoruba (yo), and Zulu (zu). MMLU-ProX provides 11{,}829 parallel questions aligned across these languages, thereby enabling a comprehensive evaluation of LLMs' multilingual reasoning capabilities.

\paragraph{XWinograd~\citep{muennighoff2022crosslingual,tikhonov2021heads}} extends the original English Winograd Schema Challenge (WSC)~\citep{DBLP:conf/kr/LevesqueDM12} to five additional languages: French (fr), Japanese (ja), Portuguese (pt), Russian (ru), and Chinese (zh). The dataset comprises pronoun-resolution problems designed to evaluate a model's commonsense reasoning ability.

\paragraph{XCOPA~\citep{ponti-etal-2020-xcopa}} is a multilingual extension of the Choice of Plausible Alternatives (COPA) task, constructed by translating and re-annotating the validation and test sets of the English (en) COPA dataset~\citep{DBLP:conf/aaaiss/RoemmeleBG11} into 11 target languages, including Estonian (et), Haitian Creole (ht), Indonesian (id), Italian (it), Quechua (qu), Swahili (sw), Tamil (ta), Thai (th), Turkish (tr), Vietnamese (vi), and Chinese (zh). Each instance consists of a premise, a question (cause or result), and two candidate alternatives. The task is to select the more plausible alternative.

\paragraph{XStoryCloze~\citep{DBLP:journals/corr/abs-2112-10668}} is constructed by~\citet{lin-etal-2022-shot} by translating the validation split of the original English StoryCloze dataset~\citep{mostafazadeh-etal-2016-corpus} into 10 typologically diverse languages: Russian (ru), Chinese (zh), Spanish (es), Arabic (ar), Hindi (hi), Indonesian (id), Telugu (te), Swahili (sw), Basque (eu), and Burmese (my). Each example consists of a four-sentence commonsense story, a correct ending, and a plausible but incorrect ending.

\section{Experimental Details of Baseline Methods}

\label{sec:all_baselines}
We conduct experiments on the Qwen2.5-3B-Instruct, Qwen2.5-7B-Instruct, Qwen2.5-32B-Instruct~\citep{yang2025qwen3} and Llama3.1-8B-Instruct~\citep{llama-series}. The detailed baseline implementation is as follows:

\subsection{Prompting-based Methods}

\label{sec:prompting-based-methods}

\begin{figure*}[t]
  \centering
\includegraphics[width=\textwidth,height=0.82\textheight,keepaspectratio]{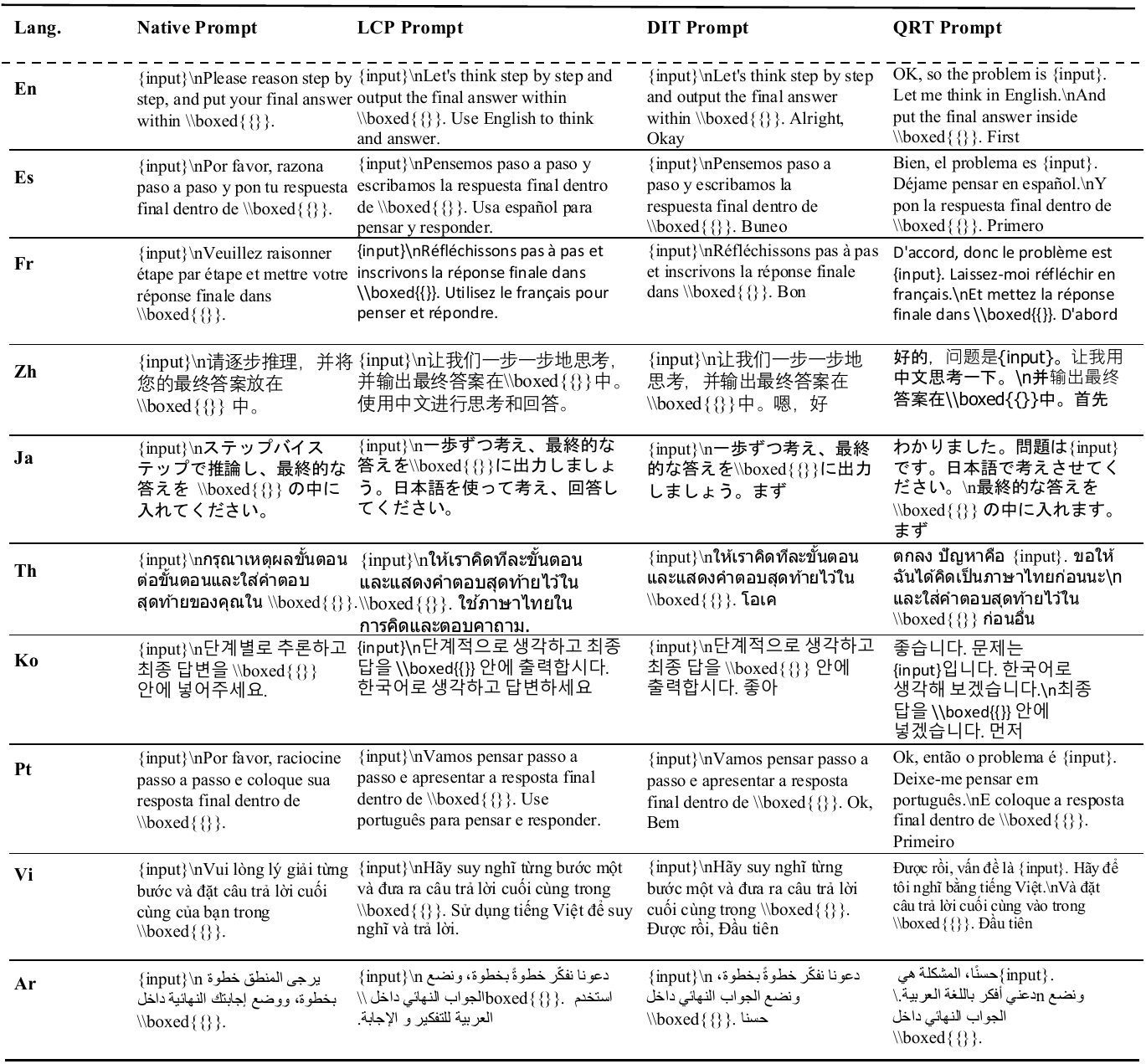}
    \caption{Prompts utilized to evaluate different methods on the MMATH test sets.}
  \label{fig:mmath_eval_prompts}
\end{figure*}

\begin{figure*}[t]
  \centering
\includegraphics[width=\textwidth,height=0.92\textheight,keepaspectratio]{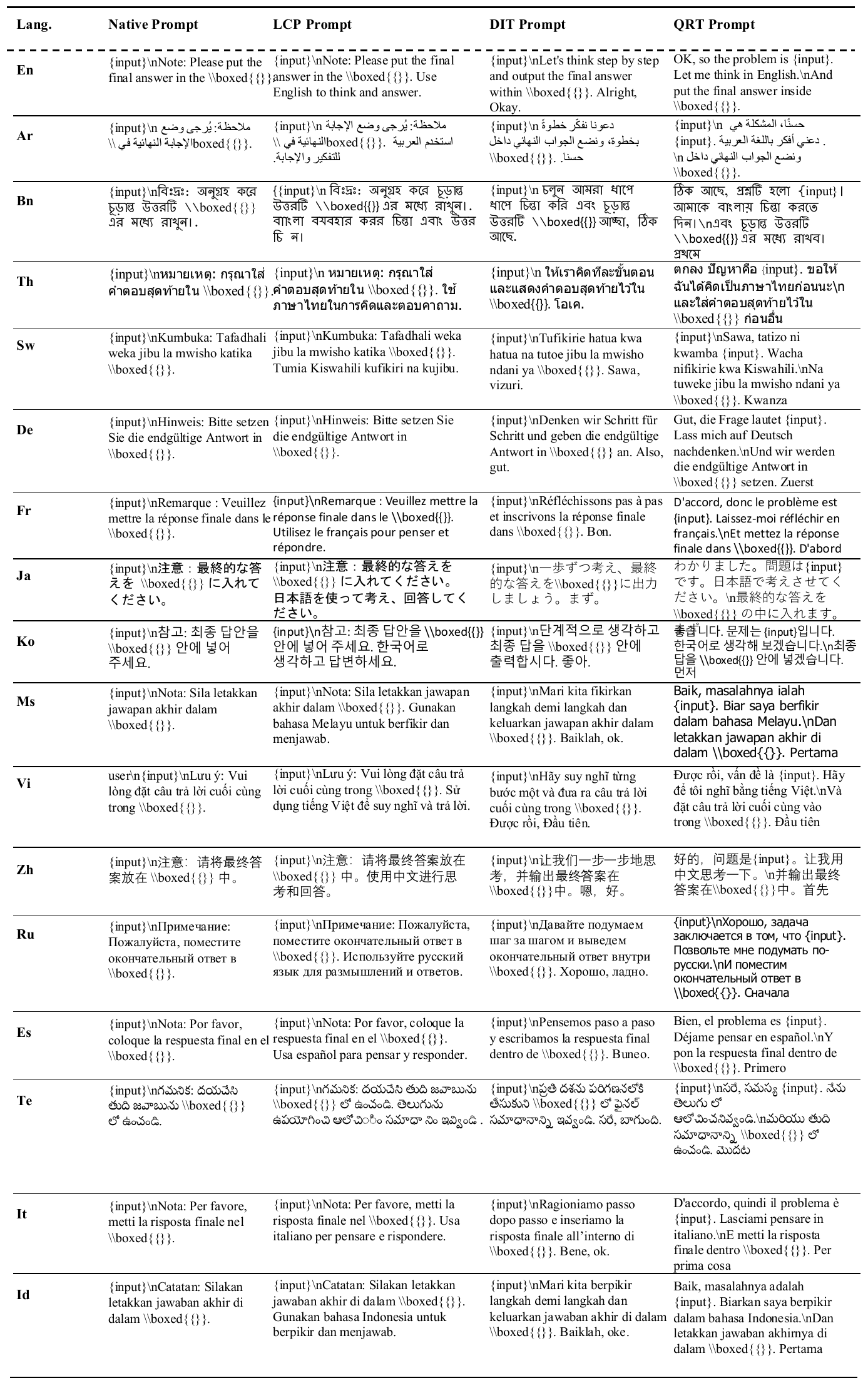}
  \caption{Prompts utilized to evaluate different methods on the PolyMath test sets.}
  \label{fig:polymath_eval_prompts}
\end{figure*}

The evaluation prompt templates of prompting-based methods used for MMATH and PolyMath datasets are shown in Figure~\ref{fig:mmath_eval_prompts} and~\ref{fig:polymath_eval_prompts}, respectively. In the template, \{\textit{input}\} can be replaced with multilingual questions.
\paragraph{Language-Constraint Prompting (LCP)~\citep{wang2025polymath}: }

For each question in the MMATH and PolyMath test sets, LCP explicitly prompts models to solve questions in targeted languages.

\paragraph{Discourse-Initiated Thinking (DIT)~\citep{luo-etal-2025-mmath}: }
Inspired by the phenomenon, the model tends to start thinking with discourse markers like “Alright” or “Okay”. DIT extracts these multilingual markers from native prompt responses and append after the “<think>” token. This approach encourages models to initiate their reasoning using discourse cues as entry points into the multilingual thinking process.

\paragraph{Question-Restatement Thinking (QRT)~\citep{luo-etal-2025-mmath}:}
Inspired by another common phenomenon, which models often restate the question before engaging in actual reasoning. QRT replicates this behavior by explicitly inserting a restated version of the question at the beginning of the thinking process. This intervention encourages the model to frame the problem before attempting to solve it.
\subsection{Training-based Methods}
\label{sec:training-based-methods}

\paragraph{Multilingual Supervised Fine-tuning (\textsc{M-SFT}):}
This method involves directly full-parameter fine-tuning models with constructed multilingual data. During training, we train all models for 3 epochs with a batch size of 256, and the learning rate is maintained at 2e-5 using NVIDIA H800 GPUs. 

\paragraph{Vanilla GRPO~\citep{guo2025deepseek}:}
We adopt the vanilla GRPO algorithm with only format and accuracy rewards to enhance the model’s multilingual reasoning ability. We use the same training data as in \ours.

\paragraph{Language-Consistency GRPO (LC-GRPO):}
To maintain input–output language consistency, we further incorporate the language consistency reward into the GRPO algorithm. To ensure a fair comparison, we use the same training hyperparameters as \ours.

\paragraph{M-Thinker~\citep{zhang2025think}:}
Following~\citet{zhang2025think}, we use the Light-R1-SFTData dataset~\citep{Light-R1-SFTData}, which contains approximately 76K samples. Each sample consists of an English question paired with a high-quality response generated by DeepSeek-R1~\citep{DeekSeek-R1}. We then use \texttt{DeepSeek-V3-0324} to translate the English questions into ten in-domain languages: Arabic, Bengali, Thai, Swahili, Japanese, Chinese, German, French, and Russian.

For cold-start training, we randomly sample 7.5K questions for each in-domain language and employ \texttt{DeepSeek-R1-0528} to generate responses in the same language as the input, yielding 75K cold-start samples in total. For RL training, we perform rejection sampling on the remaining Light-R1-SFTData with $N=8$ sampled candidates per prompt. We keep a prompt only if it yields mixed outcomes among the candidates (i.e., $0 < |O_{\text{correct}}| < N$), which avoids degenerate groups with all-zero or all-one rewards and thus provides effective advantage signals for GRPO. From the resulting filtered RL pool, we randomly select 3K samples per in-domain language to ensure a language-balanced training set under a fixed compute budget. We use \texttt{DeepSeek-V3-0324} to compute the alignment ratio by measuring the overlap between the English reasoning trace and the corresponding target-language reasoning trace, and we set the maximum generation length to 16{,}384 tokens.

\paragraph{mGRPO~\citep{mGRPO}:}
This method proposes a Polyglot Reasoning Generation Module (PRGM) to guide the LLM to generate a group of $n$ multilingual responses for each question. Given an input question, we produce $n$ candidate responses using prompts with or without explicit language instructions. Specifically, one response is generated without any language constraint, while the remaining responses are generated with prompts that explicitly specify a reasoning language randomly sampled from a predefined multilingual set, thereby encouraging broader exploration of the multilingual reasoning space. Following~\citet{mGRPO}, we adopt the mathematical reasoning dataset from MAPO~\citep{MAPO} as training data, which contains 1,703 English questions from a subset of NumGLUE along with ChatGPT-translated versions in nine languages. Since the MAPO training data does not include Arabic translations, we additionally translate the NumGLUE~\citep{NumeBLEU} questions into Arabic using GPT-4o-mini. For the Qwen2.5 series models, we use 10 in-domain languages set to guide the rollout, matching the languages covered by the training data. We set the learning rate to 1e-6, sample 5 rollouts per prompt, and train for 5 epochs with a batch size of 128.

\begin{table}[t]
\centering
\resizebox{\columnwidth}{!}{%
\begin{tabular}{l*{11}{c}}
\toprule
\textbf{Method} & \textbf{Ar} & \textbf{Th} & \textbf{Fr} & \textbf{Ja} & \textbf{Zh} & \textbf{En} & \textbf{Vi} & \textbf{Ko} & \textbf{Pt} & \textbf{Es} & \textbf{Avg.} \\
\midrule
Qwen2.5-7B-Instruct & 0.3 & 0.5 & 0.2 & 3.6 & 21.0 & 28.2 & 1.3 & 1.8 & 0.7 & 2.1 & 6.0 \\
\hspace{2mm}+ \ours & 26.1 & 27.8 & 31.8 & 28.2 & 21.4 & 33.1 & 29.5 & 23.8 & 32.6 & 31.1 & 28.5 \\
\bottomrule
\end{tabular}%
}
\caption{The LC\&Acc (\%) on MMATH under different teacher models 
for reasoning-trace generation, using Qwen2.5-7B-Instruct as the backbone.}
\label{tab:teacher_mmath}
\end{table}

\begin{table}[t]
\centering
\resizebox{\columnwidth}{!}{%
\begin{tabular}{l*{6}{c}}
\toprule
\textbf{Method} & \textbf{Ar} & \textbf{Bn} & \textbf{Th} & \textbf{Sw} & \textbf{Ja} & \textbf{Zh} \\
\midrule
Qwen2.5-7B-Instruct & 12.1 & 11.8 & 12.7 & 2.2 & 13.8 & 15.3 \\
\hspace{2mm}+ \ours         & 16.0 & 10.6 & 15.5 & 3.6 & 14.9 & 15.8 \\
\midrule
\textbf{Method} & \textbf{De} & \textbf{Fr} & \textbf{Ru} & \textbf{En} & \textbf{Te} & \textbf{Ko} \\
\midrule
Qwen2.5-7B-Instruct & 15.3 & 15.2 & 14.8 & 17.3 & 6.2 & 12.4 \\
\hspace{2mm}+ \ours         & 17.2 & 17.4 & 19.7 & 18.0 & 7.7 & 15.8 \\
\midrule
\textbf{Method} & \textbf{Vi} & \textbf{It} & \textbf{Id} & \textbf{Pt} & \textbf{Es} & \textbf{Avg.} \\
\midrule
Qwen2.5-7B-Instruct & 14.4 & 13.4 & 9.7  & 13.6 & 15.1 & 12.7 \\
\hspace{2mm}+ \ours         & 17.3 & 19.8 & 17.2 & 17.9 & 18.8 & 15.5 \\
\bottomrule
\end{tabular}%
}
\caption{The LC\&Acc (\%) on PolyMath under different teacher models 
for reasoning-trace generation, using Qwen2.5-7B-Instruct as the backbone.}
\label{tab:teacher_polymath}
\end{table}

\begin{table*}[t]
    \centering
    \small
    \definecolor{upgray}{RGB}{128, 128, 128}
\resizebox{\linewidth}{!}{%
\begin{tabular}{lccccccccccccc}
\toprule
\multirow{2}{*}{\textbf{Method}} & \multicolumn{7}{c}{\textbf{In-Domain Languages}} & \multicolumn{5}{c}{\textbf{Out-of-Domain Languages}} & \multirow{2}{*}{\textbf{ALL-Avg.}} \\
\cmidrule(lr){2-8} \cmidrule(lr){9-13}
 & \textbf{Ar} & \textbf{Th} & \textbf{Fr} & \textbf{Ja} & \textbf{Zh} & \textbf{En} & \textbf{Avg.} & \textbf{Vi} & \textbf{Ko} & \textbf{Pt} & \textbf{Es} & \textbf{Avg.} & \\
\midrule

\rowcolor{gray!30}
\textit{\textbf{Llama3.1-8B-Instruct}} & \phantom{0}7.2 & \phantom{0}9.7 & 13.0 & \phantom{0}7.9 & \phantom{0}9.3 & 15.2 & 10.4 & 10.8 & \phantom{0}8.4 & 11.2 & 12.4 & 10.7 & 10.5 \\
& \textcolor{upgray} {\phantom{0}7.4} & \textcolor{upgray}{10.0} & \textcolor{upgray}{13.1} & \textcolor{upgray}{\phantom{0}9.3} & \textcolor{upgray}{11.0} & \textcolor{upgray}{15.2} & \textcolor{upgray}{11.0} & \textcolor{upgray}{10.8} & \textcolor{upgray}{\phantom{0}8.9} &\textcolor{upgray}{11.7} & \textcolor{upgray}{12.7} & \textcolor{upgray}{11.0} & \textcolor{upgray}{11.0} \\
\hspace{2mm}+ Vanilla GRPO & \phantom{0}3.9 & \phantom{0}7.0 & \phantom{0}6.0 & \phantom{0}3.5 & \phantom{0}2.6 & \phantom{0}9.7 & \phantom{0}5.5 & \phantom{0}6.2 & \phantom{0}3.3 & \phantom{0}5.5 & \phantom{0}6.9 & \phantom{0}5.5 & \phantom{0}5.5 \\
\hspace{2mm}+ LC-GRPO & \phantom{0}3.4 & \phantom{0}8.8 & \phantom{0}6.9 & \textbf{\phantom{0}8.8} & \phantom{0}6.2 & \phantom{0}5.8 & \phantom{0}6.6 & \phantom{0}7.0 & \phantom{0}4.9 & \phantom{0}6.2 & \phantom{0}7.0 & \phantom{0}6.3 & \phantom{0}6.5 \\
\headercolorSLAM
\hspace{2mm}+ \ours & \textbf{\phantom{0}7.4} & \textbf{\phantom{0}9.9} & \textbf{14.3} & \phantom{0}8.6 & \textbf{12.2} & \textbf{16.7} & \textbf{11.5} & \textbf{11.1} & \textbf{\phantom{0}9.3} & \textbf{12.4} & \textbf{12.9} & \textbf{11.4} & \textbf{11.5} \\

\midrule

\rowcolor{gray!30}
\textit{\textbf{Qwen2.5-32B-Instruct}} & 26.2 & 26.1 & 27.9 & 27.1 & 24.6 & 34.7 & 27.8 & 27.6 & 25.8 & 30.8 & 29.6 & 28.4 & 28.0 \\
\bottomrule
\end{tabular}%
}

  \caption{The LC\&Acc (\%) on MMATH test sets for Llama3.1-8B-Instruct and Qwen2.5-32B-Instruct.} 
  \label{table:mmath-LC-Acc-llama}
\end{table*}

\begin{table*}[t]
    \centering
    \small
    \definecolor{upgray}{RGB}{128, 128, 128}
\resizebox{\linewidth}{!}{%
\begin{tabular}{lcccccccccccccccccccc}
\toprule
\multirow{2}{*}{\textbf{Method}} & \multicolumn{11}{c}{\textbf{In-Domain Languages}} & \multicolumn{8}{c}{\textbf{Out-of-Domain Languages}} & \multirow{2}{*}{\textbf{ALL-Avg.}} \\
\cmidrule(lr){2-12} \cmidrule(lr){13-20}
 & \textbf{Ar} & \textbf{Bn} & \textbf{Th} & \textbf{Sw} & \textbf{Ja} & \textbf{Zh} & \textbf{De} & \textbf{Fr} & \textbf{Ru} & \textbf{En} & \textbf{Avg.} & \textbf{Te} & \textbf{Ko} & \textbf{Vi} & \textbf{It} & \textbf{Id} & \textbf{Pt} & \textbf{Es} & \textbf{Avg.} & \\
\midrule

\rowcolor{gray!30}
\textit{\textbf{Llama3.1-8B-Instruct}} &
\phantom{0}5.9 & \phantom{0}5.6 & \phantom{0}5.9 & \phantom{0}5.1 & \phantom{0}6.9 & \phantom{0}4.3 & \phantom{0}5.8 & \phantom{0}7.0 & \phantom{0}6.3 & 10.3 & \phantom{0}6.3 & \phantom{0}5.0 & \phantom{0}6.2 &  \phantom{0}7.4 & \phantom{0}7.4 & \phantom{0}6.3 & \phantom{0}5.8 & \phantom{0}6.3 & \phantom{0}6.3 & \phantom{0}6.3 \\
&
\textcolor{upgray}{\phantom{0}5.9} & \textcolor{upgray}{\phantom{0}6.5} & \textcolor{upgray}{\phantom{0}7.1} & \textcolor{upgray}{\phantom{0}6.3} & \textcolor{upgray}{\phantom{0}6.9} & \textcolor{upgray}{\phantom{0}7.1} & \textcolor{upgray}{\phantom{0}6.6} & \textcolor{upgray}{\phantom{0}7.7} & \textcolor{upgray}{\phantom{0}7.3} & \textcolor{upgray}{10.3} & \textcolor{upgray}{\phantom{0}7.2} & \textcolor{upgray}{\phantom{0}5.2} & \textcolor{upgray}{\phantom{0}6.8} & \textcolor{upgray}{\phantom{0}7.5} & \textcolor{upgray}{\phantom{0}8.1} & \textcolor{upgray}{\phantom{0}7.1} & \textcolor{upgray}{\phantom{0}6.3} & \textcolor{upgray}{\phantom{0}6.3} & \textcolor{upgray}{\phantom{0}6.8} & \textcolor{upgray}{\phantom{0}7.0} \\
\bottomrule
\end{tabular}%
}

  \caption{The LC\&Acc (\%) on PolyMath test sets for Llama3.1-8B-Instruct and Qwen2.5-32B-Instruct.} 
  \label{table:polymath-LC-Acc-llama}
\end{table*}

\begin{table*}[t]
    \centering
    \resizebox{\linewidth}{!}{%
\begin{tabular}{lccccccccccccc}
\toprule
\multirow{2}{*}{\textbf{Method}} & \multicolumn{7}{c}{\textbf{In-Domain Languages}} & \multicolumn{5}{c}{\textbf{Out-of-Domain Languages}} & \multirow{2}{*}{\textbf{ALL-Avg.}} \\
\cmidrule(lr){2-8} \cmidrule(lr){9-13}
 & \textbf{Ar} & \textbf{Th} & \textbf{Fr} & \textbf{Ja} & \textbf{Zh} & \textbf{En} & \textbf{Avg.} & \textbf{Vi} & \textbf{Ko} & \textbf{Pt} & \textbf{Es} & \textbf{Avg.} & \\
\midrule

\rowcolor{gray!30}
\textit{\textbf{Qwen2.5-3B-Instruct}} & 13.1 & 11.0 & 16.7 & 13.7 & 15.1 & 23.0 & 15.4 & 15.9 & 13.0 & 18.5 & 16.6 & 16.0 & 15.7 \\
\hspace{2mm}+ LCP & 13.7 & 13.3 & 16.5 & 12.5 & 15.9 & 21.3 & 15.5 & 13.1 & 11.0 & 16.0 & 16.7 & 14.2 & 15.0 \\
\hspace{2mm}+ DIT & 12.6 & 12.6 & 17.0 & 12.7 & 17.4 & \textbf{23.2} & 15.9 & 14.9 & 12.5 & 17.8 & 17.9 & 15.8 & 15.8 \\
\hspace{2mm}+ QRT & 12.4 & 11.4 & 17.2 & 13.2 & 17.9 & 22.3 & 15.7 & 14.1 & 12.3 & 16.8 & 18.6 & 15.5 & 15.6 \\
\cdashline{1-14}
\hspace{2mm}+ M-SFT & 12.9 & 12.1 & 18.5 & 13.3 & 17.0 & 22.4 & 16.0 & 13.4 & 12.1 & 16.5 & 19.1 & 15.3 & 15.7 \\
\hspace{2mm}+ Vanilla GRPO & \textbf{18.6} & \textbf{15.7} & 19.5 & \textbf{18.5} & 19.4 & 22.4 & \textbf{19.0} & 16.0 & \textbf{20.4} & 20.1 & 19.1 & \textbf{18.9} & \textbf{19.0} \\
\bottomrule
\end{tabular}%
}

  \caption{The accuracy (\%) on MMATH test sets.}
  \label{table:mmath-acc}
\end{table*}

\begin{table*}[t]
    \centering
    \resizebox{\linewidth}{!}{%
\begin{tabular}{lccccccccccccccccccccc}
\toprule
\multirow{2}{*}{\textbf{Method}} & \multicolumn{11}{c}{\textbf{In-Domain Languages}} & \multicolumn{9}{c}{\textbf{Out-of-Domain Languages}} & \multirow{2}{*}{\textbf{ALL-Avg.}} \\
\cmidrule(lr){2-12} \cmidrule(lr){13-21}
 & \textbf{Ar} & \textbf{Bn} & \textbf{Th} & \textbf{Sw} & \textbf{Ja} & \textbf{Zh} & \textbf{De} & \textbf{Fr} & \textbf{Ru} & \textbf{En} & \textbf{Avg.} & \textbf{Te} & \textbf{Ko} & \textbf{Ms} & \textbf{Vi} & \textbf{It} & \textbf{Id} & \textbf{Pt} & \textbf{Es} & \textbf{Avg.} & \\
\midrule

\rowcolor{gray!30}
\textit{\textbf{Qwen2.5-3B-Instruct}} &
\phantom{0}8.4 & \phantom{0}4.5 & \phantom{0}7.6 & \phantom{0}1.3 & \phantom{0}8.6 & \phantom{0}9.4 & \phantom{0}9.3 & \phantom{0}9.6 & \phantom{0}8.9 & 11.4 & \phantom{0}7.9 &
\phantom{0}1.0 & \phantom{0}8.4 & \phantom{0}7.1 & \phantom{0}9.9 & 10.0 & \phantom{0}9.7 & 11.8 & 11.4 & \phantom{0}8.7 & \phantom{0}8.2 \\
\hspace{2mm}+ LCP &
\phantom{0}8.2 & \phantom{0}3.7 & \phantom{0}8.4 & \phantom{0}\textbf{3.4} & \phantom{0}6.8 & 11.0 & \phantom{0}9.9 & \phantom{0}9.3 & \phantom{0}8.9 & \textbf{14.1} & \phantom{0}8.4 &
\phantom{0}2.6 & \phantom{0}7.4 & \phantom{0}6.3 & \phantom{0}9.2 & \phantom{0}9.6 & \phantom{0}9.8 & 11.1 & \phantom{0}9.9 & \phantom{0}8.2 & \phantom{0}8.3 \\
\hspace{2mm}+ DIT &
\phantom{0}7.6 & \phantom{0}5.4 & \phantom{0}6.4 & \phantom{0}3.1 & \phantom{0}7.4 & 10.4 & \textbf{11.7} & \phantom{0}9.9 & 10.4 & 11.7 & \phantom{0}8.4 &
\phantom{0}1.0 & \phantom{0}8.2 & \phantom{0}9.4 & \phantom{0}8.8 & \phantom{0}9.6 & \phantom{0}9.4 & 10.5 & 10.2 & \phantom{0}8.4 & \phantom{0}8.4 \\
\hspace{2mm}+ QRT &
\textbf{10.8} & \phantom{0}3.2 & \phantom{0}6.4 & \phantom{0}1.6 & \phantom{0}9.2 & \phantom{0}7.7 & \phantom{0}9.6 & 11.0 & 10.8 & 12.7 & \phantom{0}8.3 &
\phantom{0}1.6 & \phantom{0}7.5 & \phantom{0}8.7 & 11.4 & 10.7 & 10.5 & \phantom{0}9.7 & \phantom{0}9.7 & \phantom{0}8.7 & \phantom{0}8.5 \\
\bottomrule
\end{tabular}%
}

  \caption{The accuracy (\%) on PolyMath test sets.}
  \label{table:polymath-acc}
\end{table*} 

\begin{table*}[t]
    \centering
    \resizebox{\linewidth}{!}{%
\begin{tabular}{lccccccccccccccccccccc}
\toprule
\multirow{2}{*}{\textbf{Method}} & \multicolumn{11}{c}{\textbf{In-Domain Languages}} & \multicolumn{9}{c}{\textbf{Out-of-Domain Languages}} & \multirow{2}{*}{\textbf{ALL-Avg.}} \\
\cmidrule(lr){2-12} \cmidrule(lr){13-21}
 & \textbf{Ar} & \textbf{Bn} & \textbf{Th} & \textbf{Sw} & \textbf{Ja} & \textbf{Zh} & \textbf{De} & \textbf{Fr} & \textbf{Ru} & \textbf{En} & \textbf{Avg.} & \textbf{Te} & \textbf{Ko} & \textbf{Ms} & \textbf{Vi} & \textbf{It} & \textbf{Id} & \textbf{Pt} & \textbf{Es} & \textbf{Avg.} & \\
\midrule

\rowcolor{gray!30}
\textit{\textbf{Qwen2.5-3B-Instruct}} &
\phantom{0}3.2 & \phantom{0}2.4 & \phantom{0}2.4 & \phantom{0}0.8 & \phantom{0}5.6 & \phantom{0}3.2 & \phantom{0}3.2 & \phantom{0}4.8 & \phantom{0}2.4 & \phantom{0}1.6 & \phantom{0}3.0 &
\phantom{0}0.0 & \phantom{0}2.4 & \phantom{0}2.4 & \phantom{0}4.8 & \phantom{0}4.8 & \phantom{0}4.0 & \textbf{\phantom{0}8.0} & \phantom{0}5.6 & \phantom{0}4.0 & \phantom{0}3.4 \\
\hspace{2mm}+ LCP &
\phantom{0}3.2 & \phantom{0}0.8 & \phantom{0}4.8 & \phantom{0}3.2 & \phantom{0}2.4 & \phantom{0}5.6 & \phantom{0}4.8 & \phantom{0}2.4 & \phantom{0}1.6 & \textbf{\phantom{0}6.4} & \phantom{0}3.5 &
\phantom{0}2.4 & \phantom{0}4.0 & \phantom{0}0.0 & \phantom{0}4.0 & \phantom{0}4.0 & \phantom{0}4.8 & \phantom{0}6.4 & \phantom{0}2.4 & \phantom{0}3.5 & \phantom{0}3.5 \\
\hspace{2mm}+ DIT &
\phantom{0}3.2 & \phantom{0}4.0 & \phantom{0}2.4 & \textbf{\phantom{0}3.2} & \phantom{0}1.6 & \phantom{0}4.8 & \textbf{\phantom{0}6.4} & \phantom{0}4.8 & \phantom{0}4.8 & \phantom{0}4.0 & \phantom{0}3.9 &
\phantom{0}0.0 & \phantom{0}4.0 & \phantom{0}4.8 & \phantom{0}2.4 & \phantom{0}1.6 & \phantom{0}4.0 & \phantom{0}4.8 & \phantom{0}4.0 & \phantom{0}3.2 & \phantom{0}3.6 \\
\bottomrule
\end{tabular}%
}

  \caption{The accuracy (\%) on PolyMath test sets of top-level difficulty.}
  \label{table:polymath-acc-top}
\end{table*} 

\begin{table*}[t]
    \centering
    \resizebox{\linewidth}{!}{%
\begin{tabular}{lccccccccccccccccccccc}
\toprule
\multirow{2}{*}{\textbf{Method}} & \multicolumn{11}{c}{\textbf{In-Domain Languages}} & \multicolumn{9}{c}{\textbf{Out-of-Domain Languages}} & \multirow{2}{*}{\textbf{ALL-Avg.}} \\
\cmidrule(lr){2-12} \cmidrule(lr){13-21}
 & \textbf{Ar} & \textbf{Bn} & \textbf{Th} & \textbf{Sw} & \textbf{Ja} & \textbf{Zh} & \textbf{De} & \textbf{Fr} & \textbf{Ru} & \textbf{En} & \textbf{Avg.} & \textbf{Te} & \textbf{Ko} & \textbf{Ms} & \textbf{Vi} & \textbf{It} & \textbf{Id} & \textbf{Pt} & \textbf{Es} & \textbf{Avg.} & \\
\midrule
\rowcolor{gray!30}
\textit{\textbf{Qwen2.5-3B-Instruct}} & \phantom{0}3.2 & \phantom{0}0.8 & \phantom{0}4.8 & 0.8 & \phantom{0}4.0 & \phantom{0}3.2 & \phantom{0}6.4 & \phantom{0}3.2 & \phantom{0}5.6 & \phantom{0}8.0 & \phantom{0}4.0 & 0.8 & \phantom{0}6.4 & \phantom{0}2.4 & \phantom{0}4.8 & \phantom{0}4.0 & \phantom{0}4.8 & \phantom{0}2.4 & \phantom{0}4.8 & \phantom{0}3.0 & \phantom{0}3.9 \\
\hspace{2mm}+ LCP & \phantom{0}3.2 & \phantom{0}1.6 & \phantom{0}4.8 & 2.4 & \phantom{0}4.0 & \phantom{0}5.6 & \phantom{0}6.4 & \phantom{0}7.2 & \phantom{0}5.6 & \phantom{0}8.0 & \phantom{0}4.9 & 1.6 & \phantom{0}2.4 & \phantom{0}3.2 & \phantom{0}3.2 & \phantom{0}4.0 & \phantom{0}4.8 & \phantom{0}3.2 & \phantom{0}6.4 & \phantom{0}3.6 & \phantom{0}4.3 \\
\hspace{2mm}+ DIT & \phantom{0}4.0 & \phantom{0}0.0 & \phantom{0}1.6 & 2.4 & \phantom{0}6.4 & \phantom{0}4.8 & \phantom{0}8.8 & \phantom{0}4.0 & \phantom{0}4.0 & \phantom{0}4.8 & \phantom{0}4.1 & 0.8 & \phantom{0}4.0 & \phantom{0}4.8 & \phantom{0}4.8 & \phantom{0}8.8 & \phantom{0}4.8 & \phantom{0}5.6 & \phantom{0}5.6 & \phantom{0}4.9 & \phantom{0}4.4 \\
\bottomrule
\end{tabular}%
}
  \caption{The accuracy (\%) on PolyMath test sets of high-level difficulty.}
  \label{table:polymath-acc-high}
\end{table*}

\begin{table*}[t]
    \centering
    \resizebox{\linewidth}{!}{%
\begin{tabular}{lccccccccccccccccccccc}
\toprule
\multirow{2}{*}{\textbf{Method}} & \multicolumn{11}{c}{\textbf{In-Domain Languages}} & \multicolumn{9}{c}{\textbf{Out-of-Domain Languages}} & \multirow{2}{*}{\textbf{ALL-Avg.}} \\
\cmidrule(lr){2-12} \cmidrule(lr){13-21}
 & \textbf{Ar} & \textbf{Bn} & \textbf{Th} & \textbf{Sw} & \textbf{Ja} & \textbf{Zh} & \textbf{De} & \textbf{Fr} & \textbf{Ru} & \textbf{En} & \textbf{Avg.} & \textbf{Te} & \textbf{Ko} & \textbf{Ms} & \textbf{Vi} & \textbf{It} & \textbf{Id} & \textbf{Pt} & \textbf{Es} & \textbf{Avg.} & \\
\midrule
\rowcolor{gray!30}
\textit{\textbf{Qwen2.5-3B-Instruct}} & 13.6 & \phantom{0}5.6 & 11.2 & \phantom{0}2.4 & \phantom{0}6.4 & 14.4 & 13.6 & \phantom{0}9.6 & 11.2 & 19.2 & 10.7 & \phantom{0}1.6 & \phantom{0}9.6 & \phantom{0}8.8 & 12.0 & 12.8 & 12.8 & 14.4 & 16.0 & 11.0 & 10.8 \\
\hspace{2mm}+ LCP & 13.6 & \phantom{0}4.0 & \phantom{0}8.8 & \phantom{0}4.8 & \phantom{0}8.0 & 13.6 & 11.2 & 13.6 & 15.2 & 22.4 & 11.5 & \phantom{0}2.4 & \phantom{0}8.0 & 10.4 & 12.8 & 11.2 & 10.4 & 16.0 & 14.4 & 10.7 & 11.2 \\
\hspace{2mm}+ DIT & \phantom{0}9.6 & \phantom{0}8.0 & \phantom{0}8.8 & \phantom{0}2.4 & \phantom{0}8.8 & 13.6 & 13.6 & 12.0 & 12.8 & 18.4 & 10.8 & \phantom{0}2.4 & \phantom{0}8.0 & 12.0 & 12.0 & 12.0 & 11.2 & \phantom{0}8.8 & 13.6 & 10.0 & 10.4 \\
\hspace{2mm}+ QRT & 12.8 & \phantom{0}1.6 & \phantom{0}9.6 & \phantom{0}4.0 & \phantom{0}8.8 & 12.8 & 12.8 & 16.0 & 12.0 & 21.6 & 11.2 & \phantom{0}1.6 & 13.6 & 13.6 & 12.8 & 11.2 & 16.0 & 13.6 & 14.4 & 12.1 & 11.6 \\
\cdashline{1-22}
\hspace{2mm}+ M-SFT & 12.8 & \phantom{0}3.2 & \phantom{0}8.8 & \phantom{0}3.2 & \phantom{0}5.6 & 12.8 & 12.8 & 10.4 & 10.4 & 22.4 & 10.2 & \phantom{0}0.8 & \phantom{0}8.8 & \phantom{0}6.4 & 12.8 & 12.0 & 12.8 & 12.0 & 12.8 & \phantom{0}9.8 & 10.0 \\
\bottomrule
\end{tabular}%
}
  \caption{The accuracy (\%) on PolyMath test sets of medium-level difficulty.}
  \label{table:polymath-acc-medium}
\end{table*}

\begin{table*}[t]
    \centering
    \resizebox{\linewidth}{!}{%
\begin{tabular}{lccccccccccccccccccccc}
\toprule
\multirow{2}{*}{\textbf{Method}} & \multicolumn{11}{c}{\textbf{In-Domain Languages}} & \multicolumn{9}{c}{\textbf{Out-of-Domain Languages}} & \multirow{2}{*}{\textbf{ALL-Avg.}} \\
\cmidrule(lr){2-12} \cmidrule(lr){13-21}
 & \textbf{Ar} & \textbf{Bn} & \textbf{Th} & \textbf{Sw} & \textbf{Ja} & \textbf{Zh} & \textbf{De} & \textbf{Fr} & \textbf{Ru} & \textbf{En} & \textbf{Avg.} & \textbf{Te} & \textbf{Ko} & \textbf{Ms} & \textbf{Vi} & \textbf{It} & \textbf{Id} & \textbf{Pt} & \textbf{Es} & \textbf{Avg.} & \\
\midrule
\rowcolor{gray!30}
\textit{\textbf{Qwen2.5-3B-Instruct}} & 60.8 & 33.6 & 52.8 & \phantom{0}5.6 & 55.2 & 73.6 & 61.6 & 73.6 & 68.8 & 87.2 & 57.3 & \phantom{0}8.0 & 61.6 & 60.0 & 66.4 & 69.6 & 68.0 & 74.4 & 74.4 & 60.3 & 58.6 \\
\hspace{2mm}+ LCP & 56.8 & 35.2 & 51.2 & \phantom{0}6.4 & 51.2 & 71.2 & 62.4 & 64.8 & 68.0 & 84.0 & 55.1 & \phantom{0}8.8 & 52.8 & 60.8 & 67.2 & 72.8 & 68.0 & 70.4 & 75.2 & 59.5 & 57.1 \\
\hspace{2mm}+ DIT & 52.8 & 33.6 & 52.8 & \phantom{0}6.4 & 55.2 & 71.2 & 61.6 & 70.4 & 76.0 & 87.2 & 56.7 & \phantom{0}6.4 & 58.4 & 60.0 & 69.6 & 72.0 & 68.0 & \textbf{78.4} & 72.0 & 60.6 & 58.4 \\
\hspace{2mm}+ QRT & \textbf{68.8} & 32.0 & 51.2 & \phantom{0}5.6 & 53.6 & 70.4 & 57.6 & 65.6 & 71.2 & 87.2 & 56.3 & 10.4 & 59.2 & 64.8 & 68.8 & 73.6 & 70.4 & 70.4 & 77.6 & 61.9 & 58.8 \\
\cdashline{1-22}
\hspace{2mm}+ M-SFT & 54.4 & 39.2 & 56.0 & \phantom{0}3.2 & 48.8 & \textbf{75.2} & 60.8 & 68.8 & 72.0 & 84.8 & 56.3 & \phantom{0}7.2 & 55.2 & 64.8 & 62.4 & 71.2 & 66.4 & 74.4 & 76.0 & 59.7 & 57.8 \\
\hspace{2mm}+ Vanilla GRPO & 65.6 & \textbf{47.2} & 59.2 & \textbf{11.2} & 56.8 & 74.4 & 64.0 & 74.4 & 74.4 & 86.4 & 61.4 & \phantom{0}5.6 & 59.2 & 62.4 & \textbf{72.0} & 73.6 & \textbf{73.6} & 75.2 & \textbf{79.2} & 62.6 & 61.9 \\
\bottomrule
\end{tabular}%
}

  \caption{The accuracy (\%) on PolyMath test sets of low-level difficulty.}
  \label{table:polymath-acc-low}
\end{table*}

\begin{table*}[t]
    \centering
    \resizebox{\linewidth}{!}{%
\begin{tabular}{lcccccccccccccccccccc}
\toprule
\multirow{2}{*}{\textbf{Method}} & \multicolumn{11}{c}{\textbf{In-Domain Languages}} & \multicolumn{8}{c}{\textbf{Out-of-Domain Languages}} & \multirow{2}{*}{\textbf{ALL-Avg.}} \\
\cmidrule(lr){2-12} \cmidrule(lr){13-20}
 & \textbf{Ar} & \textbf{Bn} & \textbf{Th} & \textbf{Sw} & \textbf{Ja} & \textbf{Zh} & \textbf{De} & \textbf{Fr} & \textbf{Ru} & \textbf{En} & \textbf{Avg.} & \textbf{Te} & \textbf{Ko} & \textbf{Vi} & \textbf{It} & \textbf{Id} & \textbf{Pt} & \textbf{Es} & \textbf{Avg.} & \\
\midrule

\rowcolor{gray!30}
\textit{\textbf{Qwen2.5-3B-Instruct}} & 3.2 & 2.4 & 2.4 & 0.8 & 5.6 & \phantom{0}3.2 & \phantom{0}3.2 & 4.8 & \phantom{0}2.4 & 1.6 & 3.0 & 0.0 & 2.4 & 4.8 & \phantom{0}4.0 & \phantom{0}4.0 & \textbf{8.0} & \phantom{0}5.6 & 4.1 & 3.4 \\
\hspace{2mm}+ LCP & 3.2 & 0.8 & 4.8 & 3.2 & 2.4 & \phantom{0}5.6 & \phantom{0}4.8 & 2.4 & \phantom{0}1.6 & \textbf{6.4} & 3.5 & 2.4 & 4.0 & 4.0 & \phantom{0}4.0 & \phantom{0}4.8 & 6.4 & \phantom{0}2.4 & 4.0 & 3.7 \\
\hspace{2mm}+ DIT & 3.2 & 4.0 & 2.4 & \textbf{3.2} & 1.6 & \phantom{0}4.8 & \textbf{\phantom{0}6.4} & 4.8 & \phantom{0}4.8 & 4.0 & 3.9 & 0.0 & 4.0 & 2.4 & \phantom{0}1.6 & \phantom{0}4.0 & 4.8 & \phantom{0}4.0 & 3.0 & 3.5 \\
\hspace{2mm}+ QRT & \textbf{6.4} & 1.6 & 2.4 & 0.8 & 4.8 & \phantom{0}1.6 & \phantom{0}5.6 & 4.8 & \textbf{\phantom{0}6.4} & 4.0 & 3.8 & 0.8 & 0.8 & 6.4 & \textbf{\phantom{0}6.4} & \phantom{0}4.8 & 3.2 & \phantom{0}3.2 & 3.7 & 3.8 \\
\cdashline{1-21}
\hspace{2mm}+ M-SFT & 1.6 & 2.4 & 1.6 & 0.8 & 1.6 & \phantom{0}3.2 & \phantom{0}0.8 & 1.6 & \phantom{0}4.0 & 4.0 & 2.2 & 0.0 & 1.6 & 0.8 & \phantom{0}1.6 & \phantom{0}0.0 & 1.6 & \phantom{0}1.6 & 1.0 & 1.7 \\
\hspace{2mm}+ Vanilla GRPO & 4.0 & 0.0 & 2.4 & 0.0 & 4.0 & \phantom{0}5.6 & \phantom{0}3.2 & 4.8 & \phantom{0}3.2 & 4.8 & 3.2 & 1.6 & \textbf{4.8} & 4.0 & \phantom{0}0.8 & \textbf{\phantom{0}7.2} & 4.0 & \textbf{\phantom{0}5.6} & 4.0 & 3.5 \\
\bottomrule
\end{tabular}%
}
  \caption{The LC\&Acc (\%) on PolyMath test sets of top-level difficulty.}
  \label{table:fffn}
\end{table*}

\begin{table*}[t]
    \centering
    \resizebox{\linewidth}{!}{%
\begin{tabular}{lcccccccccccccccccccc}
\toprule
\multirow{2}{*}{\textbf{Method}} & \multicolumn{11}{c}{\textbf{In-Domain Languages}} & \multicolumn{8}{c}{\textbf{Out-of-Domain Languages}} & \multirow{2}{*}{\textbf{ALL-Avg.}} \\
\cmidrule(lr){2-12} \cmidrule(lr){13-20}
 & \textbf{Ar} & \textbf{Bn} & \textbf{Th} & \textbf{Sw} & \textbf{Ja} & \textbf{Zh} & \textbf{De} & \textbf{Fr} & \textbf{Ru} & \textbf{En} & \textbf{Avg.} & \textbf{Te} & \textbf{Ko} & \textbf{Vi} & \textbf{It} & \textbf{Id} & \textbf{Pt} & \textbf{Es} & \textbf{Avg.} & \\
\midrule

\rowcolor{gray!30}
\textit{\textbf{Qwen2.5-3B-Instruct}} & \phantom{0}3.2 & 0.8 & \phantom{0}4.8 & 0.8 & \phantom{0}4.0 & 3.2 & \phantom{0}6.4 & \phantom{0}3.2 & \phantom{0}5.6 & \phantom{0}8.0 & \phantom{0}4.0 & 0.8 & \phantom{0}6.4 & \phantom{0}4.8 & \phantom{0}4.0 & 4.8 & \phantom{0}2.4 & \phantom{0}4.8 & \phantom{0}4.0 & \phantom{0}4.0 \\
\hspace{2mm}+ LCP & \phantom{0}3.2 & 1.6 & \phantom{0}4.8 & 2.4 & \phantom{0}4.0 & 5.6 & \phantom{0}6.4 & \phantom{0}7.2 & \phantom{0}5.6 & \phantom{0}8.0 & \phantom{0}4.9 & 0.8 & \phantom{0}2.4 & \phantom{0}3.2 & \phantom{0}4.0 & 4.8 & \phantom{0}3.2 & \phantom{0}6.4 & \phantom{0}3.5 & \phantom{0}4.3 \\
\hspace{2mm}+ DIT & \phantom{0}4.0 & 0.0 & \phantom{0}1.6 & \textbf{2.4} & \phantom{0}6.4 & 4.8 & \phantom{0}8.8 & \phantom{0}4.0 & \phantom{0}4.0 & \phantom{0}4.8 & \phantom{0}4.1 & 0.8 & \phantom{0}4.0 & \phantom{0}4.8 & \phantom{0}8.8 & 4.0 & \phantom{0}5.6 & \phantom{0}5.6 & \phantom{0}4.8 & \phantom{0}4.4 \\
\hspace{2mm}+ QRT & \phantom{0}4.0 & 0.0 & \phantom{0}1.6 & 0.8 & \textbf{\phantom{0}7.2} & 1.6 & \phantom{0}4.0 & \textbf{\phantom{0}7.2} & \phantom{0}4.0 & \phantom{0}7.2 & \phantom{0}3.8 & 0.8 & \phantom{0}4.8 & \phantom{0}6.4 & \phantom{0}3.2 & 4.0 & \phantom{0}5.6 & \phantom{0}3.2 & \phantom{0}4.0 & \phantom{0}3.9 \\
\bottomrule
\end{tabular}%
}
  \caption{The LC\&Acc (\%) on PolyMath test sets of high-level difficulty.}
  \label{table:fffn}
\end{table*}

\begin{table*}[t]
    \centering
    \resizebox{\linewidth}{!}{%
\begin{tabular}{lcccccccccccccccccccc}
\toprule
\multirow{2}{*}{\textbf{Method}} & \multicolumn{11}{c}{\textbf{In-Domain Languages}} & \multicolumn{8}{c}{\textbf{Out-of-Domain Languages}} & \multirow{2}{*}{\textbf{ALL-Avg.}} \\
\cmidrule(lr){2-12} \cmidrule(lr){13-20}
 & \textbf{Ar} & \textbf{Bn} & \textbf{Th} & \textbf{Sw} & \textbf{Ja} & \textbf{Zh} & \textbf{De} & \textbf{Fr} & \textbf{Ru} & \textbf{En} & \textbf{Avg.} & \textbf{Te} & \textbf{Ko} & \textbf{Vi} & \textbf{It} & \textbf{Id} & \textbf{Pt} & \textbf{Es} & \textbf{Avg.} & \\
\midrule

\rowcolor{gray!30}
\textit{\textbf{Qwen2.5-3B-Instruct}} & 12.8 & \phantom{0}5.6 & 10.4 & 2.4 & \phantom{0}5.6 & 12.8 & 13.6 & \phantom{0}9.6 & 11.2 & 19.2 & 10.3 & \phantom{0}1.6 & \phantom{0}9.6 & 12.0 & 12.8 & 12.8 & 14.4 & 15.2 & 11.2 & 10.7 \\
\hspace{2mm}+ LCP & \textbf{13.6} & \phantom{0}3.2 & \phantom{0}8.8 & 4.8 & \phantom{0}8.0 & 13.6 & 11.2 & 13.6 & 15.2 & 22.4 & 11.4 & \phantom{0}2.4 & \phantom{0}8.0 & 12.8 & 11.2 & 10.4 & 16.0 & 14.4 & 10.7 & 11.2 \\
\hspace{2mm}+ DIT & \phantom{0}9.6 & \phantom{0}7.2 & \phantom{0}8.8 & 2.4 & \phantom{0}8.8 & 12.0 & 13.6 & 12.0 & 12.8 & 18.4 & 10.6 & \phantom{0}2.4 & \phantom{0}8.0 & 12.0 & 12.0 & 11.2 & \phantom{0}8.8 & 13.6 & \phantom{0}9.7 & 10.2 \\
\hspace{2mm}+ QRT & 12.8 & \phantom{0}1.6 & \phantom{0}9.6 & 4.0 & \phantom{0}8.8 & 12.8 & 12.8 & 16.0 & 12.0 & 21.6 & 11.2 & \phantom{0}1.6 & 13.6 & 12.8 & 11.2 & 16.0 & 13.6 & 14.4 & 11.9 & 11.5 \\
\cdashline{1-21}
\hspace{2mm}+ M-SFT & \phantom{0}0.8 & \phantom{0}3.2 & \phantom{0}3.2 & 1.6 & \phantom{0}4.8 & \phantom{0}4.0 & \phantom{0}4.8 & \phantom{0}2.4 & \phantom{0}6.4 & \phantom{0}3.2 & \phantom{0}3.4 & \phantom{0}0.8 & \phantom{0}1.6 & \phantom{0}6.4 & \phantom{0}2.4 & \phantom{0}1.6 & \phantom{0}4.0 & \phantom{0}6.4 & \phantom{0}3.3 & \phantom{0}3.4 \\
\bottomrule
\end{tabular}%
}
  \caption{The LC\&Acc (\%) on PolyMath test sets of medium-level difficulty.}
  \label{table:fffn}
\end{table*}

\begin{table*}[t]
    \centering
    \resizebox{\linewidth}{!}{%
\begin{tabular}{lcccccccccccccccccccc}
\toprule
\multirow{2}{*}{\textbf{Method}} & \multicolumn{11}{c}{\textbf{In-Domain Languages}} & \multicolumn{8}{c}{\textbf{Out-of-Domain Languages}} & \multirow{2}{*}{\textbf{ALL-Avg.}} \\
\cmidrule(lr){2-12} \cmidrule(lr){13-20}
 & \textbf{Ar} & \textbf{Bn} & \textbf{Th} & \textbf{Sw} & \textbf{Ja} & \textbf{Zh} & \textbf{De} & \textbf{Fr} & \textbf{Ru} & \textbf{En} & \textbf{Avg.} & \textbf{Te} & \textbf{Ko} & \textbf{Vi} & \textbf{It} & \textbf{Id} & \textbf{Pt} & \textbf{Es} & \textbf{Avg.} & \\
\midrule

\rowcolor{gray!30}
\textit{\textbf{Qwen2.5-3B-Instruct}} & 60.0 & 33.6 & 52.8 & \phantom{0}5.6 & 55.2 & 55.2 & 60.0 & 73.6 & 67.2 & 87.2 & 55.0 & \phantom{0}8.0 & 61.6 & 66.4 & 68.8 & 67.2 & 68.0 & 73.6 & 59.1 & 56.7 \\
\hspace{2mm}+ LCP & 56.8 & 35.2 & 51.2 & \phantom{0}6.4 & 51.2 & 58.4 & 62.4 & 63.2 & 66.4 & 84.0 & 53.5 & \phantom{0}8.8 & 52.8 & 67.2 & 72.8 & 68.0 & 68.0 & 75.2 & 59.0 & 55.8 \\
\hspace{2mm}+ DIT & 52.8 & 33.6 & 52.8 & \phantom{0}6.4 & 54.4 & 56.0 & 61.6 & 69.6 & 75.2 & 87.2 & 55.0 & \phantom{0}6.4 & 58.4 & 69.6 & 71.2 & 68.0 & 74.4 & 71.2 & 59.9 & 57.0 \\
\hspace{2mm}+ QRT & \textbf{68.8} & 32.0 & 50.4 & \phantom{0}5.6 & 53.6 & 58.4 & 56.0 & 64.8 & 71.2 & 87.2 & 54.8 & 10.4 & 59.2 & 68.8 & 72.8 & 70.4 & 66.4 & \textbf{77.6} & 60.8 & 57.3 \\
\cdashline{1-21}
\hspace{2mm}+ M-SFT & 23.2 & 12.8 & 32.8 & \phantom{0}0.8 & 12.0 & 22.4 & 22.4 & 19.2 & 30.4 & 22.4 & 19.8 & \phantom{0}0.8 & 48.8 & 52.0 & 22.4 & 27.2 & 28.0 & 46.4 & 32.2 & 24.9 \\
\hspace{2mm}+ Vanilla GRPO & 65.6 & \phantom{0}4.8 & 59.2 & \textbf{10.4} & 56.8 & 61.6 & 62.4 & 74.4 & 73.6 & 86.4 & 55.5 & \phantom{0}5.6 & 59.2 & \textbf{72.0} & 72.8 & \textbf{73.6} & 72.8 & 79.2 & 62.2 & 58.3 \\
\bottomrule
\end{tabular}%
}
  \caption{The LC\&Acc (\%) on PolyMath test sets of low-level difficulty.}
  \label{table:fffn}
\end{table*}

\begin{table*}[t]
    \centering
    \begin{tabular}{l c c c c c c c}
\toprule
\multirow{2}{*}{\textbf{Method}}
& \multicolumn{7}{c}{\textbf{XWinograd}} \\
\cmidrule(lr){2-8}
& pt & ru & fr & jp & zh & en & Avg. \\
\midrule
\textit{\textbf{Qwen2.5-7B-Instruct}} 
& 68.8 & 61.6 & 71.1 & 61.5 & 61.3 & 68.4 & 65.4 \\
\hspace{2mm}+ Vanilla GRPO
& 64.6 & 57.7 & 66.3 & 61.7 & 57.1 & 64.2 & 61.9 \\
\hspace{2mm}+ LC-GRPO
& 63.9 & 59.1 & 68.7 & 60.8 & 54.6 & 66.5 & 62.3 \\
\hspace{2mm}+ \ours
& \textbf{79.1} & \textbf{70.8} & \textbf{72.3} & \textbf{72.1} & \textbf{81.0} & \textbf{84.5} & \textbf{76.6} \\
\bottomrule
\end{tabular}
  \caption{The accuracy (\%) for all languages on the XWinograd test sets.}
  \label{table:Xwinograd}
\end{table*}

\begin{table*}[t]
    \centering
    \resizebox{\linewidth}{!}{%
\begin{tabular}{l c c c c c c c c c c c c}
\toprule
\multirow{2}{*}{\textbf{Method}}
& \multicolumn{12}{c}{\textbf{XStoryCloze}} \\
\cmidrule(lr){2-13}
& eu & my & sw & te & hi & id & ar & es & ru & zh & en & Avg. \\
\midrule
\textit{\textbf{Qwen2.5-7B-Instruct}}
& 49.8 & 49.6 & 53.0 & 55.9 & 58.8 & 64.3 & 63.6 & 67.2 & 68.5 & 66.2 & 72.5 & 60.9 \\
\hspace{2mm}+ Vanilla GRPO
& 50.2 & 49.2 & 51.0 & 56.3 & 56.6 & 58.1 & 60.6 & 61.2 & 63.4 & 60.4 & 65.9 & 57.5 \\
\hspace{2mm}+ LC-GRPO
& \textbf{52.1} & 49.4 & \textbf{52.5} & \textbf{56.3} & 58.6 & 63.7 & 62.2 & 64.7 & 67.6 & 64.7 & 70.2 & 60.2 \\
\hspace{2mm}+ \ours
& 51.0 & \textbf{50.7} & 52.3 & 56.2 & \textbf{61.5} & \textbf{68.2} & \textbf{64.0} & \textbf{70.9} & \textbf{71.5} & \textbf{71.5} & \textbf{80.9} & \textbf{63.5} \\
\bottomrule
\end{tabular}
}
  \caption{The accuracy (\%) for all languages on the XStoryCloze test sets.}
  \label{table:XStoryCloze}
\end{table*}

\begin{table*}[t]
    \centering
    \resizebox{\linewidth}{!}{%
\begin{tabular}{l c c c c c c c c c c c c}
\toprule
\multirow{2}{*}{\textbf{Method}}
& \multicolumn{12}{c}{\textbf{XCOPA}} \\
\cmidrule(lr){2-13}
& et & ht & qu & sw & ta & id & th & tr & vi & it & zh & Avg. \\
\midrule
\textit{\textbf{Qwen2.5-7B-Instruct}}
& \textbf{53.6} & 50.2 & 50.6 & \textbf{54.0} & 54.2 & 69.2 & 62.6 & 61.4 & 68.8 & 72.2 & 71.2 & 60.7 \\
\hspace{2mm}+ Vanilla GRPO
& 51.2 & 49.6 & 49.6 & 52.6 & \textbf{56.0} & 66.0 & 61.8 & 59.2 & 65.4 & 67.2 & 69.2 & 58.9 \\
\hspace{2mm}+ LC-GRPO
& 51.4 & 51.4 & \textbf{50.8} & 50.6 & 55.6 & 69.2 & \textbf{63.2} & 61.4 & 69.0 & 71.0 & 71.4 & 60.5 \\
\hspace{2mm}+ \ours
& 50.6 & \textbf{53.0} & 49.4 & 52.6 & 55.4 & \textbf{74.0} & 61.0 & \textbf{62.0} & \textbf{78.2} & \textbf{75.6} & \textbf{80.2} & \textbf{62.9} \\
\bottomrule
\end{tabular}
}
  \caption{The accuracy (\%) for all languages on the XCOPA test sets.}
  \label{table:XCOPA}
\end{table*}

\begin{table*}[t]
    \centering
    \begin{tabular}{l *{6}{c}}
\toprule
\multirow{2}{*}{\textbf{Method}} & \multicolumn{6}{c}{\textbf{MMLU-ProX}} \\
\cmidrule(lr){2-7}
& zu & yo & wo & ur & te & ne \\
\midrule
\textit{\textbf{Qwen2.5-7B-Instruct}} & \phantom{0}8.7 & \phantom{0}6.4 & 16.5 & 10.9 & 25.7 & 19.5 \\
\hspace{2mm}+ Vanilla GRPO           & \phantom{0}0.1 & \phantom{0}0.0 & \phantom{0}0.1  & \phantom{0}1.1  & \phantom{0}0.2  & \phantom{0}0.0  \\
\hspace{2mm}+ LC-GRPO                & 10.4& 19.2& 10.9 & 30.9 & 22.4 & 28.5 \\
\hspace{2mm}+ \ours                  & \textbf{15.7}& \textbf{20.9}& \textbf{21.3} & \textbf{34.5} & \textbf{26.4} & \textbf{31.4} \\
\midrule

\multirow{2}{*}{\textbf{Method}} & \multicolumn{6}{c}{\textbf{MMLU-ProX}} \\
\cmidrule(lr){2-7}
& mr & sw & bn & af & sr & uk \\
\bottomrule
\end{tabular}

  \caption{The extract match (EM) (\%) for all languages on the MMLU-ProX test sets.}
  \label{table:MMLU-ProX}
\end{table*}

\begin{figure*}[htbp]
    \centering
    \begin{tikzpicture}
\begin{axis}[
    width=\linewidth,
    height=7cm,
    ymin=0, ymax=101,
    ytick={0,10,20,30,40,50,60,70,80,90,100},
    ytick distance=10,
    grid=major,
    major grid style={dotted},
    ylabel={Language Consistency Ratio},
    xlabel={Language Code},
    symbolic x coords={Ar,Th,Fr,Ja,Zh,En,Vi,Ko,Pt,Es},
    xtick=data,
    xticklabels={
    {\strut Ar},
    {\strut Th},
    {\strut Fr},
    {\strut Ja},
     {\strut Zh},
  {\strut En},
 {\strut Vi},
  {\strut Ko},
  {\strut Pt},
  {\strut Es},
},
    xticklabel style={anchor=north},
    enlarge x limits=0.02,
    legend style={
      draw=none,
      line width=1.5pt,
      at={(0.5,1.05)},
      anchor=south,
      legend columns=5,
      column sep=3pt,
      font=\footnotesize,
    },
]

\addplot[
  sharp plot,
  orange!65,
  thick, line width=1.5pt,
  mark=pentagon*, mark size=3pt,
  mark options={fill=white,draw=orange!65,line width=1.5pt}
] coordinates {
  (Ar,1.2) (Th,1.4) (Fr,2.2) (Ja,5.1) (Zh,82.0)
(En,99.1) (Vi,3.3) (Ko,2.5) (Pt,3.6) (Es,4.9)
};
\addlegendentry{Qwen2.5-3B-Instruct}

\addplot[
  sharp plot,
  ublue!80,
  thick, line width=1.5pt,
  mark=square*, mark size=2.6pt,
  mark options={fill=white,draw=ublue!80,line width=1.5pt}
] coordinates {
  (Ar,99.4) (Th,99.5) (Fr,99.9) (Ja,99.1) (Zh,91.6)
  (En,99.7) (Vi,100.0) (Ko,99.2) (Pt,98.9) (Es,99.5)
};
\addlegendentry{LCP}

\addplot[
  sharp plot,
  udporange!90,
  thick, line width=1.5pt,
  mark=*, mark size=3pt,
  mark options={fill=white,draw=udporange!90,line width=1.5pt}
] coordinates {
  (Ar,99.3) (Th,98.5) (Fr,99.5) (Ja,96.2) (Zh,89.9)
  (En,99.9) (Vi,99.9) (Ko,98.9) (Pt,99.4) (Es,98.7)
};
\addlegendentry{DIT}

\addplot[
  sharp plot,
  usemiblue!65,
  thick, line width=1.5pt,
  mark=triangle*, mark size=3pt,
  mark options={fill=white,draw=usemiblue!65,line width=1.5pt}
] coordinates {
  (Ar,99.2) (Th,99.7) (Fr,99.8) (Ja,99.3) (Zh,92.2)
  (En,99.8) (Vi,99.9) (Ko,99.2) (Pt,99.0) (Es,96.1)
};
\addlegendentry{QRT}

\addplot[
  sharp plot,
  udark!65,
  thick, line width=1.5pt,
  mark=pentagon*, mark size=3pt,
  mark options={fill=white,draw=udark!65,line width=1.5pt}
] coordinates {
  (Ar,79.0) (Th,88.0) (Fr,99.9) (Ja,98.4) (Zh,90.5)
  (En,99.7) (Vi,78.5) (Ko,88.6) (Pt,98.9) (Es,89.2)
};
\addlegendentry{M-SFT}

\addplot[
  sharp plot,
  usemidark!65,
  thick,
  line width=1.5pt,
  mark=*,
  mark size=3pt,
  mark options={fill=white,draw=usemidark!65,line width=1.5pt}
] coordinates {
  (Ar,17) (Th,62.8) (Fr,93.9) (Ja,36) (Zh,88.5)
  (En,100) (Vi,93.2) (Ko,5.7) (Pt,53.4) (Es,92.4)
};
\addlegendentry{Vanilla GRPO}

\addplot[
  sharp plot,
  udpdark!65,
  thick, line width=1.5pt,
  mark=triangle*, mark size=2.8pt,
  mark options={fill=white,draw=udpdark!65,line width=1.5pt}
] coordinates {
  (Ar,99.9) (Th,99.4) (Fr,99.9) (Ja,99.8) (Zh,87.0)
  (En,99.9) (Vi,100.0) (Ko,99.3) (Pt,99.7) (Es,99.8)
};
\addlegendentry{LC-GRPO}

\addplot[
  sharp plot,
  upurple!75,
  thick, line width=1.5pt,
  mark=square*, mark size=2.8pt,
  mark options={fill=white,draw=upurple!75,line width=1.5pt}
] coordinates {
  (Ar,99.4) (Th,97.8) (Fr,99.4) (Ja,98.7) (Zh,90.3)
  (En,99.7) (Vi,98.0) (Ko,99.1) (Pt,98.9) (Es,98.9)
};
\addlegendentry{M-Thinker}

\addplot[
  sharp plot,
  udpblue!40,
  thick, line width=1.5pt,
  mark=pentagon*, mark size=3pt,
  mark options={fill=white,draw=udpblue!40,line width=1.5pt}
] coordinates {
  (Ar,40.5) (Th,50.4) (Fr,70.7) (Ja,88.9) (Zh,77.0)
  (En,99.7) (Vi,60.3) (Ko,80.7) (Pt,85.3) (Es,88.6)
};
\addlegendentry{mGRPO}

\addplot[
  sharp plot,
  red!90,
  thick, line width=1.5pt,
  mark=oplus, mark size=3pt,
  mark options={fill=white,draw=red!90,line width=1.5pt}
] coordinates {
  (Ar,99.0) (Th,99.2) (Fr,99.5) (Ja,99.3) (Zh,92.6)
  (En,99.9) (Vi,99.7) (Ko,99.4) (Pt,99.0) (Es,99.4)
};
\addlegendentry{\ours}

\end{axis}
\end{tikzpicture}
  \caption{The input-output language consistency ratio of different methods for each language on MMATH test sets with Qwen2.5-3B-Instruct model.}
  \label{table:mmath-3B-LC}
\end{figure*}

\begin{figure*}[htbp]
    \centering
    \begin{tikzpicture}
\begin{axis}[
    width=\linewidth,
    height=7cm,
    ymin=0, ymax=103,
    ytick={0,10,20,30,40,50,60,70,80,90,100},
    ytick distance=10,
    grid=major,
    major grid style={dotted},
    ylabel={Language Consistency Ratio},
    xlabel={Language Code},
    symbolic x coords={ar,th,fr,ja,zh,en,vi,ko,pt,es},
    xtick=data,
    enlarge x limits=0.02,
    xticklabels={
    {\strut Ar},
    {\strut Th},
    {\strut Fr},
    {\strut Ja},
     {\strut Zh},
  {\strut En},
 {\strut Vi},
  {\strut Ko},
  {\strut Pt},
  {\strut Es},
},
xticklabel style={anchor=north},
    legend style={
      draw=none,
      line width=1.5pt,
      at={(0.5,1.05)},
      anchor=south,
      legend columns=5,
      column sep=3pt,
      font=\footnotesize,
    },
]

\addplot[
  sharp plot,
  orange!65,
  thick,
  line width=1.5pt,
  mark=pentagon*,
  mark size=3pt,
  mark options={fill=white,draw=orange!65,line width=1.5pt}
] coordinates {
  (ar,0.9) (th,2.9) (fr,1.2) (ja,18.7) (zh,87.0)
(en,99.8) (vi,5.7) (ko,9.1) (pt,1.4) (es,6.6)
};
\addlegendentry{Qwen2.5-7B-Instruct}

\addplot[
  sharp plot,
  ublue!80,
  thick,
  line width=1.5pt,
  mark=square*,
  mark size=2.6pt,
  mark options={fill=white,draw=ublue!80,line width=1.5pt}
] coordinates {
  (ar,94.8) (th,79.9) (fr,99.7) (ja,99.1) (zh,88.2)
(en,99.6) (vi,99.5) (ko,94.3) (pt,99.1) (es,99.3)
};
\addlegendentry{LCP}

\addplot[
  sharp plot,
  udporange!90,
  thick,
  line width=1.5pt,
  mark=*,
  mark size=3pt,
  mark options={fill=white,draw=udporange!90,line width=1.5pt}
] coordinates {
  (ar,84.6) (th,89.0) (fr,96.9) (ja,80.9) (zh,85.7)
(en,99.8) (vi,90.6) (ko,77.9) (pt,67.4) (es,88.0)
};
\addlegendentry{DIT}

\addplot[
  sharp plot,
 usemiblue!65,
  thick,
  line width=1.5pt,
  mark=triangle*,
  mark size=3pt,
  mark options={fill=white,draw=usemiblue!65,line width=1.5pt}
] coordinates {
  (ar,88.6) (th,80.3) (fr,77.1) (ja,98.9) (zh,85.8)
(en,99.7) (vi,95.7) (ko,67.0) (pt,62.6) (es,95.1)
};
\addlegendentry{QRT}

\addplot[
  sharp plot,
  udark!65,
  thick,
  line width=1.5pt,
  mark=pentagon*,
  mark size=3pt,
  mark options={fill=white,draw=udark!65,line width=1.5pt}
] coordinates {
  (ar,93.3) (th,84.6) (fr,90.2) (ja,86.1) (zh,85.9)
(en,99.7) (vi,91.0) (ko,90.5) (pt,91.8) (es,95.7)
};
\addlegendentry{M-SFT}

\addplot[
  sharp plot,
  usemidark!65,
  thick,
  line width=1.5pt,
  mark=*,
  mark size=3pt,
  mark options={fill=white,draw=usemidark!65,line width=1.5pt}
] coordinates {
  (ar,0.0) (th,3.4) (fr,0.0) (ja,0.3) (zh,8.8)
(en,99.9) (vi,0.7) (ko,0.0) (pt,0.0) (es,0.2)
};
\addlegendentry{Vanilla GRPO}

\addplot[
  sharp plot,
  udpdark!65,
  thick,
  line width=1.5pt,
  mark=triangle*,
  mark size=2.8pt,
  mark options={fill=white,draw=udpdark!65,line width=1.5pt}
] coordinates {
  (ar,98.9) (th,99.5) (fr,100.0) (ja,98.7) (zh,79.1)
(en,99.9) (vi,99.9) (ko,97.1) (pt,99.3) (es,99.3)
};
\addlegendentry{LC-GRPO}

\addplot[
  sharp plot,
  upurple!75,
  thick,
  line width=1.5pt,
  mark=square*,
  mark size=2.8pt,
  mark options={fill=white,draw=upurple!75,line width=1.5pt}
] coordinates {
  (ar,98.4) (th,97.9) (fr,99.6) (ja,99.4) (zh,79.1)
(en,99.7) (vi,98.8) (ko,98.8) (pt,99.4) (es,99.3)
};
\addlegendentry{M-Thinker}

\addplot[
  sharp plot,
  udpblue!40,
  thick,
  line width=1.5pt,
  mark=pentagon*,
  mark size=3pt,
  mark options={fill=white,draw=udpblue!40,line width=1.5pt}
] coordinates {
  (ar,65.4) (th,79.7) (fr,87.0) (ja,66.0) (zh,82.8)
(en,99.7) (vi,89.0) (ko,73.6) (pt,91.6) (es,95.7)
};
\addlegendentry{mGRPO}

\addplot[
  sharp plot,
  red!90,               
  thick,
  line width=1.5pt,
  mark=oplus,
  mark size=3pt,
  mark options={fill=white,draw=red!90,line width=1.5pt}
] coordinates {
  (ar,99.2) (th,99.4) (fr,100.0) (ja,98.7) (zh,89.1)
(en,99.9) (vi,99.9) (ko,96.9) (pt,99.3) (es,99.1)
};
\addlegendentry{\ours}
\end{axis}
\end{tikzpicture}
  \caption{The input-output language consistency ratio of different methods for each language on MMATH test sets with Qwen2.5-7B-Instruct model.}
  \label{table:mmath-7B-LC}
\end{figure*}

\begin{figure*}[htbp]
    \centering
    \begin{tikzpicture}
\begin{axis}[
    width=\linewidth,
    height=8cm,
    ymin=0, ymax=105,
    ytick={0,10,20,30,40,50,60,70,80,90,100},
    grid=major,
    major grid style={dotted},
    ylabel={Language Consistency Ratio},
    xlabel={Language Code},
    symbolic x coords={ar,bn,th,sw,ja,zh,de,fr,ru,en,te,ko,vi,it,id,pt,es},
    xticklabels={
      {\strut Ar},{\strut Bn},{\strut Th},{\strut Sw}, {\strut Ja},{\strut Zh},{\strut De},{\strut Fr},{\strut Ru},{\strut En},
      {\strut Te},{\strut Ko},{\strut Vi},{\strut It},{\strut Id},{\strut Pt},{\strut Es}
    },
xtick=data,
    xticklabel style={anchor=north},
    enlarge x limits=0.02,
    legend style={
      draw=none,
      at={(0.5,1.05)},
      anchor=south,
      legend columns=5,
      font=\footnotesize,
    },
]

\addplot[
  sharp plot,
  orange!65,
  thick, line width=1.5pt,
  mark=pentagon*, mark size=3pt,
  mark options={fill=white,draw=orange!65,line width=1.5pt}
] coordinates {
  (ar,89.6) (bn,85.5) (th,92.3) (sw,44.1) (ja,76.4)
  (zh,83.3) (de,88.6) (fr,91.6) (ru,83.6) (en,98.2)
  (te,79.4) (ko,91.6) (vi,89.0) (it,77.9) (id,48.9)
  (pt,86.5) (es,98.1)
};
\addlegendentry{Qwen2.5-3B-Instruct}

\addplot[
  sharp plot,
  ublue!80,
  thick,
  line width=1.5pt,
  mark=square*, mark size=2.6pt,
  mark options={fill=white,draw=ublue!80,line width=1.5pt}
] coordinates {
  (ar,99.7) (bn,98.6) (th,98.8) (sw,98.1) (ja,100.0)
  (zh,94.3) (de,99.4) (fr,99.5) (ru,99.7) (en,97.5)
  (te,98.6) (ko,98.8) (vi,100.0) (it,98.9) (id,99.6)
  (pt,99.1) (es,99.2)
};
\addlegendentry{LCP}

\addplot[
  sharp plot,
  udporange!90,
  thick,
  line width=1.5pt,
  mark=*, mark size=3pt,
  mark options={fill=white,draw=udporange!90,line width=1.5pt}
] coordinates {
  (ar,98.8) (bn,99.6) (th,99.5) (sw,95.7) (ja,99.4)
  (zh,90.1) (de,99.5) (fr,99.3) (ru,99.3) (en,97.9)
  (te,99.3) (ko,98.7) (vi,99.5) (it,98.6) (id,99.4)
  (pt,98.1) (es,99.0)
};
\addlegendentry{DIT}

\addplot[
  sharp plot,
  usemiblue!65,
  thick,
  line width=1.5pt,
  mark=triangle*, mark size=3pt,
  mark options={fill=white,draw=usemiblue!65,line width=1.5pt}
] coordinates {
  (ar,99.5) (bn,99.8) (th,99.7) (sw,98.0) (ja,99.3)
  (zh,93.9) (de,99.5) (fr,99.5) (ru,99.7) (en,99.0)
  (te,100.0) (ko,100.0) (vi,99.8) (it,99.1) (id,99.8)
  (pt,98.2) (es,99.7)
};
\addlegendentry{QRT}

\addplot[
  sharp plot,
  udark!65,
  thick,
  line width=1.5pt,
  mark=pentagon*, mark size=3pt,
  mark options={fill=white,draw=udark!65,line width=1.5pt}
] coordinates {
  (ar,87.2) (bn,89.8) (th,79.8) (sw,87.2) (ja,99.6)
  (zh,83.5) (de,96.4) (fr,97.8) (ru,99.6) (en,96.8)
  (te,89.0) (ko,91.3) (vi,79.9) (it,96.1) (id,86.2)
  (pt,96.4) (es,95.7)
};
\addlegendentry{M-SFT}

\addplot[
  sharp plot,
  usemidark!65,
  thick,
  line width=1.5pt,
  mark=*, mark size=2.5pt,
  mark options={fill=white,draw=usemidark!65,line width=1.5pt}
] coordinates {
  (ar,98.2) (bn,2.5) (th,99.6) (sw,83.9) (ja,96.7)
  (zh,92.5) (de,99.5) (fr,99.8) (ru,99.9) (en,100.0)
  (te,55.1) (ko,75.4) (vi,100.0) (it,97.9) (id,99.7)
  (pt,94.3) (es,99.4)
};
\addlegendentry{Vanilla GRPO}

\addplot[
  sharp plot,
  udpdark!65,
  thick,
  line width=1.5pt,
  mark=triangle*, mark size=2.8pt,
  mark options={fill=white,draw=udpdark!65,line width=1.5pt}
] coordinates {
  (ar,99.4) (bn,99.0) (th,100.0) (sw,99.7) (ja,98.5)
  (zh,91.5) (de,99.8) (fr,99.8) (ru,99.8) (en,99.8)
  (te,98.9) (ko,99.6) (vi,99.7) (it,99.0) (id,99.8)
  (pt,99.6) (es,99.5)
};
\addlegendentry{LC-GRPO}

\addplot[
  sharp plot,
  upurple!75,
  thick,
  line width=1.5pt,
  mark=square*, mark size=2.8pt,
  mark options={fill=white,draw=upurple!75,line width=1.5pt}
] coordinates {
  (ar,99.0) (bn,99.1) (th,99.6) (sw,97.9) (ja,98.4)
  (zh,89.1) (de,99.6) (fr,98.4) (ru,99.5) (en,98.0)
  (te,99.4) (ko,98.7) (vi,99.5) (it,99.4) (id,99.6)
  (pt,98.6) (es,99.1)
};
\addlegendentry{M-Thinker}

\addplot[
  sharp plot,
  udpblue!40,
  thick,
  line width=1.5pt,
  mark=pentagon*, mark size=3pt,
  mark options={fill=white,draw=udpblue!40,line width=1.5pt}
] coordinates {
  (ar,59.4) (bn,58.8) (th,69.8) (sw,47.2) (ja,88.5)
  (zh,89.1) (de,89.6) (fr,89.3) (ru,68.7) (en,99.1)
  (te,29.1) (ko,79.5) (vi,59.4) (it,47.5) (id,26.0)
  (pt,98.8) (es,89.7)
};
\addlegendentry{mGRPO}

\addplot[
  sharp plot,
  red!90,
  thick,
  line width=1.5pt,
  mark=oplus, mark size=3pt,
  mark options={fill=white,draw=red!90,line width=1.5pt}
] coordinates {
  (ar,99.8) (bn,99.6) (th,98.0) (sw,97.8) (ja,97.8)
  (zh,94.7) (de,98.8) (fr,99.5) (ru,99.1) (en,98.5)
  (te,99.4) (ko,98.1) (vi,99.6) (it,99.1) (id,99.3)
  (pt,97.5) (es,99.4)
};
\addlegendentry{\ours}

\end{axis}
\end{tikzpicture}
  \caption{The input-output language consistency ratio of different methods across languages and levels on PolyMath with Qwen2.5-3B-Instruct model.}
  \label{table:polymath-3B-LC}
\end{figure*}

\begin{figure*}[htbp]
    \centering
    \begin{tikzpicture}
\begin{axis}[
    width=\linewidth,
    height=7cm,
    ymin=0, ymax=105,
    ytick={0,10,20,30,40,50,60,70,80,90,100},
    grid=major,
    major grid style={dotted},
    ylabel={Language Consistency Ratio},
    xlabel={Language Code},
    symbolic x coords={ar,bn,th,sw,ja,zh,de,fr,ru,en,te,ko,vi,it,id,pt,es},
    xtick=data,
    xticklabels={
      {\strut Ar},{\strut Bn},{\strut Th},{\strut Sw}, {\strut Ja},{\strut Zh},{\strut De},{\strut Fr},{\strut Ru},{\strut En},
      {\strut Te},{\strut Ko},{\strut Vi},{\strut It},{\strut Id},{\strut Pt},{\strut Es}
    },
    xticklabel style={anchor=north},
    enlarge x limits=0.02,
    legend style={
      draw=none,
      line width=1.5pt,
      at={(0.5,1.05)},
      anchor=south,
      legend columns=5,
      column sep=3pt,
      font=\footnotesize,
    },
]

\addplot[
  sharp plot,
  orange!65,
  thick, line width=1.5pt,
  mark=pentagon*, mark size=3pt,
  mark options={fill=white,draw=orange!65,line width=1.5pt}
] coordinates {
  (ar,96.5) (bn,94.5) (th,97.9) (sw,50.4) (ja,97.9)
  (zh,83.3) (de,88.3) (fr,97) (ru,88.2) (en,99.5)
  (te,98.3) (ko,96.7) (vi,98.1) (it,97.4) (id,55.2)
  (pt,90.9) (es,90.9)
};
\addlegendentry{Qwen2.5-7B-Instruct}

\addplot[
  sharp plot,
  ublue!80,
  thick, line width=1.5pt,
  mark=square*, mark size=2.6pt,
  mark options={fill=white,draw=ublue!80,line width=1.5pt}
] coordinates {
  (ar,96.7) (bn,95.8) (th,100.0) (sw,96.3) (ja,99.9)
  (zh,89.4) (de,99.5) (fr,99.1) (ru,99.5) (en,100.0)
  (te,98.7) (ko,98.9) (vi,97.1) (it,98.0) (id,96.3)
  (pt,97.0) (es,96.8)
};
\addlegendentry{LCP}

\addplot[
  sharp plot,
  udporange!90,
  thick, line width=1.5pt,
  mark=*, mark size=3pt,
  mark options={fill=white,draw=udporange!90,line width=1.5pt}
] coordinates {
  (ar,96.6) (bn,98.9) (th,99.0) (sw,85.1) (ja,98.7)
  (zh,91.3) (de,99.0) (fr,99.6) (ru,99.4) (en,99.5)
  (te,98.8) (ko,98.6) (vi,99.0) (it,99.7) (id,97.7)
  (pt,98.5) (es,98.2)
};
\addlegendentry{DIT}

\addplot[
  sharp plot,
  usemiblue!65,
  thick, line width=1.5pt,
  mark=triangle*, mark size=3pt,
  mark options={fill=white,draw=usemiblue!65,line width=1.5pt}
] coordinates {
  (ar,97.0) (bn,99.8) (th,98.4) (sw,98.8) (ja,100.0)
  (zh,92.3) (de,99.4) (fr,99.8) (ru,99.4) (en,99.1)
  (te,97.7) (ko,97.2) (vi,98.3) (it,99.3) (id,100.0)
  (pt,99.5) (es,99.5)
};
\addlegendentry{QRT}

\addplot[
  sharp plot,
  udark!65,
  thick, line width=1.5pt,
  mark=pentagon*, mark size=3pt,
  mark options={fill=white,draw=udark!65,line width=1.5pt}
] coordinates {
  (ar,88.7) (bn,89.8) (th,99.4) (sw,86.5) (ja,99.4)
  (zh,75.3) (de,96.4) (fr,99.1) (ru,89.0) (en,98.3)
  (te,69.9) (ko,60.5) (vi,77.5) (it,76.6) (id,84.9)
  (pt,95.8) (es,93.7)
};
\addlegendentry{M-SFT}

\addplot[
  sharp plot,
  usemidark!65,
  thick, line width=1.5pt,
  mark=*, mark size=3pt,
  mark options={fill=white,draw=usemidark!65,line width=1.5pt}
] coordinates {
  (ar,94.6) (bn,8.6) (th,11.0) (sw,43.0) (ja,92.7)
  (zh,88.7) (de,99.9) (fr,98.2) (ru,97.1) (en,99.0)
  (te,98.8) (ko,82.3) (vi,96.3) (it,99.5) (id,98.3)
  (pt,95.2) (es,96.6)
};
\addlegendentry{Vanilla GRPO}

\addplot[
  sharp plot,
  udpdark!65,
  thick, line width=1.5pt,
  mark=triangle*, mark size=2.8pt,
  mark options={fill=white,draw=udpdark!65,line width=1.5pt}
] coordinates {
  (ar,100.0) (bn,99.6) (th,99.9) (sw,24.2) (ja,99.8)
  (zh,89.2) (de,99.7) (fr,100.0) (ru,99.8) (en,98.9)
  (te,99.6) (ko,99.3) (vi,99.7) (it,99.7) (id,100.0)
  (pt,99.6) (es,99.9)
};
\addlegendentry{LC-GRPO}

\addplot[
  sharp plot,
  upurple!75,
  thick, line width=1.5pt,
  mark=square*, mark size=2.8pt,
  mark options={fill=white,draw=upurple!75,line width=1.5pt}
] coordinates {
  (ar,99.8) (bn,99.6) (th,98.0) (sw,97.8) (ja,97.8)
  (zh,94.0) (de,98.8) (fr,99.5) (ru,99.1) (en,98.5)
  (te,99.4) (ko,98.1) (vi,99.6) (it,99.1) (id,99.3)
  (pt,97.3) (es,99.4)
};
\addlegendentry{M-Thinker}

\addplot[
  sharp plot,
  udpblue!40,
  thick, line width=1.5pt,
  mark=pentagon*, mark size=3pt,
  mark options={fill=white,draw=udpblue!40,line width=1.5pt}
] coordinates {
  (ar,78.0) (bn,88.8) (th,88.5) (sw,49.4) (ja,79.6)
  (zh,88.1) (de,95.6) (fr,99.6) (ru,78.6) (en,99.9)
  (te,45.8) (ko,77.2) (vi,69.6) (it,85.0) (id,45.8)
  (pt,94.1) (es,86.3)
};
\addlegendentry{mGRPO}

\addplot[
  sharp plot,
  red!90,
  thick, line width=1.5pt,
  mark=oplus, mark size=3pt,
  mark options={fill=white,draw=red!90,line width=1.5pt}
] coordinates {
  (ar,99.6) (bn,100.0) (th,100.0) (sw,97.9) (ja,100.0)
  (zh,91.5) (de,99.3) (fr,99.6) (ru,99.9) (en,99.7)
  (te,99.8) (ko,100.0) (vi,99.4) (it,99.1) (id,100.0)
  (pt,98.9) (es,99.9)
};
\addlegendentry{\ours}

\end{axis}
\end{tikzpicture}
  \caption{The input-output language consistency ratio of different methods across languages and levels on PolyMath with Qwen2.5-7B-Instruct model.}
  \label{table:polymath-7B-LC}
\end{figure*}

\end{document}